\documentclass[journal]{IEEEtran}
\usepackage{xcolor}
\usepackage{xpatch}

\usepackage{changes}
\usepackage{color}
\usepackage[para]{threeparttable}
\usepackage{footnote}
\usepackage{amssymb,amsmath}
\usepackage{colortbl}
\usepackage{lineno}

\usepackage[pagebackref=false,colorlinks]{hyperref}
\usepackage{cite}
\usepackage{hyperref}   

\usepackage{amsmath}

\usepackage{booktabs}
\usepackage{multirow}
\usepackage{caption}
\usepackage{microtype}

\usepackage{verbatim}
\usepackage{bm}

\usepackage{indentfirst}
\usepackage{graphicx}
\usepackage{float}
\usepackage{rotate}
\usepackage[section]{placeins}  

\usepackage{setspace} 
\usepackage{paralist}

\title{CbwLoss:Constrained Bidirectional Weighted Loss for Self-supervised Learning of Depth and Pose}

\author{Fei~Wang,~\IEEEmembership{Student Member,~IEEE}, Jun~Cheng,~\IEEEmembership{Member,~IEEE}, Penglei~Liu,~\IEEEmembership{Student Member,~IEEE}
\thanks{This manuscript was submitted to the Transactions on Intelligent Transportation Systems on May 9, 2021 and accepted by the Transactions on Intelligent Transportation Systems on February 20, 2023. \newline\indent (Corresponding author: Jun~Cheng.) \newline\indent Fei~Wang, Jun~Cheng, Penglei~Liu are with the Guangdong-Hong Kong-Macao Joint Laboratory of Human-Machine Intelligence-Synergy Systems, Shenzhen Institute of Advanced Technology, Chinese Academy of Sciences, Shenzhen, 518055, China. Fei~Wang, Penglei~Liu are also with University of Chinese Academy of Sciences, Beijing,100049, China. They are also with The Chinese University of Hong Kong (email:\{fei.wang2, jun.cheng, pl.liu\}@siat.ac.cn).}}

\begin{document}

\maketitle
\begin{abstract}
Photometric differences are widely used as supervision signals to train neural networks for estimating depth and camera pose from unlabeled monocular videos. However, this approach is detrimental for model optimization because occlusions and moving objects in a scene violate the underlying static scenario assumption. In addition, pixels in textureless regions or less discriminative pixels hinder model training. To address these problems, in this paper, we deal with moving objects and occlusions by utilizing the differences between the flow fields, and the differences between the depth structure generated by affine transformation and view synthesis, respectively. Secondly, we mitigate the effect of textureless regions on model optimization by measuring the differences between features with more semantic and contextual information without requiring additional networks. In addition, although the bidirectionality component is used in each sub-objective function, a pair of images is reasoned about only once, which helps reduce overhead. Extensive experiments and visual analysis demonstrate the effectiveness of the proposed method, which outperforms existing state-of-the-art self-supervised methods under the same conditions and without introducing additional auxiliary information.
 
	\end{abstract}


\IEEEpeerreviewmaketitle
  \vspace{-10pt}
\section{Introduction}\label{sec:intro}

\IEEEPARstart  {T}{he} estimation of depth and camera pose from monocular videos is a fundamental and valuable but challenging task with applications in mobile robot vision and navigation \cite{yasuda2020autonomous}, driverless cars \cite{arnold2019survey}, and other scenarios. In these scenarios, an odometer based on a wheel encoder, which is susceptible to cumulative error from imprecision of the angular measurements, wheel slippage and conversion of rotation into distance, can be replaced by the direct estimation of the camera pose from a sequence of consecutive images \cite{yasuda2020autonomous}. Furthermore, the depth information inferred from detailed object size and location information, which can be directly obtained from monocular videos without expensive depth sensors, can be used for precise location, object detection, and obstacle avoidance for autonomous vehicles \cite{arnold2019survey,wang20203d}. Compared with traditional methods \cite{eom2019temporally,chen2020denao,qi2020geonet++,park2019high,su2020monocular,fu2018deep}, although the existing methods benefitting from the powerful data fitting ability of deep neural networks can achieve competitive performance in depth prediction from video sequences, these methods require expensive depth sensors and considerable labor to obtain sufficiently large amounts of data labeled with pixel-level depth information or even require stereo video sequences \cite{park2019high} for network training. To solve these problems, researchers have recently attempted to jointly predict depth and camera pose in a self-supervised fashion by employing geometric priors directly learned from large amounts of easily accessible unlabeled videos captured using the least expensive, least restrictive, and most ubiquitous cameras \cite{wang2020unsupervised,zhang2020unsupervised,yang2020d3vo,li2020self,casser2019depth,ranjan2019competitive,bian2019unsupervised,godard2019digging,yin2018geonet,godard2017unsupervised,zhou2017unsupervised,luo2019every}.

The main principle of these self-supervised methods is that one can transform one frame into another frame based on the relative camera pose, expressed in terms of rotation angles and translations, as well as the camera intrinsic matrix and a depth map estimated using spatial transformer networks \cite{jaderberg2015spatial}; then the corresponding photometric differences can be utilized as the supervision signal for model optimization. However, the training of deep neural networks based on photometric differences requires some underlying assumptions, namely, that the scene is static (without moving objects) and that there are no occlusions between adjacent frames. The performance of geometric image reconstruction is limited if these underlying assumptions are violated. Regarding these challenges, recent works address them by leveraging prior knowledge (e.g. the velocity \cite{guizilini20203d}, geometry structure information \cite{bian2019unsupervised,bian2021unsupervised} of the objects) or multi-task learning \cite{ranjan2019competitive,wang2020unsupervised,chen2020denao,qi2020geonet++}, which either fails to handle objects with different velocities or requires additional overhead. In addition, the use of the photometric difference between a pixel warped from the reference frame and the corresponding pixel captured in the target frame is often problematic because pixels in textureless regions do not provide a good foundation for a neural network to find the global minimum. This problem could be mitigated, while either off-the-shelf stereo algorithms \cite{watson2019self} or additional networks \cite{zhou2019unsupervised,shu2020feature} are required. Furthermore, to take full advantage of the information contained in both images, stacked frames in normal order and stacked frames in reversed order \cite{bian2019unsupervised,bian2021unsupervised,gordon2019depth,li2020unsupervised} are individually fed to the network for predicting both forward and backward motion, whereas it is time-consuming and computationally expensive because stacked frames in normal and reversed order are all reasoned about by networks.

Considering these basic problems, in this paper, a bidirectional weighted photometric loss is first proposed to effectively handle moving objects and occlusions effectively while taking full advantage of the information contained in both the target and reference images to improve the robustness of the algorithm. Furthermore, benefitting from this bidirectional calculation, the estimated results can be well verified online. Specifically, the photometric loss is reweighted using both adaptive weights, which are obtained by measuring the difference between the estimated depth and the depth obtained through projection transformation (which should theoretically be consistent), and camera flow occlusion masks, which are based on our observation that corresponding pixels between adjacent frames should be similar if there is no occlusion but dissimilar in the presence of occlusion (as, in the latter case, the corresponding occluded pixels are not visible). In prior work \cite{gordon2019depth,li2020unsupervised}, the error between the motion fields, obtained by applying the networks on the frames in normal and reversed order, is minimized to deal with dynamic scenes. In this paper, (1) the bidirectional image reconstruction error (that is, differences between the target image and the synthesized target image, and the differences between the reference image and the reconstructed reference image) is employed as an optimization objective. During view synthesis, the synthesized target image is obtained by warping a reference frame $I_{ref}$ based on the depth  map $D_{tgt}$ predicted by DepthNet, the camera intrinsic matrix K, and the relative pose $T_{tgt\rightarrow ref}$ predicted by CameraNet, the reconstructed reference image is obtained by warping a target image $I_{tgt}$ based on the reference depth map $D_{ref}$ estimated by DepthNet, K and the relative pose $T_{ref \rightarrow tgt}$ obtained by calculating the inverse of $T_{tgt\rightarrow ref}$ rather than the pose predicted by CameraNet; (2) we weighted the bidirectional image reconstruction error utilizing the masks calculated from both camera flow consistency check and depth structure consistency check to effectively handle moving objects and occlusions; (3) although our method is bidirectional, only one-way prediction is required. Compared with existing work \cite{bian2021unsupervised,bian2019unsupervised}, where $T_{ref\rightarrow tgt}$ is obtained by applying CameraNet on the frames in reversed order again, it is economic because a pair of images is reasoned about by CameraNet only once.

Second, the photometric information in textureless regions (e.g. uniformly colored regions) can be ambiguous, in such cases, the features error, which depends on the extracted deep features from raw images by utilizing an encoder network, is more robust than per-pixel loss, which only relies on the low-level pixel information \cite{johnson2016perceptual}. Furthermore, dense feature loss can be employed as an auxiliary signal for image reconstruction loss based on color intensity \cite{zhan2018unsupervised}, as it can incorporate more semantic and contextual information by encoding larger-scale patterns. Accordingly, a bidirectional feature perception loss is proposed to prevent the image gradient from tending toward zero in textureless regions during network training, which can enhance the perception ability of the model for weakly textured areas. Compared with work \cite{zhan2018unsupervised}, our method eliminates the need to measure stereo feature maps and use additional pre-trained models for feature extraction.

Third, a bidirectional depth structure consistency loss is proposed to not only constrain the difference between the depth obtained from the multiview geometric transformation {\footnotesize $p_{ref} \sim K\hat T_{tgt\rightarrow ref}\hat D(p_{tgt})K^-p_{tgt}$} and the depth predicted using DepthNet from the corresponding reference frame but also minimize the difference between the depth obtained from the transformation {\footnotesize $p_{tgt} \sim K \hat T_{ref\rightarrow tgt}\hat D(p_{ref})K^-p_{ref}$ } and the depth estimated from the corresponding target frame. More importantly, the scale of depth can be kept consistent based on the above constraint and explicit constraints which force forward and backward pose to have a consistent scale by calculating the inverse.

Finally, the proposed constrained bidirectional weighted loss (CbwLoss), which is obtained by constraining the bidirectional weighted photometric loss using the bidirectional feature perception loss and the bidirectional depth structure consistency loss, is employed to guide the learning of the model's parameters.

In summary, the main contributions of this paper are summarized as follows:

\begin{itemize}
	\item 
	We proposed a scheme to deal with the moving objects in dynamic scenes. In this scheme, we locate the moving objects using both the camera flow consistency check and the depth structure consistency check. We then calculate the masks based on the consistency check and use these masks to weight photometric loss for reducing the contribution of the corresponding region, thereby satisfying the basic assumption of image reconstruction based on the static scene.
	
	\item 
	We proposed a simple and economic scheme to improve the robustness of the algorithm in weak texture regions. In this scheme, neither the stereo features extracted by the pre-trained model nor the additional trained auto-encoder networks are required. We directly utilize mid-level features obtained from depth estimation networks to define feature perception loss, aiming at enhancing the perception ability
	of the depth estimation networks in textureless regions without increasing overhead.
	
	\item 
	We propose a simple bidirectional scheme that not only explicitly constrains the bidirectional pose scale, but also maximizes the use of the limited data. Furthermore, although it is bidirectional, only one-way prediction is required.
\end{itemize}

This paper is organized as follows. In section \ref{sec:relatedwork}, we introduce the previous work on depth and camera pose estimation based on deep learning. The proposed algorithm is presented in section \ref{sec:method}. Experimental results are reported in section \ref{sec:experiment}. Finally, we conclude the paper in section \ref{sec:conclusion}.
 \vspace{-5pt}

\section{Related Work}\label{sec:relatedwork}
Recently, with the rapid development of deep learning and high-performance computing devices, artificial intelligence technology that uses deep neural networks to analyze scene depth and camera motion to accurately perceive the surrounding environment is increasingly playing an irreplaceable role in robot navigation \cite{yasuda2020autonomous} and autonomous driving \cite{arnold2019survey,wang20203d}. Therefore, methods of taking advantage of the remarkable learning ability of deep neural networks to estimate depth and camera pose from a large number of videos captured by the least expensive, most ubiquitous cameras have attracted considerable attention \cite{eigen2014depth,liu2015learning,fu2018deep,eom2019temporally,park2019high,chen2020denao,qi2020geonet++,su2020monocular,konda2015learning,wang2017deepvo,wang2018end,xue2019beyond,beauvisage2020multimodal,ju2021scene,garg2016unsupervised,godard2017unsupervised,yin2018geonet,ranjan2019competitive,luo2019every,wang2020unsupervised,godard2019digging,bian2019unsupervised,watson2019self,zhan2019self,zhao2020towards}. These methods can be divided into supervised and self-supervised depth-pose estimation methods depending on whether the ground truth is required.
\vspace{-10pt}
\subsection{Supervised Depth-Pose Estimation}

For depth estimation, Eigen et al. \cite{eigen2014depth} first predicted a depth map from a single image by employing two stacked deep neural networks --- a coarse-scale network and a fine-scale network. The coarse-scale network was used to estimate the depth of the scene at the global level, and then, the estimated depth was refined within local regions using the fine-scale network to allow fine-scale details to be incorporated into the global prediction. Soon afterwards, depth estimation was formulated as a structured learning problems by jointly combining convolutional neural networks and conditional random fields, but no geometric priors were considered \cite{liu2015learning}. In another work, the strong ordinal correlation of depth values was taken into account to constrain the objective function, and the depth estimation problem was cast as an ordinal regression problem \cite{fu2018deep}. Thereafter, to further improve the quality of dense depth maps, Eom et al. \cite{eom2019temporally} predicted depth maps by means of a two-stream convolutional neural network with convolutional gated recurrent units, which could leverage both temporal information and spatial information in video sequences, while Park et al. \cite{park2019high} predicted high-precision depth map by leveraging the complementary properties of light detection and ranging (LiDAR) point clouds and stereo images. More recently, various other approaches have been adopted that can greatly benefit the depth estimation task, such as attention mechanisms \cite{su2020monocular}, which can be applied to emphasize the interdependency between low-resolution features capturing long-range context and fine-grained features describing local context; the exploitation of the geometric relationships between depth and surface normals \cite{qi2020geonet++}; and the combination of an epipolar geometry constraint with auxiliary optical flow \cite{chen2020denao}.

For pose estimation, Konda et al. \cite{konda2015learning} designed the first convolutional network for predicting camera pose from visual information extracted from consecutive frames. Wang et al. \cite{wang2017deepvo} formulated camera pose estimation as a sequence learning problem in which poses are directly inferred from videos based on recurrent convolutional neural networks without adopting any module in the traditional visual odometry pipeline. Based on the method of \cite{wang2017deepvo}, the uncertainty of the camera pose has been considered and modeled using its covariance to correct the drift and bound the uncertainty \cite{wang2018end}. Most recently, a multimodal localization fusion framework \cite{ju2021scene}, in which LiDAR odometry and visual odometry are modeled simultaneously, has been proposed to obtain more robust results in a complex environment; however, the cost of the necessary calibration data and sensors is also increased under this framework.
\vspace{-10pt}
\subsection{Self-supervised Depth-Pose Estimation}
In contrast to supervised depth-pose estimation methods, in which depth estimation and pose estimation are treated independently as two unrelated problems, in self-supervised depth-pose estimation methods, these two problems are tightly coupled to model their correlations. Most importantly, the ground truth information, which is time consuming and laborious to acquire for real-world scenes, is not required.

As one possible approach, Garg et al. \cite{garg2016unsupervised} first explicitly reconstructed the reference frame by explicitly generating an inverse warp of the target image using the estimated depth and camera pose and then employed the reconstruction error as the supervision signal to train networks to estimate the depth map. The standard photometric warp loss was later improved by taking contextual information into consideration, instead of relying solely on per-pixel color matching \cite{zhan2018unsupervised}. However, in the above methods, not only the relative camera pose between the reference image and the target image must be known, but also stereo video sequences are required to reconstruct the reference view from the live view. In addition, although high-quality images can be reconstructed by minimizing the reconstruction error, the estimated depth map is of poor quality. To overcome this problem, a left-right consistency check \cite{godard2017unsupervised} has been proposed to improve the quality of synthesized depth images, but in this case, stereo images are required for network training. Different from the method in \cite{godard2017unsupervised}, Zhou et al. \cite{zhou2017unsupervised} directly estimated the depth map and camera pose from monocular videos in a fully unsupervised fashion by jointly training a depth network and a pose network for the first time. To achieve increased robustness to outliers and non-Lambertian regions, Yin et al. \cite{yin2018geonet} designed a cascaded architecture consisting of two stages to model static scenes and dynamic objects independently. Similarly, Ranjan et al. \cite{ranjan2019competitive} segmented the scene of interest into static and moving regions by employing additional segmentation networks. Specifically, the static scene in a video sequence can be analyzed based on a depth network and a pose network, while the whole scene consisting of both static and moving objects can be analyzed by employing an additional optical flow network. Then, the pixels in the scene can be assigned as belonging to either static or independently moving regions using segmentation networks. Similar to the above method \cite{ranjan2019competitive}, Luo et al. \cite{luo2019every} decomposed the scene of interest into background and foreground using three parallel networks: one to predict the camera motion, one to predict the dense depth map, and one to predict the per-pixel optical flow. Based  on the methods presented in \cite{ranjan2019competitive} and \cite{luo2019every}, a less-than-mean mask \cite{wang2020unsupervised} was subsequently designed to further exclude mismatched pixels disturbed by motion or illumination changes during the training of the depth and pose networks and was also used to exclude trivial mismatched pixels in the training of the optical flow network. 

In contrast to the above methods of explicitly segmenting scenes into static and dynamic object regions, Godard et al. \cite{godard2019digging} designed a per-pixel minimum reprojection loss, instead of averaging the photometric error over all source images, to handle occlusions. Bian et al. \cite{bian2019unsupervised} adopted a geometric consistency constraint to explicitly enforce scale consistency between different samples in order to handle dynamic scenes. In addition, to improve the quality of the predicted depth maps, Watson et al. \cite{watson2019self} enhanced the existing photometric loss using depth cues generated from off-the-shelf algorithms. Zhan et al. \cite{zhan2019self} jointly predicted depth and surface normals using an additional network. Zhao et al. \cite{zhao2020towards} directly solved the fundamental matrix based on optical flow correspondence and calculated the camera pose without PoseNet; in addition, a double view triangulation module was used to recover the up-to-scale scene structure. Subsequently, feature-metric loss \cite{shu2020feature}, which shares the same spirit as previous work \cite{zhan2018unsupervised}, was introduced to make up for the shortcomings that the standard photometric warp loss is not robust to uniformly textured areas. However, features, which are utilized to define feature-metric loss for depth estimation, must be explicitly learned by additional networks, resulting in increased overhead. Furthermore, only unidirectional warped features were considered. In order to further improve the quality of the estimated depth maps, either semantic information \cite{klingner2020self,guizilini2020semantically} obtained from additional semantic segmentation networks, which were jointly optimized with the depth estimation task, or the relations between depths \cite{jia2021self}, or more complex depth networks \cite{guizilini20203d} are also leveraged to improved depth prediction.

In parallel, recent work \cite{gordon2019depth,li2020unsupervised} has shown the consistency across neighboring frames could be well constrained by imposing the forward motion field\added{,} estimated by feeding a pair of frames in normal order to network, and the backward motion field\added{,} predicted by feeding the frames in reversed order to the same network again\added{,} to be the opposite of each other. Instead of directly constraining the motion field, the forward view differences between the source view and the warped view, obtained by transforming a neighboring view utilizing the forward relative pose and the backward photometric loss computed by inverting their roles again are imposed simultaneously \cite{bian2019unsupervised,bian2021unsupervised}. Despite these methods producing good results, a pair of frames need to be propagated twice across the network in order to get forward and backward either motion field or relative pose, resulting in increasing the training and inference overhead.

Notably, the above methods require either ground truth information for network training; additional networks (e.g., segmentation and/or optical flow networks) to filter out outliers, or off-the-shelf algorithms to enhance the photometric loss. Different from previous work \cite{gordon2019depth,li2020unsupervised,bian2019unsupervised,bian2021unsupervised}, in this paper, the relative pose, obtained by applying a network on frames in normal order, is reused by the inverse operation, and then is combined with camera intrinsic matrix and the corresponding predicted depth to obtain the corresponding synthetic frame, which is utilized to compute backward photometric loss. Instead of features generated from pre-trained models \cite{zhan2018unsupervised} or additional networks \cite{shu2020feature}, our method utilizes mid-features, which were obtained from depth estimation networks, to define feature perception loss, aiming at enhancing the perception ability of the model in textureless regions without increasing overhead. Similarly, the pose, obtained by applying a network on frames in normal order, is also utilized to compute both projected depth and the corresponding synthetic feature in the other direction by utilizing economic inverse operations rather than costly reasoning about the frames in reversed order again. Based on these, backward feature perception error and backward depth structure consistency error could be obtained. In addition, to mitigate the adverse impact of moving objects and occlusions, the camera flow occlusion masks based on flow fields are used to weight the above photometric loss.

\vspace{-5pt}
\section{Method}\label{sec:method}
\vspace{-5pt}

\begin{figure*}[t]	
	\vspace{-1.2cm}
	\centering
	\includegraphics[width=1\textwidth]{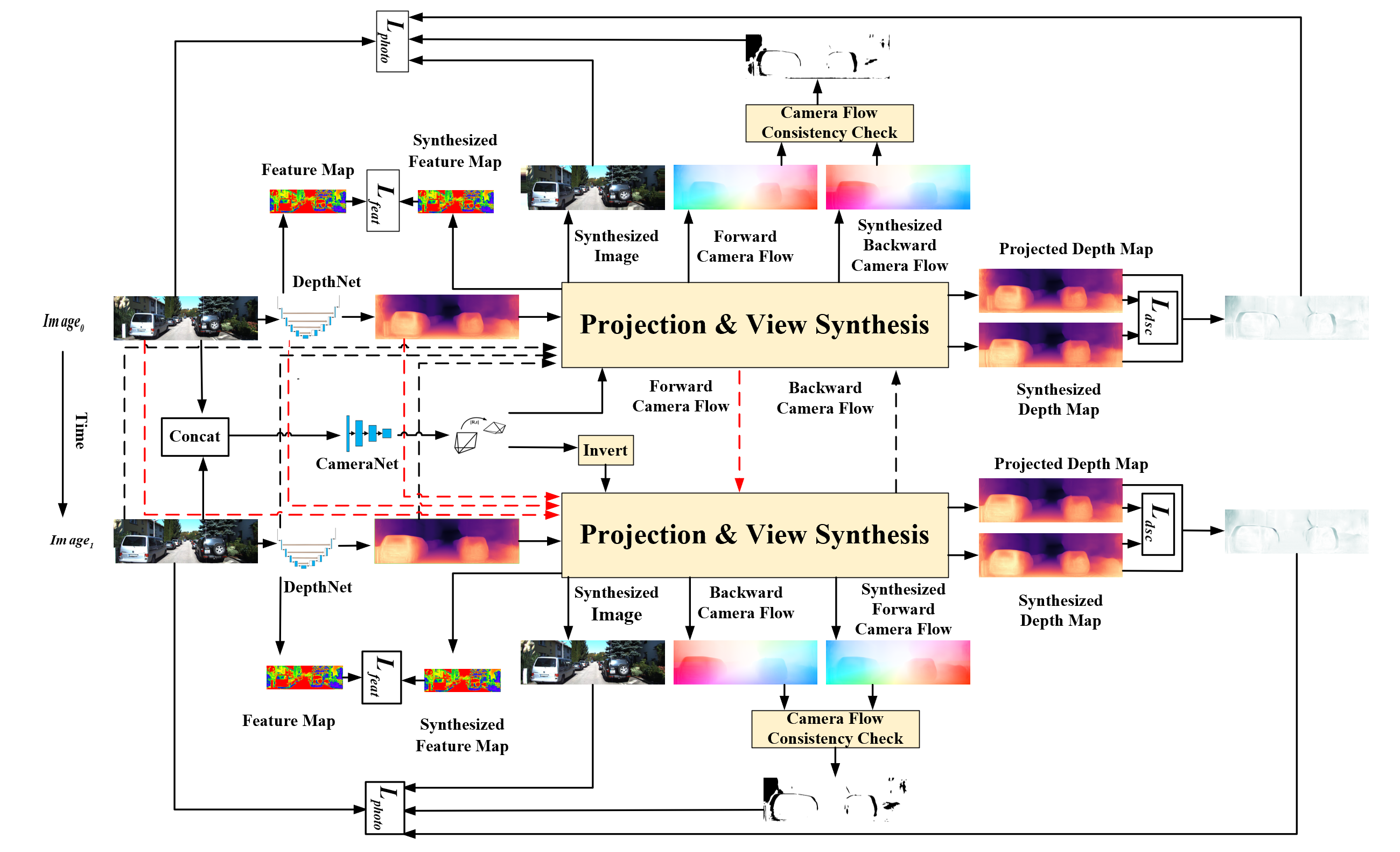}	
	\caption{Diagram of the general framework. Given two consecutive frames $I_0$ and $I_1$, the depth maps $D_0$ and $D_1$ can be estimated by DepthNet, the feature maps $f_0$ and $f_1$ of maximum resolution can be extracted from the corresponding images using the encoder network of DepthNet, and the camera pose $T$ can be estimated by CameraNet. Then the forward/backward camera flow and the synthesized forward/backward camera flow can be obtained for a consistency check based on projection transformation and a bilinear sampling mechanism. The corresponding image, depth map, and feature map can then be synthesized for the view reconstruction loss $L_{photo}$, the depth structure consistency loss $L_{dsc}$, and the feature perception loss $L_{feat}$.} \label{fig:overview}
	 
\end{figure*}

In this section, the theory of view synthesis is briefly introduced, and then, the bidirectional weighted photometric loss is presented. Afterwards, the bidirectional feature perception loss that is used to deal with weakly textured or textureless regions is described. Thereafter, the bidirectional depth structure consistency constraint is shown. The general framework of our proposed method is depicted in Fig. \ref{fig:overview}.

\vspace{-10pt}
\subsection{View Synthesis and Bidirectional Weighted Photometric Loss}
  
\subsubsection{View Synthesis and Bidirectional Photometric Loss}

\
\newline
\indent 
Our self-supervised learning problem is treated as a novel view synthesis problem. Specifically, consider a training image sequence ${I_1,I_2,...I_N}$, where one of the frames is the target frame $I_{tgt}$ and the remaining frames are reference frames $I_{ref}$ in a monocular video, with $1\leq ref \leq N,ref\neq tgt$. The photometric differences between the target frame $I_{tgt}$ and synthetic target frame $\hat I_{tgt}^{pose}$, obtained by warping a reference frame $I_{ref}$ based on the predicted depth map $D_{tgt}$, the camera intrinsic matrix $K$ and the predicted camera pose $T_{tgt\rightarrow ref}$, which is described in terms of camera rotation angles $\alpha ,\beta $ and $\gamma$ and translations $t_x ,t_y$ and $t_z$, are utilized as the supervision signal for model optimization. For convenience of description, suppose that the image coordinates of a point of interest are expressed as $img_{xy}^{tgt}=(x,y)$ and the predicted depth is $d_{tgt}$ at the image coordinates $img_{xy}^{tgt}$ then, we can transform the image coordinates into the world coordinates $img_{xyz}^{tgt}=(x_w^{tgt},y_w^{tgt},z_w^{tgt})$ in accordance with formula \eqref{eq_1}:

\vspace{-10pt}
\begin{subequations}
	\label{eq_1}
	\begin{equation}
	\begin{aligned}
	\label{eq_1_1}
	x_w^{tgt}=\frac{d_{tgt}}{f}(x-c_x)
	\end{aligned}
	\end{equation}
\vspace{-10pt}
	\begin{equation}
	\begin{aligned}
	\label{eq_1_2}
	y_w^{tgt}=\frac{d_{tgt}}{f}(y-c_y)
	\end{aligned}
	\end{equation}
\vspace{-10pt}
	\begin{equation}
	\begin{aligned}
	\label{eq_1_3}
	z_w^{tgt}=d_{tgt}
	\end{aligned}
	\end{equation}
\vspace{-10pt}
	\begin{equation}
	\begin{aligned}
	\label{eq_1_4}
	img_{xyz}^{tgt}=D_{tgt}K^{-}img_{xy}^{tgt}
	\end{aligned}
	\end{equation}
\end{subequations} 

where $(c_x,c_y,f)$ represents the principal point offset and focal length. Thereafter, the transformed world coordinates $\hat{img}_{xyz}^{ref}=(\hat x_w^{ref},\hat y_w^{ref},\hat z_w^{ref})$ can be calculated in accordance with formula \eqref{eq_2}:
\begin{equation}
	\begin{aligned}	
\hat{img}_{xyz}^{ref}=R_xR_yR_zimg_{xyz}^{tgt}+t
	   \label{eq_2} \\
\end{aligned} 
\end{equation}

where $(R_x,R_y,R_z,t)\in SE3$ denotes a $3D$ rotation and translation belonging to a special Euclidean group\footnote{\scriptsize A special Euclidean group is a set of Euclidean transformations denoted by an algebraic structure composed of sets and operations.}, constituting a homogeneous transformation matrix $T_{tgt\rightarrow ref}$. Thereafter, the transformed image coordinates $\hat {img}_{xy}^{ref}=(\hat x,\hat y)$ and the backward camera flow $u_{b}$ can be obtained according to formula \eqref{eq_3}, similarly, the forward camera flow $u_f$ can be also acquired.

\vspace{-5pt}
\begin{subequations}
	\label{eq_3}
	\begin{equation}
	\begin{aligned}
	\label{eq_3_1}
	\hat x=\frac{f}{\hat z_w^{ref}}+c_x
	\end{aligned}
	\end{equation}
\vspace{-5pt}
	\begin{equation}
	\begin{aligned}
	\label{eq_3_2}
	\hat y=\frac{f}{\hat z_w^{ref}}+c_y
	\end{aligned}
	\end{equation}
\vspace{-5pt}
	\begin{equation}
	\begin{aligned}
	\label{eq_3_3}
	\hat {img}_{xy}^{ref}=KT_{tgt\rightarrow ref} {img}_{xyz}^{tgt}
	\end{aligned}
	\end{equation}
\vspace{-5pt}
	\begin{equation}
	\begin{aligned}
	\label{eq_3_4}
	u_{b}=(\hat x-x,\hat y-y)
	\end{aligned}
	\end{equation}
\end{subequations} 

Based on the transformed image coordinates $\hat {img}_{xy}^{ref}$, the value of the synthesized target frame $\hat I_{tgt}^{pose}$ can be obtained via the differentiable bilinear sampling mechanism proposed in \cite{jaderberg2015spatial}. Similarly, the transformed image coordinates $\hat {img}_{xy}^{tgt}$ can be calculated, and the synthetic reference frame $\hat I_{ref}^{pose}$ can be also obtained by warping the target frame $I_{tgt}$ based on the camera intrinsic matrix $K$, the predicted depth map $D_{ref}$, and the computed camera pose $T_{ref\rightarrow tgt}$. Thus, the bidirectional photometric loss can be formulated as shown in formula \eqref{eq_4}:
\begin{equation}
\begin{aligned}
\label{eq_4}
L_{photo}^{bi} =L_{photo}^{ref\rightarrow tgt}(I_{tgt},\hat I_{tgt}^{pose}) 
+ L_{photo}^{tgt\rightarrow ref}(\hat I_{ref}^{pose},I_{ref}) 
\end{aligned}
\end{equation}

where $L_{photo}^{ref\rightarrow tgt}(\cdot)$ and $L_{photo}^{tgt\rightarrow ref}(\cdot)$ are the corresponding photometric error functions that measure the differences between the target frame $I_{tgt}$ and the corresponding synthesized frame $\hat I_{tgt}^{pose}$ and between the reference frame $I_{ref}$ and the corresponding synthesized frame $\hat I_{ref}^{pose}$, respectively.

As is common practice \cite{ranjan2019competitive,zhou2019unsupervised,pillai2019superdepth,godard2017unsupervised}, we also adopt the robust image similarity measure (the structural similarity index measure, SSIM) \cite{wang2004image} shown in formula \eqref{eq_5} for the photometric losses shown in the formulas \eqref{eq_6} and \eqref{eq_7}, while the robust error function shown in formula \eqref{eq_8} is adopted instead of the $L1$ norm.
\vspace{-1pt}
\begin{equation}
\begin{aligned}
\label{eq_5}
SSIM(a,b) = \frac{(2\mu_{a}\mu_{b}+c_1)(2\delta_{ab}+c_2)}{(\mu_a^2+\mu_b^2+c_1)(\delta_a^2+\delta_b^2+c_2)}
\end{aligned}
\end{equation}
\vspace{-1pt}
\begin{small}
\begin{equation}
\begin{aligned}
\label{eq_6}
 L_{photo}^{ref\rightarrow tgt}(I_{tgt},\hat I_{tgt}^{pose})  &= \alpha\frac{1-SSIM(I_{tgt},\hat I_{tgt}^{pose})}{2} \\&
+(1-\alpha)ERF(I_{tgt},\hat I_{tgt}^{pose})
\end{aligned}
\end{equation}
\end{small}
\vspace{-1pt}
\begin{small}
\begin{equation}
\begin{aligned}
\label{eq_7}
L_{photo}^{tgt\rightarrow ref}(\hat I_{ref}^{pose},I_{ref} &= \alpha\frac{1-SSIM(I_{ref},\hat I_{ref}^{pose})}{2} \\ &+(1-\alpha)ERF(I_{ref},\hat I_{ref}^{pose})
\end{aligned}
\end{equation}
\end{small}
\vspace{-1pt}
\begin{equation}
\begin{aligned}
\label{eq_8}
ERF(m,n)=\sqrt {(m-n)^2+\epsilon^2}
\end{aligned}
\end{equation}

Here $\mu$ and $\delta$ are the local mean and variance, respectively,  over the pixel neighborhood with $c_1=0.01^2$ and $c_2=0.03^2$; $\alpha$ is taken to be 0.85; and $ERF(\cdot)$ is our robust error measure function, where $\epsilon=0.01$ in this paper.

\subsubsection{Bidirectional Camera Flow Occlusion Masks}

\
\newline
\indent 
The prerequisites that there are no occlusions or moving objects in the scene of interest need to be satisfied when the photometric error is employed as the supervision signal to optimize a model for estimating depth and camera pose from large amounts of video data. If any of these assumptions are violated during neural network training, the gradients could be disrupted, impeding the training process. To mitigate the adverse impact from moving objects and occlusions, we propose bidirectional camera flow occlusion masks based on the observation that in general, the pixels in one frame should be similar to the pixels in another consecutive frame; however, in the case of occlusion, the pixels should not be similar because the corresponding pixels in the occluded frame are not visible. Similar to the optical flow estimation tasks \cite{meister2018unflow}, pixels will be marked as occlusions whenever the mismatch between different flow fields occurs. Nevertheless, different from the method \cite{meister2018unflow} where flow fields between adjacent frames are directly estimated by FlowNetC, our flow fields are generated from the corresponding transformed image coordinates, which are obtained during affine transformation utilizing the estimated relative pose. Specifically, the corresponding backward camera flow $u_{b}$ can be computed according to formula \eqref{eq_3_4}; then, the synthetic forward camera flow $\hat u_{f}$ can be obtained by both the transformed image coordinates $\hat {img}_{xy}^{ref}$ and the forward camera flow $u_{f}$ via the differentiable bilinear sampling mechanism \cite{jaderberg2015spatial}. Thereafter, the backward camera flow occlusion mask can be defined as shown in formula \eqref{eq_9}. Using a similar scheme, the synthetic backward camera flow $\hat u_{b}$ can be acquired based on $\hat {img}_{xy}^{tgt}$ and the backward camera flow $u_b$, and then the forward camera flow occlusion mask can be defined as shown in formula \eqref{eq_10}.
\vspace{-1pt}
\begin{equation}
\begin{aligned}
\label{eq_9}
M_{occ}^{ref\rightarrow tgt}=\Gamma(\|u_{b}+\hat {u}_{f}\|^2,\alpha_1(\|u_b\|^2+\|\hat u_{f}\|^2)+\alpha_2)
\end{aligned}
\end{equation}
\vspace{-10pt}
\begin{equation}
\begin{aligned}
\label{eq_10}
 M_{occ}^{tgt\rightarrow ref}=\Gamma(\|u_f+\hat u_b\|^2, 
\alpha_1(\|u_f\|^2+\|\hat u_b\|^2)+\alpha_2)
\end{aligned}
\end{equation}

where $\Gamma(a,b)$ represents the indicator function defined in formula \eqref{eq_indicator}. We set $\alpha_1=0.01$ and $\alpha_2=0.5$ in all our experiments.
\vspace{-10pt}
\begin{equation} 
\begin{aligned}
\label{eq_indicator}
\Gamma(a,b)=\begin{cases}
	 1,& a \textless b \cr 0, & otherwise
\end{cases}
\end{aligned}
\end{equation}

Based on the bidirectional camera occlusion masks shown in formulas \eqref{eq_9} and \eqref{eq_10}, adaptive weights (described in detail in formulas \eqref{eq_aw_ref2tgt} and \eqref{eq_aw_tgt2ref} in subsection \ref{subssec:depth_structure_consistency}), and the bidirectional photometric loss shown in formula \eqref{eq_4}, the bidirectional weighted photometric loss can be defined as shown in  formula \eqref{eq_biw_loss}.
\begin{equation} 
	\begin{aligned}
	\label{eq_biw_loss}
		L_{photo}^{biw} &=\lambda_p^{tgt}*M_{valid}^{ref \rightarrow tgt}*W_{aw}^{ref\rightarrow tgt}*L_{photo}^{ref\rightarrow tgt} \\ 
		&+ \lambda_p^{ref}*M_{valid}^{tgt\rightarrow ref}*W_{aw}^{tgt\rightarrow ref}*L_{photo}^{tgt\rightarrow ref} 
	\end{aligned}
\end{equation}

\begin{subequations}
	\label{eq_13M}
	\begin{equation}
	\begin{aligned}
	\label{eq_13M_1}
	M_{valid}^{ref \rightarrow tgt}=1-\lambda_{occ}^{tgt} * M_{occ}^{ref \rightarrow tgt}
	\end{aligned}
	\end{equation}
\vspace{-8pt}
	\begin{equation}
	\begin{aligned}
	\label{eq_13M_2}
	M_{valid}^{tgt\rightarrow ref}=1-\lambda_{occ}^{ref} * M_{occ}^{tgt\rightarrow ref}
	\end{aligned}
	\end{equation}	
\end{subequations}

\subsection{Bidirectional Feature Perception Loss}

The photometric error between the target frame and the corresponding frame synthesized from the reference frame is employed as the supervision signal to update the gradients and weights of the model. Therefore, the gradients play a crucial role during neural network training. Here, the loss function is reanalyzed from the gradient update perspective. To simplify the description, we analyze only the photometric loss $L_{photo}^{ref\rightarrow tgt} =L_{photo}^{ref\rightarrow tgt}(I_{tgt}(p),\hat I_{tgt}^{pose}(\hat p|T,D))$. Based on the chain rule, we can express the gradients of  $L_{photo}^{ref\rightarrow tgt}$ with respect to the depth $D$ and the camera pose $T$ as shown in formula \eqref{eq_12}.

\begin{scriptsize}
\begin{subequations}
	\label{eq_12}
	\begin{equation}
	\begin{aligned}
	\label{eq_12_1}
	\frac{\partial L_{photo}^{ref\rightarrow tgt}}{\partial D} =\frac{\partial L_{photo}^{ref\rightarrow tgt}}{\partial \hat I_{tgt}^{pose}(\hat p|T,D)} *\frac{\partial \hat I_{tgt}^{pose}(\hat p|T,D)}{\partial \hat p}*\frac{\partial \hat p}{\partial D} 
	\end{aligned}
	\end{equation}

	\begin{equation}
	\begin{aligned}
	\label{eq_12_2}
	\frac{\partial L_{photo}^{ref\rightarrow tgt} }{\partial T}=\frac{\partial L_{photo}^{ref\rightarrow tgt}}{\partial \hat I_{tgt}^{pose}(\hat p|T,D))} *\frac{\partial \hat I_{tgt}^{pose}(\hat p|T,D))}{\partial \hat p}*\frac{\partial \hat p}{\partial T}
	\end{aligned}
	\end{equation}	
\end{subequations} 
\end{scriptsize}

As seen from formula \eqref{eq_12}, the gradient $\frac{\partial L_{photo}^{ref\rightarrow tgt}}{\partial D}$ and the gradient $\frac{\partial L_{photo}^{ref\rightarrow tgt} }{\partial T}$ both depend on the
image gradient $\frac{\partial \hat I_{tgt}^{pose}(\hat p|T,D)}{\partial \hat p}$.
For textureless regions, the image gradients are close to zero and thus make no contribution to the gradients $\frac{\partial L_{photo}^{ref\rightarrow tgt}}{\partial D}$ and $\frac{\partial L_{photo}^{ref\rightarrow tgt} }{\partial T}$, thereby hindering network training. However, the deep features extracted from images by the encoder network can encode larger-scale patterns in the images, with redundancies and noise removed. Therefore, these features are more discriminative than the raw RGB image features for textureless regions. Features error is more robust than the per-pixel loss \cite{johnson2016perceptual} in textureless regions. Furthermore, dense feature loss can be used as an auxiliary signal for image reconstruction loss based on color intensity \cite{zhan2018unsupervised,shu2020feature}. Instead of features generated from pre-trained models \cite{johnson2016perceptual,zhan2018unsupervised} or by training additional auto-encoder networks \cite{shu2020feature}, we directly utilize mid-level features obtained from depth estimation networks to define feature perception loss aiming at enhancing the perception ability of the depth estimation networks in textureless regions without increasing overhead. Inspired by the concept of view synthesis, we can force networks to pay more attention to textureless regions by simultaneously minimizing the difference between these features. More precisely, given a target image $I_{tgt}$ and a reference image $I_{ref}$, the corresponding target features $f_{tgt}$ and reference features $f_{ref}$ can be extracted by the encoder network, and then, the target features $\hat f_{tgt}$ can be synthesized from the reference features $f_{ref}$ based on the transformed image coordinates $\hat {img}_{xy}^{ref}$ using the differentiable bilinear sampling mechanism. Similarly, the reference features $\hat f_{ref}$ can also be synthesized based on the target features $f_{tgt}$ and the image coordinates $\hat {img}_{xy}^{tgt}$. Accordingly, the photometric loss function can be constrained by measuring the differences between the target features $f_{tgt}$ obtained by encoding the target frame $I_{tgt}$ and the synthesized target features $\hat f_{tgt}$ and between the reference features $f_{ref}$ obtained by encoding the reference frame $I_{ref}$ and the synthesized reference features $\hat f_{ref}$. Here, we define this constraint as the bidirectional feature perception loss, formulated as shown in formula \eqref{eq_13}.

\vspace{-1pt}
\begin{equation}
	\begin{aligned}
		\label{eq_13}
		L_{feat}^{bi}=\lambda_{feat}^{tgt}*\|f_{tgt}-\hat f_{tgt}\| 
		+\lambda_{frat}^{ref}*\|f_{ref}-\hat f_{ref}\|
	\end{aligned}
\end{equation}

\subsection{Bidirectional Depth Structure Consistency Loss}\label{subssec:depth_structure_consistency}

Given a target image $I_{tgt}$, the corresponding transformed world coordinates $\hat{img}_{xyz}^{ref}=(\hat x_w^{ref},\hat y_w^{ref},\hat z_w^{ref})$ can be obtained according to formula \eqref{eq_1_4}. The depths in the target and reference images can be estimated by DepthNet and are denoted by $d_{net}^{tgt}$ and $d_{net}^{ref}$, respectively. Because CameraNet is naturally coupled with DepthNet during training, the computed depth $\hat z_w^{ref}$ and the estimated depth $d_{net}^{ref}$ should conform to the same 3D scene structure and should be consistent. However, as shown in Fig. \ref{fig:depthocc}, the depths $\hat z_w^{ref}$ and $d_{net}^{ref}$ are not always equal, especially in regions with moving objects and occlusions. Intuitively, we can enforce consistency by minimizing the difference between $\hat z_w^{ref}$ and $d_{net}^{ref}$. In addition, moving objects and occlusions can be located using this difference. Formulas \eqref{eq_1} and \eqref{eq_3} show that the computed depth is only affected by camera pose, and the depth of another frame because the camera intrinsic is a fixed constant. Nevertheless, the predicted camera pose scale, inverse pose scale, and predicted depth scale are all unknown. Recent work \cite{bian2019unsupervised} enforces predicted depth scale consistency between each consecutive image by geometry consistency constraint, whereas this constraint does not guarantee that the depth scale of adjacent frames is completely consistent because the predicted camera pose and predicted inverse pose have inconsistent scales and do not necessarily satisfy the condition of invertibility. For example, given two consecutive frames $I_1,I_2$, the predicted depth is $D_1,D_2$. Assume that the scale factors, which are used to align the predicted depth to the absolute scale depth, are $s_{d1}$ and $s_{d2}$ respectively. The predicted relative pose and the predicted relative inverse pose are $T_1,T_2$, and the corresponding scale factors are $s_{t1},s_{t2}$, respectively. Based on formulas \eqref{eq_1} and \eqref{eq_3}, we can compute the depth $\hat D_1$ and $\hat D_2$ of $I_1$ and $I_2$ respectively. And assume the scale factors are $\hat s_{d1}$ and $\hat s_{d2}$. Then they should satisfy the following relationships. 
\begin{subequations}
	\label{eq_scale}
	\begin{equation}
	\begin{aligned}
	\label{eq_scale_1}
	\hat s_{d1}=s_{t1}*s_{d2}
	\end{aligned}
	\end{equation}
	\vspace{-8pt}
	\begin{equation}
	\begin{aligned}
	\label{eq_scale_2}
	\hat s_{d2}=s_{t2}*s_{d1}
	\end{aligned}
	\end{equation}	
\end{subequations} 

The constraint of $\|\hat D_1-D_1\|$  and  $\|\hat D_2-D_2\|$ will only drive $\hat s_{d1}$ closer to $s_{d1}$ and $\hat s_{d2}$ closer to $s_{d2}$. However, forcing $s_{t1}*s_{d2}=s_{d1}$ or $s_{t2}*s_{d1}=s_{d2}$ does not guarantee consistency of depth between adjacent frames because $s_{t1}$ and $s_{t2}$ are unknown and the predicted $T_1$ and the predicted $T_2$ are not guaranteed to be invertible. In order to ensure that this condition is satisfied, we explicitly constrain these two poses to be invertible by means of jointly estimating and computing, aiming at ensuring the depth scale of adjacent frames to be completely consistent.
Note, however, that we cannot directly compute the difference between $\hat z_w^{ref}$ and $d_{net}^{ref}$ because the estimated depth does not depend on the pixel grid. Therefore, we instead minimize the difference between $\hat z_w^{ref}$ and $d_{net\_{interp}}^{ref}$, which is obtained based on the predicted depth map $d_{net}^{ref}$ and the grid coordinates $\hat {img}_{xy}^{ref}=(\hat x,\hat y)$, obtained by formula \eqref{eq_3}, through bilinear interpolation. For pixel $p_{ref}$, the depth structure difference is defined as follows:

\begin{footnotesize}
	\begin{equation}
	\begin{aligned}
	\label{eq_17}
	depth_{diff}^{ref\rightarrow tgt}(p_{ref}) =\frac{ERF(\hat z_w^{ref},d_{{net}\_{interp}}^{ref}(p_{ref}))}{\hat z_w^{ref}+d_{net\_{interp}}^{ref}(p_{ref})}
	\end{aligned}
	\end{equation}
\end{footnotesize}
where $ERF(\cdot)$ is the robust error function shown in formula \eqref{eq_8}.

Similarly, for pixel $p_{tgt}$, the corresponding transformed world coordinates $\hat{img}_{xyz}^{tgt}=(\hat x_w^{tgt},\hat y_w^{tgt},\hat z_w^{tgt})$ and image coordinates $\hat {img}_{xy}^{tgt}$ can be acquired, and then the depth structure difference between $\hat z_w^{tgt}$ and $d_{net}^{tgt}$ can be obtained as shown in formula \eqref{eq_18}:

\begin{footnotesize}
	\begin{equation}
	\begin{aligned}
	\label{eq_18}
	depth_{diff}^{tgt\rightarrow ref}(p_{tgt}) =\frac{ERF(\hat z_w^{tgt},d_{{net}\_{interp}}^{tgt}(p_{tgt}))}{\hat z_w^{tgt}+d_{net\_{interp}}^{tgt}(p_{tgt})}
	\end{aligned}
	\end{equation}
\end{footnotesize}

Then, the bidirectional depth structure consistency loss is defined as shown in formula \eqref{eq_20}, and the adaptive weights, which are obtained based on the depth differences, are defined as shown in formulas \eqref{eq_aw_ref2tgt} and \eqref{eq_aw_tgt2ref}:

\vspace{-10pt}
\begin{equation}
\begin{aligned}
\label{eq_20}
& L_{dsc}^{bi} =\lambda_{dsc}^{tgt}*L_{dsc}^{ref\rightarrow tgt} + \lambda_{dsc}^{ref}*L_{dsc}^{tgt\rightarrow ref} \\ &
=\lambda_{dsc}^{tgt}*\frac{ \sum depth_{diff}(p_{tgt})}{N_{tgt}} 
+ \lambda_{dsc}^{ref}*\frac{\sum depth_{diff}(p_{ref}) }{N_{ref}}
\end{aligned}
\end{equation}

\begin{equation}
\begin{aligned}
\label{eq_aw_ref2tgt}
W_{aw}^{ref\rightarrow tgt} = 1-\lambda_{aw}^{tgt}*depth_{diff}^{ref\rightarrow tgt}(p_{ref})
\end{aligned}
\end{equation}
\vspace{-10pt}
\begin{equation}
\begin{aligned}
\label{eq_aw_tgt2ref}
W_{aw}^{tgt\rightarrow ref} = 1-\lambda_{aw}^{ref}*depth_{diff}^{tgt\rightarrow ref}(p_{tgt})
\end{aligned}
\end{equation}
where $N_{ref}$ and $N_{tgt}$ denote the numbers of valid grid coordinates $\hat {img}_{xy}^{ref}$ and $\hat {img}_{xy}^{tgt}$ respectively.

\subsection{Smoothness Loss}

As is common practice \cite{bian2019unsupervised,ranjan2019competitive,godard2019digging}, a smoothness loss is also employed as a regularizer for the estimated depth maps and feature maps. Here, an edge-aware term is used to weight the cost based on the depth map gradients and feature map gradients. The smoothness loss is formulated as follows:

\begin{equation}
\begin{aligned}
\label{eq_22}
 L_{s}^{bi} &=  
\lambda_f^{tgt}*\sum|\partial f_{tgt} |*e^{-|\partial I_{tgt}|} \\&
 +\lambda_f^{ref}*\sum |\partial  f_{ref}|*e^{-{|\partial I_{ref}}|} \\& +
 \lambda_d^{ref}*\sum | \partial depth_{ref}|*e^{-| \partial I_{ref}|} \\& +\lambda_d^{tgt}*\sum | \partial depth_{tgt} |*e^{-|\partial I_{tgt}|}
\end{aligned}
\end{equation}

\subsection{Summary}

The proposed CbwLoss is the bidirectional weighted photometric loss constrained by both the bidirectional feature perception loss and the bidirectional depth structure consistency loss, as shown in formula \eqref{eq_cbwloss}.
\begin{equation}
\begin{aligned}
	\label{eq_cbwloss}
L_{Cbw}=& L_{photo}^{biw} + L_{feat}^{bi} +L_{dsc}^{bi}
\end{aligned}
\end{equation}
The total loss is shown in formula \eqref{eq_total}.

\begin{equation} 
\begin{aligned}
\label{eq_total}
L_{total}=L_{Cbw} + L_{s}^{bi}
\end{aligned}
\end{equation}

\section{Experiments}\label{sec:experiment}
\subsection{Dataset}
We conducted experiments on the KITTI RAW dataset \cite{geiger2013vision} and the KITTI Odometry dataset \cite{geiger2012we} DDAD dataset \cite{guizilini20203d}. Similar to previous related work \cite{zhou2017unsupervised,yin2018geonet,ranjan2019competitive,godard2017unsupervised}, the KITTI RAW dataset was split as in \cite{eigen2014depth}, with approximately 40k frames used for training and 5k frames used for validation. The images were resized to $256 \times 832$ for depth estimation and camera pose estimation experiments. We evaluated DepthNet on test data consisting of 697 test frames in accordance with Eigen's testing split and tested CameraNet on sequences $09-10$ of the KITTI Odometry dataset. We also evaluate the proposed method on the improved ground-truth depths dataset \cite{uhrig2017sparsity}. The DDAD dataset was split as in \cite{guizilini20203d}, with 17,050 frames used for training and 4,150 frames used for evaluation.
\vspace{-12pt}
\subsection{Network Architectures}

\begin{table*}[h]\small
	\setlength\tabcolsep{4pt}
	\centering
	\begin{tabular}{lcccccccccccccccccccccccccccc} 
		\toprule 
		\multicolumn{1}{l}{\multirow{1}*{\small Method}}&
		\multicolumn{1}{c}{\multirow{1}*{\small $\lambda_{p}^{tgt}$}}&
		\multicolumn{1}{c}{\multirow{1}*{\small $\lambda_{p}^{ref}$}}&
		\multicolumn{1}{c}{\multirow{1}*{\small $\lambda_{occ}^{tgt}$}}&
		\multicolumn{1}{c}{\multirow{1}*{\small $\lambda_{occ}^{ref}$}}&
		\multicolumn{1}{c}{\multirow{1}*{\small $\lambda_{aw}^{tgt}$}}&
		\multicolumn{1}{c}{\multirow{1}*{\small $\lambda_{aw}^{ref}$}}&
		\multicolumn{1}{c}{\multirow{1}*{\small $\lambda_{dsc}^{tgt}$}}&
		\multicolumn{1}{c}{\multirow{1}*{\small $\lambda_{dsc}^{ref}$}}&
		\multicolumn{1}{c}{\multirow{1}*{\small $\lambda_{feat}^{tgt}$}}&
		\multicolumn{1}{c}{\multirow{1}*{\small $\lambda_{feat}^{ref}$}}&
		\multicolumn{1}{c}{\multirow{1}*{\small $\lambda_{d}^{tgt}$}}&
		\multicolumn{1}{c}{\multirow{1}*{\small $\lambda_{d}^{ref}$}}&
		\multicolumn{1}{c}{\multirow{1}*{\small $\lambda_{f}^{tgt}$}}&
		\multicolumn{1}{c}{\multirow{1}*{\small $\lambda_{f}^{ref}$}}& \\
		\midrule
		\multicolumn{1}{l}{{\small Baseline}}&1.0&0.0&0.0&0.0&0.0&0.0&0.0&0.0&0.0&0.0&0.01&0.0&0.0&0.0\\ 
		\multicolumn{1}{l}{{\small $L_{p}^{bi}$}}&1.0&1.0&0.0&0.0&0.0&0.0&0.0&0.0&0.0&0.0&0.01&0.01&0.0&0.0\\

		\multicolumn{1}{l}{{{\small $L_{p}$ + $M_{occ}$}}}&{1.0}&{0.0}&{1.0}&{0.0}&{0.0}&{0.0}&{0.0}&{0.0}&{0.0}&{0.0}&{0.01}&{0.0}&{0.0}&{0.0}\\
		\multicolumn{1}{l}{{\small $L_{p}^{bi}$ + $M_{occ}^{bi}$}}&1.0&1.0&1.0&1.0&0.0&0.0&0.0&0.0&0.0&0.0&0.01&0.01&0.0&0.0\\

		\multicolumn{1}{l}{{{\small $L_{p}$ + $M_{occ}$ + $L_{dsc}$}}}&{1.0}&{0.0}&{1.0}&{0.0}&{0.0}&{0.0}&{0.5}&{0.0}&{0.0}&{0.0}&{0.01}&{0.0}&{0.0}&{0.0}\\
		\multicolumn{1}{l}{{\small $L_{p}^{bi}$ + $M_{occ}^{bi}$ + $L_{dsc}^{bi}$}}&1.0&1.0&1.0&1.0&0.0&0.0&0.5&0.5&0.0&0.0&0.01&0.01&0.0&0.0\\

		\multicolumn{1}{l}{{{\small $L_{p}$ + $M_{occ}$ + $L_{dsc}$ + $W_{aw}$}}}&{1.0}&{0.0}&{1.0}&{0.0}&{1.0}&{0.0}&{0.5}&{0.0}&{0.0}&{0.0}&{0.01}&{0.01}&{0.0}&{0.0}\\
		\multicolumn{1}{l}{{\small $L_{p}^{bi}$ + $M_{occ}^{bi}$ + $L_{dsc}^{bi}$ + $W_{aw}^{bi}$}}&1.0&1.0&1.0&1.0&1.0&1.0&0.5&0.5&0.0&0.0&0.01&0.01&0.0&0.0\\

		\multicolumn{1}{l}{{{\small $L_{p}$ + $M_{occ}$ + $L_{dsc}$ + $W_{aw}$ + $L_{feat}$}}}&{1.0}&{0.0}&{1.0}&{0.0}&{1.0}&{0.0}&{0.5}&{0.0}&{0.05}&{0.0}&{0.01}&{0.0}&{0.001}&{0.0}\\
		\multicolumn{1}{l}{{\small $L_{p}^{bi}$ + $M_{occ}^{bi}$ + $L_{dsc}^{bi}$ + $W_{aw}^{bi}$ + $L_{feat}^{bi}$}}&1.0&1.0&1.0&1.0&1.0&1.0&0.5&0.5&0.05&0.05&0.01&0.01&0.001&0.001\\
		
		\bottomrule
	\end{tabular}
	\caption{Parameter Settings}\label{tab:param_set}
	
	\vspace{-10pt}
\end{table*}

\paragraph{DepthNet}
The U-Net architecture \cite{ronneberger2015u} with an encoder-decoder structure is adopted in our depth estimation network, which can extract both deep abstract feature information and local information. We use ResNet-50 \cite{he2016deep} without a fully-connected layer as our encoder and finally output the deepest feature maps with a $1/32$ resolution relative to the input image after five rounds of subsampling. The decoder contains five convolution blocks, each consisting of a $3 \times 3$ convolutional layer with reflection padding and an exponential linear unit (ELU) nonlinear layer, followed by an upsampling layer. The decoder outputs feature maps with the same resolution as the input image after five rounds of upsampling. The feature maps of the last three scales are fed to a $3\times 3$ convolutional layer with a sigmoid function for synthesizing multiscale images and are employed for estimating the corresponding depth maps via a $3\times 3$ convolutional layer followed by a sigmoid function. The feature map of the maximum resolution extracted from the encoder is used for the feature perception loss. Finally, the predicted depths are constrained using $1/(\alpha *x+\beta)$ with $\alpha =10$ and $\beta =0.01$, following previous work \cite{zhou2017unsupervised}.

\paragraph{CameraNet}

CameraNet takes the image sequences concatenated from the target and reference frames along the channel dimension as input and outputs the relative camera poses between adjacent frames. For fairness, we use a similar network architecture as in \cite{zhou2017unsupervised} for our CameraNet, which consists of seven convolutional layers with stride 2, whose output is then fed to  a $1\times 1$ convolutional layer with $6*N_{ref}$ output channels. Finally, we use global average pooling to aggregate the predictions at all spatial locations.
\vspace{-1pt}
\subsection{Training Details}

The proposed learning framework was implemented using the PyTorch Library \cite{paszke2017automatic}. DepthNet and CameraNet were coupled by the loss function and trained jointly with a batch size of 2 and the learning rate of $10^{-4}$ using the AdamW \cite{loshchilov2017decoupled} optimizer; during the testing phase, however, each model could be used separately. The corresponding lambda parameters in our experiments were shown in Tab. \ref{tab:param_set}. 

We preprocessed the training set using random scaling, cropping and horizontal flipping. Then, the data to be used as input to the model were processed into the form of a tensor with a height of 256 and a width of 832. During training, following Ranjan et al. \cite{ranjan2019competitive}, five consecutive video frames were used as a training sample for model optimization, where the third image was regarded as the target image to calculate the losses with respect to the other four images, and the roles were then inverted to make the most of the limited available data. The model was trained for 150 epochs and validated in each epoch.

\vspace{-10pt}

\subsection{Performance Metrics}

\paragraph{Monocular Depth Estimation}

For depth evaluation, standard metrics from previous related work \cite{garg2016unsupervised,zhou2017unsupervised,eigen2014depth} were used, as shown in formula \eqref{eq_25}:

\begin{subequations}
	\label{eq_25}
	\begin{equation}
	\begin{aligned}
	\label{eq_25_1}
	AbsRel:\frac{1}{|D|}\sum_{d\in D}\frac{\|d^*-d\|}{d^*}
	\end{aligned}
	\end{equation}
	\begin{equation}
	\begin{aligned}
	\label{eq_25_2}
	RMSE:\sqrt{\frac{1}{|D|}\sum_{d\in D}\|d^*-d\|^2}
	\end{aligned}
	\end{equation}
	\begin{equation}
	\begin{aligned}
	\label{eq_25_3}
	SqRel:\frac{1}{|D|}\sum_{d \in D}\frac{\|d^*-d\|^2}{d^*}
	\end{aligned}
	\end{equation}
	\begin{equation}
	\begin{aligned}
	\label{eq_25_4}
	RMSElog:\sqrt{\frac{1}{|D|}\sum_{d\in D}\|logd^*-logd\|^2}
	\end{aligned}
	\end{equation}
	\begin{equation}
	\begin{aligned}
	\label{eq_25_5}
	\delta_t:
	\frac{1}{|D|}|\{{d\in D| max(\frac{d^*}{d},\frac{d}{d^*})<1.25^t}\}| *100\% 
	\end{aligned}
	\end{equation}		
\end{subequations} 

where $ d^*$ and $d$ denote the ground truth depth and the predicted depth, respectively. During the evaluation, the depth was capped at 50 m and 80 m in our experiments. To match the median with the ground truth, we needed to multiply the estimated depth maps by the scale factor computed from formula \eqref{eq_24} following the method in \cite{zhou2017unsupervised} because the depth estimated from monocular videos using our method is defined only up to a scale factor.

\begin{equation} 
\label{eq_24}
scale=\frac{D_{median}^{gt}}{D_{median}^{pred}} 
\end{equation}

\paragraph{Camera Pose Estimation}

For camera pose estimation, we used the absolute trajectory error (ATE) \cite{mur2015orb} as the performance metric. In our experiments, we computed the ATE by employing five frame snippets and optimized the scale factor according to formula \eqref{eq_24} such that the predictions were best aligned with the ground truth.

\begin{table*}[htbp]\small 
	\setlength\tabcolsep{2pt}
	\centering
	\footnotesize	
	\begin{tabular}{lccccccccccccccccccc} 
		\toprule 
		\multicolumn{1}{l}{\multirow{2}*{\footnotesize Method}}&
		\multicolumn{1}{c}{\multirow{2}*{\footnotesize Data}}&
		\multicolumn{1}{c}{\multirow{2}*{\footnotesize Cap}}&
		\multicolumn{1}{c}{\multirow{2}*{\footnotesize Resolutions}}&
		\multicolumn{1}{c}{\multirow{2}*{\footnotesize DE}} &
		\multicolumn{1}{c}{\multirow{2}*{\footnotesize PE}} &
		\multicolumn{1}{c}{\multirow{2}*{\footnotesize N}} &
		\multicolumn{1}{c}{\multirow{2}*{\footnotesize Sup}} &
		\multicolumn{4}{c}{\footnotesize Error$\downarrow$}& &
		\multicolumn{3}{c}{\footnotesize Accuracy$\uparrow$}\\
		\multicolumn{8}{c}{}&\footnotesize AbsRel&\footnotesize SqRel&\footnotesize RMSE&\footnotesize RMSElog& &\footnotesize $\delta_1$ &\footnotesize $ \delta_2$&\footnotesize $\delta_3$&	\\		
		\midrule		
		
		

		\multicolumn{1}{l}{{Guizilini et al.} \cite{guizilini20203d}}&K&80&192$\times$640 &PackNet&{PN7}&{1}&{{\footnotesize M+V}}&0.111&0.829&4.788&0.199& &0.864&0.954&0.980&\\

		\multicolumn{1}{l}{{Guizilini et al.} \cite{guizilini20203d}}&K+CS&80&192$\times$640 &PackNet&{PN7}&{1}&{{\footnotesize M+V}}&0.108&0.803&4.642&0.195& &0.875&0.958&0.980&\\
		
		\multicolumn{1}{l}{Luo et al.\cite{luo2019every}}&K&-&256$\times$832 &{VGG}&{PN7}&{1}&{M+PWCNet}&0.141&1.029&5.350&0.216& &0.816&0.941&0.976&\\
		
		
		\multicolumn{1}{l}{Wang et al.\cite{wang2020unsupervised}}&K&80&256$\times$832 &{RN50}&{PN7}&{1}&{{\footnotesize M+PWCNet+MaskNet}}&0.140&1.068&5.255&0.217& &0.827&0.943&0.977&\\

		\multicolumn{1}{l}{Ranjan et al.\cite{ranjan2019competitive}}&K+CS&80&256$\times$832 &{DRN}&{PN7}&{1}&{{\footnotesize M+PWCNet+MaskNet}}&0.139&1.032&5.199&0.213& &0.827&0.943&0.977&\\

		\multicolumn{1}{l}{Wang et al.\cite{wang2020unsupervised}}&K+CS&80&256$\times$832 &{{\footnotesize RN50}}&{{\footnotesize PN7}}&1&{{\footnotesize M+PWCNet+MaskNet}}&0.132&0.986&5.173&0.212& &0.835&0.945&0.977&\\
		
		\multicolumn{1}{l}{Zhao et al.\cite{zhao2020towards}}&K&-&256$\times$832 &{RN18}&-&-&{M+PWCNet}&0.130&0.893&5.062&0.205& &0.832&0.949&0.981&\\

		\multicolumn{1}{l}{{Bian et al.} \cite{bian2021unsupervised}}&{K}&{80}&{256$\times$832} &{RN50}&{RN18}&{2}&{M+ORBSLAM2}&{0.114}&{0.813}&{4.706}&{0.191}& &{0.873}&{0.960}&{0.982}&\\

		\multicolumn{1}{l}{{Shu et al.} \cite{shu2020feature}}&{K}&{80}&{320$\times$1024} &{RN50}&{RN18}&{1}&{M+FeatureNet}&{0.104}&{0.729}&{4.481}&{0.179}& &{\textit{0.893}}&{0.965}&{0.984}&\\
		
		\multicolumn{1}{l}{{Ma et al.} \cite{ma2022towards}}&{K}&{80}&{320$\times$1024}&{RN50}&{RN18}&{1}&{M+Semantic}&{\textit{0.099}}&{\textbf{0.624}}&{\textbf{4.165}}&{\textit{0.171}}& &{\textbf{0.902}}&{\textbf{0.969}}&{\textbf{0.986}}&\\

		\multicolumn{1}{l}{{Petrovai et al.} \cite{petrovai2022exploiting}}&{K}&{80}&{320$\times$1024}&{RN50}&{RN18}&{1}&{M+Pseudo}&{\textbf{0.098}}&{\textit{0.674}}&{\textit{4.187}}&{\textbf{0.170}}& &{\textbf{0.902}}&{\textit{0.968}}&{\textit{0.985}}&\\
	
		\multicolumn{1}{l}{{Klingner et al.} \cite{klingner2020self}}&{K+CS}&{80}&{384$\times$1280} &{RN18}&{RN18}&{1}&{M+Semantic}&{0.107}&{0.768}&{4.468}&{0.186}& &{0.891}&{0.963}&{0.982}&\\

		\multicolumn{1}{l}{{Guizilini et al.} \cite{guizilini2020semantically}}&{K}&{80}&{384$\times$1280}&{PackNet}&{PN7}&{1}&{M+Semantic}&{0.100}&{0.761}&{4.270}&{0.175}& &{\textbf{0.902}}&{0.965}&{0.982}&\\

		\midrule



		\multicolumn{1}{l}{Godard et al.\cite{godard2019digging}}&K&80&192$\times$640 &{RN18}&{RN18}&{1}&{M}&0.132&1.044&5.142&0.210& &0.845&0.948&0.977&\\
		
		\multicolumn{1}{l}{Godard et al.$^\ddagger$\cite{godard2019digging}}&K&80&192$\times$640 &{RN50}&{RN50}&{1}&{M}&0.131&1.020&5.060&0.206& &0.849&0.951&0.979&\\
		
		\multicolumn{1}{l}{{Guizilini et al.} \cite{guizilini20203d}}&K+IN&80&192$\times$640 &RN50&{PN7*}&{1}&{{\footnotesize M}}&0.117&0.900&4.826&0.196& &0.873&-&-&\\
		
		\multicolumn{1}{l}{Godard et al.\cite{godard2019digging}}&K+IN&80&192$\times$640 &{RN18}&{RN18}&{1}&{M}&0.115&0.903&4.863&0.193& &0.877&0.959&0.981&\\

		\multicolumn{1}{l}{{Guizilini et al.} \cite{guizilini20203d}}&K&80&192$\times$640 &PackNet&{PN7*}&{1}&{{\footnotesize M}}&0.111&0.785&4.601&0.189& &0.878&0.960&0.982&\\

		\multicolumn{1}{l}{Godard et al.$^\ddagger$\cite{godard2019digging}}&K+IN&80&192$\times$640 &{RN50}&{RN50}&{1}&{M}&0.110&0.835&4.644&0.187& &0.883&0.962&0.982&\\

		\multicolumn{1}{l}{{Guizilini et al.} \cite{guizilini20203d}}&K+CS&80&192$\times$640 &PackNet&{PN7*}&{1}&{{\footnotesize M}}&0.108&\textit{0.727}&4.426&0.184& &0.885&0.963&\textit{0.983}&\\
		
		\multicolumn{1}{l}{{He et al.} \cite{he2022ra}}&{K+IN}&{80}&{192$\times$640}&{HRNet}&{RN18}&{1}&{M}&{\textbf{0.096}}&{\textbf{0.632}}&{\textbf{4.216}}&{\textbf{0.171}}& &{\textbf{0.903}}&{\textbf{0.968}}&{\textbf{0.985}}&\\

		
		\multicolumn{1}{l}{{Jia et al.}\cite{jia2021self}}&{K}&{80}&{256$\times$832} &{RN18}&{PN7}&{1}&{M}&{0.136}&{0.895}&{4.834}&{0.199}& &{0.832}&{0.950}&{0.982}&\\

		\multicolumn{1}{l}{Bian et al.\cite{bian2019unsupervised}}&K+CS&80&256$\times$832 &{DRN}&{PN7}&{2}&{M}&0.128&1.047&5.234&0.208& &0.846&0.947&0.976&\\

		\multicolumn{1}{l}{{Gordon et al.} \cite{gordon2019depth}}&{K}&{80}&-&{RN18}&{RN18}&{2}&{M}&{0.128}&{0.959}&{5.23}&{0.212}& &{0.845}&{0.947}&{0.976}&\\

		\multicolumn{1}{l}{{Gordon et al.} \cite{gordon2019depth}}&{K+CS}&{80}&-&{RN18}&{RN18}&{2}&{M}&{0.124}&{0.930}&{5.12}&{0.206}& &{0.851}&{0.950}&{0.978}&\\

		\multicolumn{1}{l}{{Godard et al.}\cite{godard2019digging}}&{K+IN}&{80}&{320$\times$1024} &{RN18}&{RN18}&{1}&{M}&{0.115}&{0.882}&{4.701}&{0.190}& &{0.879}&{0.961}&{0.982}&\\

		\multicolumn{1}{l}{{Guizilini et al.} \cite{guizilini20203d}}&{K}&{80}&{384$\times$1280} &{PackNet}&{PN7*}&{1}&{M}&{0.107}&{0.802}&{4.538}&{0.186}& &{0.889}&{0.962}&{0.981}&\\

		\multicolumn{1}{l}{{Guizilini et al.} \cite{guizilini20203d}}&{K+CS}&{80}&{384$\times$1280} &{PackNet}&{PN7*}&{1}&{M}&{\textit{0.104}}&{0.758}&{\textit{4.386}}&{\textit{0.182}}& &{\textit{0.895}}&{\textit{0.964}}&{0.982}&\\

		\multicolumn{1}{l}{\textbf{Ours}}&K&80&256$\times$832&{RN50}&{PN7}&{1}&{M}&0.120&0.947&4.941&0.197& &0.863&0.957&0.981&\\

		\multicolumn{1}{l}{{\textbf{Ours}}}&K+CS&80&256$\times$832&{RN50}&{PN7}&{1}&{M}&0.110&0.847&4.654&0.189& &0.882&0.960&0.981&\\

		\multicolumn{1}{l}{{\textbf{Ours}}}&{K}&{80}&{384$\times$1280}&{RN50}&{PN7}&{1}&{M}&{0.110}&{0.829}&{4.614}&{0.185}& &{0.880}&{0.962}&{\textit{0.983}}&\\

		\multicolumn{1}{l}{{\textbf{Ours}}}&{K+CS}&{80}&{384$\times$1280}&{RN50}&{PN7}&{1}&{M}&{\textit{0.104}}&{0.798}&{4.501}&{0.184}& &{0.889}&{0.961}&{0.982}&\\

		\midrule

		
		

		\multicolumn{1}{l}{Luo et al.\cite{luo2019every}}&K&-&256$\times$832 &{VGG}&{PN7}&{1}&{M+PWCNet}&{0.141}&1.029&5.350&{0.216}& &{0.816}&{0.941}&0.976&\\
		
		\multicolumn{1}{l}{Zhao et al.\cite{zhao2020towards}}&K&-&256$\times$832 &{RN18}&-&-&{M+PWCNet}&{0.130}&{0.893}&5.062&{0.205}& &{0.832}&{0.949}&{0.981}&\\
		

		\multicolumn{1}{l}{\textbf{Ours}}&K&50&256$\times$832&{RN50}&{PN7}&{1}&{M}&0.116&0.817&4.025&0.188& &0.876&0.962&\textit{0.983}&\\

		\multicolumn{1}{l}{\textbf{Ours }}&K+CS&50&256$\times$832&{RN50}&{PN7}&{1}&{M}&\textit{0.105}&\textit{0.649}&\textit{3.571}&0.179& &\textit{0.895}&0.964&{\textit{0.983}}&\\

		\multicolumn{1}{l}{{\textbf{Ours}}}&{K}&{50}&{384$\times$1280}&{RN50}&{PN7}&{1}&{M}&{0.106}&{0.716}&{3.745}&{\textit{0.177}}& &{0.892}&{\textbf{0.966}}&{\textbf{0.984}}&\\

		\multicolumn{1}{l}{{\textbf{Ours }}}&{K+CS}&{50}&{384$\times$1280}&{RN50}&{PN7}&{1}&{M}&{\textbf{0.099}}&{\textbf{0.599}}&{\textbf{3.418}}&{\textbf{0.173}}& &{\textbf{0.900}}&{\textit{0.965}}&{\textbf{0.984}}&\\

		\bottomrule 
	\end{tabular}	
	\caption{Comparison of performance for monocular depth estimation on the KITTI dataset. `-' indicates that the Cap parameter is not specified in the corresponding paper.
		K denotes that our models were trained only on KITTI, and CS/IN+K means that the models were fine-tuned on KITTI after pretraining on the Cityscapes/ImageNet dataset. The best performance in each column is highlighted in bold and the second best is highlighted in italics. `DE/PE' refers to the backbone of the used encoder in depth/pose estimation network, `N' is the number of times that a pair of images need to be reasoned about, `M' refers to methods that only monocular(M) images are used to train network, `RN50/RN18/VGG/HRNet' refers to the corresponding encoder based on ResNet50 \cite{he2016deep}/ ResNet18 \cite{he2016deep}/ VGGNet \cite{simonyan2014very}/  HRNet\cite{wang2020deep}, `DRN' refers to the designed depth estimation network in  \cite{ranjan2019competitive}, `PN7' refers to a simple pose estimation network consisting of seven layers convolution being designed in \cite{zhou2017unsupervised}. `PackNet' is a more advanced and more complex network architecture \cite{guizilini20203d} compared with ResNet50 \cite {he2016deep}. `PWCNet' \cite{sun2018pwc}/ `FeatureNet' (based on ResNet50 \cite {he2016deep}) are the corresponding additional networks required for jointly optimizing the depth estimation model. `ORBSLAM2' refers to a method that is integrated with ORBSLAM2 \cite{mur2017orb} for optimizing the predicted depths and poses. `Semantic' refers to methods that require an additional network (e.g. Feature Pyramid Network \cite {lin2017feature} with ResNet \cite {he2016deep}) and semantic label to guide depth estimation.  $^\ddagger$ indicates the results, which are derived from github's latest weight. `*'  indicates that Group Normalization is used after each convolution layer in PN7.  }\label{tab:depth_compared_previous_method_80m_50m}
	\vspace{-6pt}
\end{table*}

\begin{table*}[htbp]
	\setlength\tabcolsep{2pt}
	\centering	 
	\begin{tabular}{lccccccccccccccccccc} 
		\toprule 
		
		\multicolumn{1}{l}{\multirow{2}*{\footnotesize Method}}&
		\multicolumn{1}{c}{\multirow{2}*{\footnotesize Data}}&		
		\multicolumn{1}{c}{\multirow{2}*{\footnotesize Resolutions}}&		
		\multicolumn{4}{c}{ Error$\downarrow$}& &
		\multicolumn{3}{c}{\footnotesize Accuracy$\uparrow$}\\
		\multicolumn{3}{c}{}&\footnotesize AbsRel&\footnotesize SqRel&\footnotesize RMSE&\footnotesize RMSElog& &\footnotesize $\delta_1$ &\footnotesize $ \delta_2$&\footnotesize $\delta_3$&	\\		
		\midrule
		
		\multicolumn{1}{l}{Zhou et al.\cite{zhou2017unsupervised}}&K+CS&{128$\times$416} &{0.176}&{1.532}&{6.129}&{0.244}& &{0.758}&{0.921}&{0.971}&\\

		\multicolumn{1}{l}{Ranjan et al.\cite{ranjan2019competitive}}&K+CS&{256$\times$832} &{0.1049}&{0.6569}&{4.3128}&{0.1572}& &{0.8869}&{0.9721}&{0.9914}&\\
	
		\multicolumn{1}{l}{Bian et al.\cite{bian2019unsupervised}}&K+CS&{256$\times$832} &{0.0984}&{0.6495}&{4.3975}&{0.1526}& &{0.8917}&{0.9717}&{0.9906}&\\
		\multicolumn{1}{l}{{Godard et al.}\cite{godard2019digging}}&K+IN&{192$\times$640} &{0.090}&{0.545}&{3.942}&{0.137}& &{0.914}&{0.983}&{0.995}&\\
		
		\multicolumn{1}{l}{{Guizilini et al.} \cite{guizilini20203d}}&K+CS&{192$\times$640} &{\textit{0.078}}&{\textbf{0.420}}&{\textbf{3.485}}&{\textit{0.121}}& &{\textit{0.931}}&{\textbf{0.986}}&{\textbf{0.996}}&\\
		\multicolumn{1}{l}{\textbf{Ours}}&K+CS&256$\times$832&\textbf{0.0766}&\textit{0.4249}&\textit{3.5371}&\textbf{0.1207}& &\textbf{0.9336}&\textit{0.9849}&\textit{0.9959}&\\

		\hline
		\multicolumn{1}{l}{{Godard et al.}\cite{godard2019digging}}&K+IN&{320$\times$1024} &{0.0858}&{0.4619}&{3.5768}&{0.1270}& &{0.9242}&{0.9861}&{0.9962}&\\
		\multicolumn{1}{l}{{Guizilini et al.} \cite{guizilini20203d}}&K+CS&{384$\times$1280} &\textbf{0.071}&{\textbf{0.359}}&{\textbf{3.153}}&{\textbf{0.109}}& &{\textbf{0.944}}&{\textbf{0.990}}&{\textbf{0.997}}&\\
		
		\multicolumn{1}{l}{\textbf{Ours}}&K+CS&384$\times$1280&\textit{0.0723}&\textit{0.3767}&\textit{3.3584}&\textit{0.1147}& &\textit{0.9390}&\textit{0.9872}&\textit{0.9964}&\\

		\bottomrule 
	\end{tabular}	
	\caption{Comparison of performance for monocular depth estimation on the improved KITTI dataset \cite{uhrig2017sparsity}.  }\label{tab:depth_compared_on_improved_kitti}
	\vspace{-6pt}
\end{table*}

\begin{table*}[htbp]
	\setlength\tabcolsep{2pt}
	\centering

	\begin{tabular}{lccccccccccccccccccc} 
		\toprule 
		\multicolumn{1}{l}{\multirow{2}*{\footnotesize Method}}&
		\multicolumn{1}{l}{\multirow{2}*{\footnotesize Data}}&		
		\multicolumn{1}{c}{\multirow{2}*{\footnotesize Resolutions}}&		
		\multicolumn{4}{c}{ Error$\downarrow$}& &
		\multicolumn{3}{c}{\footnotesize Accuracy$\uparrow$}\\
		\multicolumn{3}{c}{}&\footnotesize AbsRel&\footnotesize SqRel&\footnotesize RMSE&\footnotesize RMSElog& &\footnotesize $\delta_1$ &\footnotesize $ \delta_2$&\footnotesize $\delta_3$&	\\ 	 
		
		\midrule
		
		\multicolumn{1}{l}{Zhou et al.\cite{zhou2017unsupervised}}&K+CS&{128$\times$416} &{0.1926}&{1.3625}&{6.1366}&{0.2767}& &{0.7094}&{0.8999}&{0.9587}&\\
		
		\multicolumn{1}{l}{Ranjan et al.\cite{ranjan2019competitive}}&K+CS&{256$\times$832} &{0.1475}&{1.0216}&{4.9665}&{0.2265}& &{0.8197}&{0.9392}&{0.9724}&\\
		
		\multicolumn{1}{l}{Bian et al.\cite{bian2019unsupervised}}&K+CS&{256$\times$832} &{0.1416}&{1.0717}&{5.1491}&{0.2299}& &{0.8281}&{0.9367}&{0.9683}&\\
		
		\multicolumn{1}{l}{{Godard et al.}\cite{godard2019digging}}&K+IN&{192$\times$640} &{0.1251}&{1.0205}&{4.8573}&{0.2137}& &{0.8685}&{0.9522}&{0.9757}&\\
		
		\multicolumn{1}{l}{{Guizilini et al.} \cite{guizilini20203d}}&K+CS&{192$\times$640} &{\textit{0.1209}}&{\textit{0.9012}}&{\textit{4.6110}}&{\textit{0.2079}}& &{\textit{0.8716}}&{\textit{0.9542}}&{\textit{0.9768}}&\\
		
		\multicolumn{1}{l}{\textbf{Ours}}&K+CS&{256$\times$832} &{\textbf{0.1145}}&{\textbf{0.8215}}&{\textbf{4.4561}}&{\textbf{0.2027}}& &{\textbf{0.8793}}&{\textbf{0.9569}}&{\textbf{0.9780}}&\\

		\hline
		
		\multicolumn{1}{l}{{Godard et al.}\cite{godard2019digging}}&K+IN&{320$\times$1024} &{0.1249}&{0.9541}&{\textit{4.5813}}&{0.2099}& &{0.8686}&{\textit{0.9553}}&{\textit{0.9766}}&\\

		\multicolumn{1}{l}{{Guizilini et al.} \cite{guizilini20203d}}&K+CS&{384$\times$1280} &{\textit{0.1159}}&{\textit{0.8936}}&{4.5912}&{\textit{0.2079}}& &{\textbf{0.8824}}&{0.9539}&{0.9757}&\\
	
		\multicolumn{1}{l}{\textbf{Ours}}&K+CS&{384$\times$1280} &{\textbf{0.1128}}&{\textbf{\textbf{0.7848}}}&{\textbf{4.3275}}&{\textbf{0.1994}}& &{\textit{0.8727}}&{\textbf{0.9565}}&{\textbf{0.9793}}&\\

		\bottomrule
	\end{tabular}	
	\caption{Comparison of performance for monocular depth estimation on the  282 images selected from 697 images in accordance with Eigen's testing split. (272 images with moving objects and 10 images in which most areas are textureless.)  }\label{tab:depth_compared_on_282_moving_object}
	\vspace{-6pt}
\end{table*}

\begin{table*}[htbp]
	
	\setlength\tabcolsep{2pt}
	\centering	
	\begin{tabular}{lccccccccccccccccccc} 
		\toprule 
		\multicolumn{1}{l}{\multirow{2}*{\footnotesize Method}}&		
		\multicolumn{1}{c}{\multirow{2}*{\footnotesize Resolutions}}&		
		\multicolumn{4}{c}{ Error$\downarrow$}& &
		\multicolumn{3}{c}{\footnotesize Accuracy$\uparrow$}\\
		\multicolumn{2}{c}{}&\footnotesize AbsRel&\footnotesize SqRel&\footnotesize RMSE&\footnotesize RMSElog& &\footnotesize $\delta_1$ &\footnotesize $ \delta_2$&\footnotesize $\delta_3$&	\\		
		\midrule
		\multicolumn{1}{l}{{Guizilini et al.} \cite{guizilini20203d}}&{384$\times$640} &{0.162}&{3.917}&{13.452}&{0.269}& &{0.823}&{-}&{-}&\\
		\multicolumn{1}{l}{\textbf{Ours}}&384$\times$640&\textbf{0.1210}&\textbf{1.2288}&\textbf{6.7214}&\textbf{0.1872}& &\textbf{0.8533}&\textbf{0.9575}&\textbf{0.9845}&\\		
		\bottomrule 
	\end{tabular}	
	\caption{Comparison of performance for monocular depth estimation on the DDAD dataset \cite{guizilini20203d}.  }\label{tab:depth_compared_on_ddad}
	\vspace{-6pt}
\end{table*}

\subsection{Comparison with State-of-the-Art Methods} 

\paragraph{Monocular Depth Estimation}

In Tab. \ref{tab:depth_compared_previous_method_80m_50m}, we compare the depth estimation results with those of current state-of-the-art self-supervised methods trained on the KITTI dataset and with the results of methods with parameters pretrained on the Cityscapes dataset and then fine-tuned on KITTI, which are taken from the corresponding published papers. Depths capped at 50 m and 80 m were used to evaluate the model performance.
The results show that the quality of the recovered depth map could be significantly improved by jointing learning different tasks (e.g., optical flow task \cite{yin2018geonet,zou2018df,wang2020unsupervised,ranjan2019competitive,luo2019every,zhao2020towards}, segmentation task \cite{ranjan2019competitive,wang2020unsupervised}, feature representation task \cite{shu2020feature}, semantic learning task \cite{klingner2020self,guizilini2020semantically} ) utilizing different networks. Besides, it can be also seen from previous work \cite{bian2019unsupervised,bian2021unsupervised,godard2019digging,guizilini20203d} that both the resolution of the input image and the adopted network architecture play an important role in dense depth estimation. We are more interested here in monocular methods where neither additional tasks are required nor the complexity of the model is increased. In the absence of additional auxiliary tasks and pretraining, our proposed method outperforms previous methods \cite{zhou2017unsupervised,bian2019unsupervised,li2020unsupervised,godard2019digging,gordon2019depth,jia2021self,godard2019digging}, except \cite{ guizilini20203d} where more complex network architecture is adopted. To be relatively fair, we also trained the network using the image with the same resolution as \cite{guizilini20203d} and using additional training data. It shows that competitive results can be obtained under the same resolution conditions compared with the methods \cite{guizilini20203d}, while the depth estimator in reference \cite{guizilini20203d} is four times as large as ours (see \cite{guizilini20203d}). We believe that the better network architecture for DepthNet (e.g \cite{guizilini20203d}) and CameraNet (e.g. \cite{godard2019digging}) is utilized, the greater improvements will be achieved in depth estimation. Moreover, although our approach introduces no additional information, it outperforms most previous methods \cite{wang2020unsupervised,ranjan2019competitive,zou2018df,yin2018geonet,luo2019every}. We attribute this to the fact that our objective function can provide a better optimization direction for the network.

In addition, to further verify depth estimation results, in Tab. \ref{tab:depth_compared_on_improved_kitti}, we also evaluate the model performance on the improved KITTI dataset \cite{uhrig2017sparsity}. It also shows that our proposed method outperforms previous methods \cite{zhou2017unsupervised,ranjan2019competitive,bian2019unsupervised,godard2019digging} and can also achieve competitive performance compared to the current state-of-the-art method \cite{guizilini20203d} without additional auxiliary tasks.

\begin{figure*}[htbp]
	\centering
	\tiny
	\rotatebox{90}{
		$\quad$$\quad$Ours$\ $ 
		$\quad$$\quad$$\quad$$\quad$Ours$\ $  
		$\quad$$\quad$$\quad$$\quad$Ours$\ $ 
		$\quad$$\quad$$\quad$$\quad$Ours$\ $		
		$\quad$$\quad$Guizilini\cite{guizilini20203d} $ \ $
		$\quad$Godard\cite{godard2019digging} $ \ $
		$\quad$Bian\cite{bian2019unsupervised} $ \ $ 
		$\quad$Ranjan\cite{ranjan2019competitive}$\ $	
		$\quad$$\quad$GT $\quad$$\quad$$\quad$$\quad$$\quad$$\quad$Input
	} 
	\rotatebox{90}{$\quad$$\quad$K+CS $\quad$$\quad$$\quad$$\quad$K+CS $\quad$$\quad$$\quad$$\quad$$\quad$K$\quad$$\quad$$\quad$$\quad$$\quad$ K $\quad$$\quad$$\quad$$\quad$K+CS $\quad$$\quad$$\quad$$\quad$$\quad$K+IN$\quad$$\quad$$\quad$$\quad$$\quad$$\quad$K$\quad$$\quad$$\quad$$\quad$$\quad$$\quad$K}
     \rotatebox{90}{
    	${\tiny 384\times1280}$
    	${\tiny 256\times832}$
    	${\tiny 384\times1280}$
    	${\tiny 256\times832}$
    	${\tiny 384\times1280}$
    	${\tiny 320\times1024}$
    	${\tiny 256\times832}$
    	$\quad$${\tiny 256\times832}$

    }
	\includegraphics[scale=0.13]{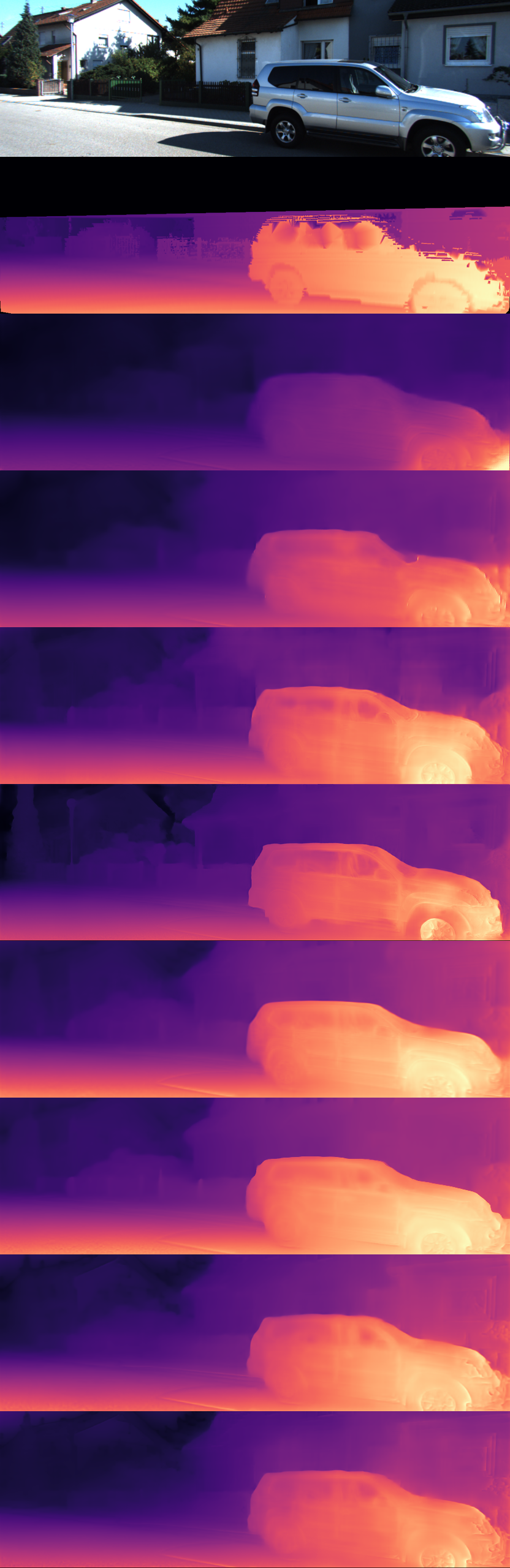}
	\includegraphics[scale=0.13]{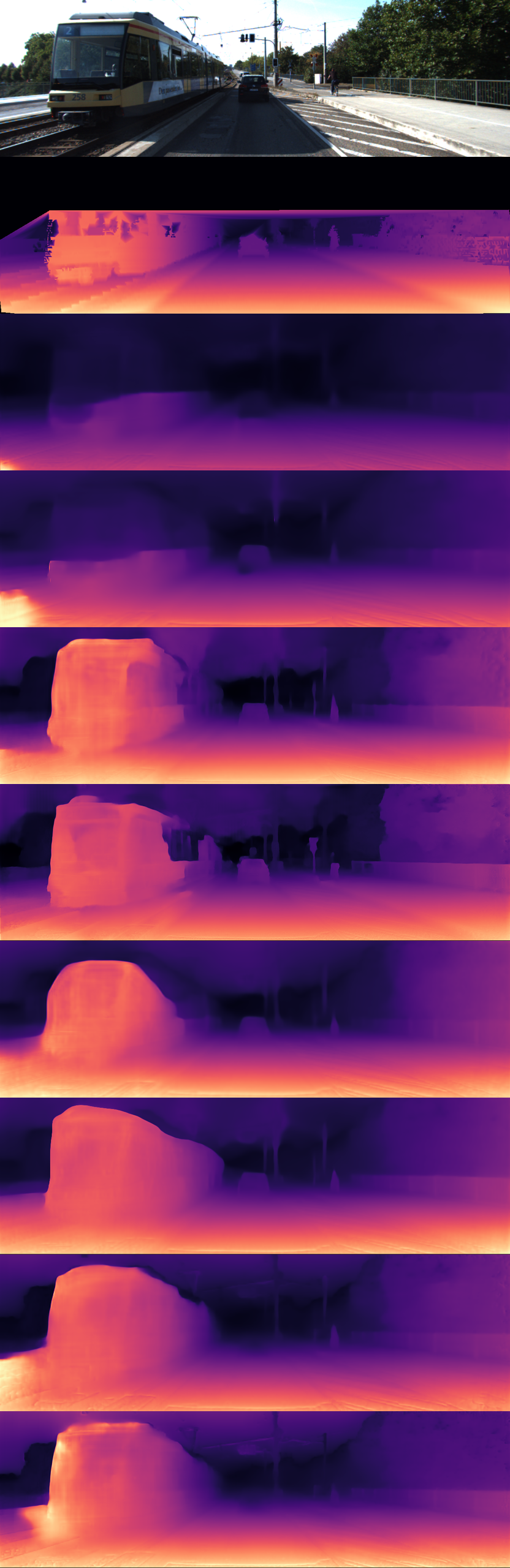}
	\includegraphics[scale=0.13]{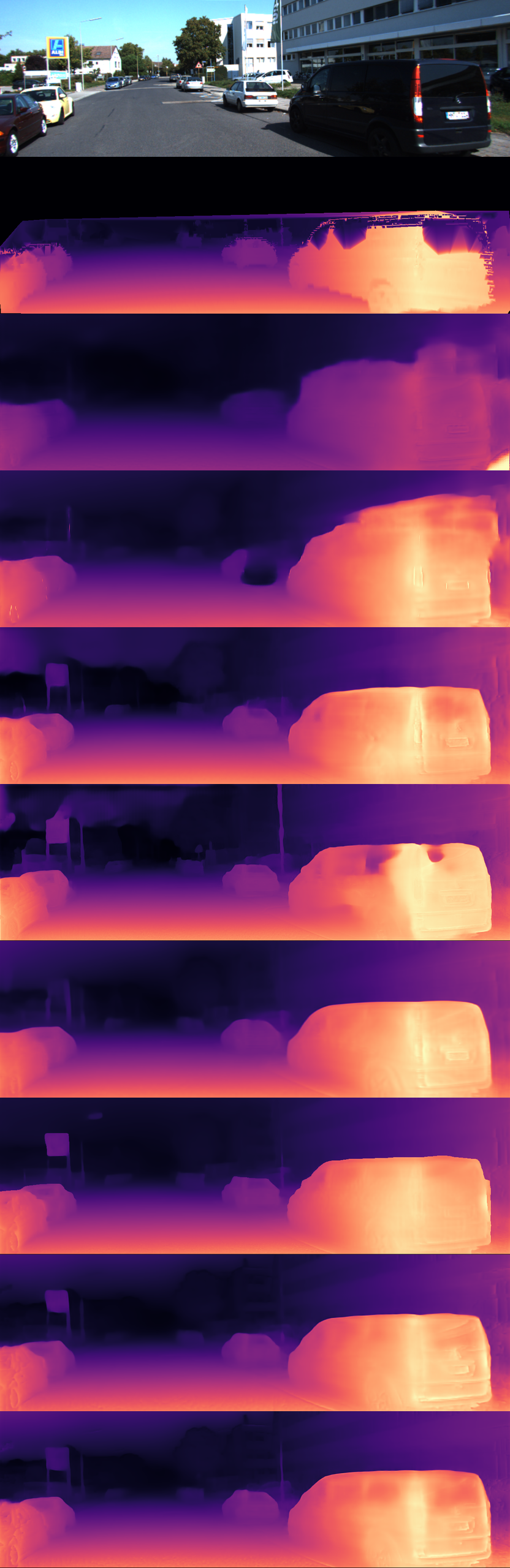}	\includegraphics[scale=0.13]{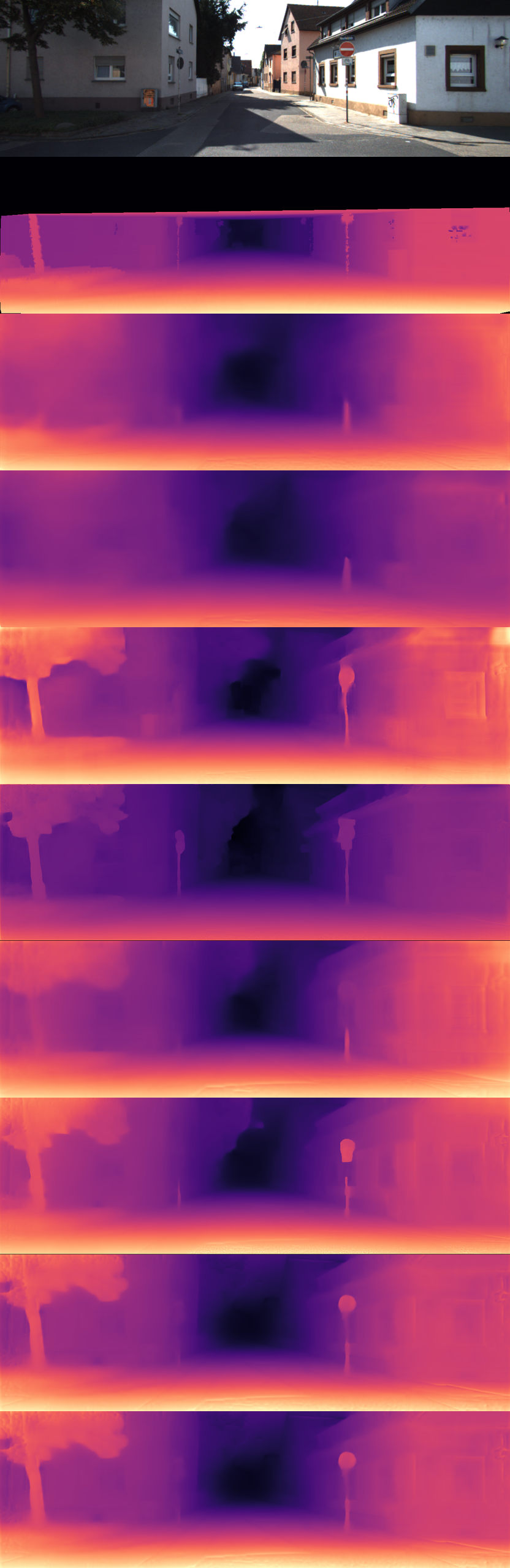}  \\
	
	$\qquad$	\\
	$\qquad$ \\
	
	\rotatebox{90}{
		$\quad$$\quad$Ours$\ $ 
		$\quad$$\quad$$\quad$$\quad$Ours$\ $  
		$\quad$$\quad$$\quad$$\quad$Ours$\ $ 
		$\quad$$\quad$$\quad$$\quad$Ours$\ $		
		$\quad$$\quad$Guizilini\cite{guizilini20203d} $ \ $
		$\quad$Godard\cite{godard2019digging} $ \ $
		$\quad$Bian\cite{bian2019unsupervised} $ \ $ 
		$\quad$Ranjan\cite{ranjan2019competitive}$\ $		
		$\quad$$\quad$GT $\quad$$\quad$$\quad$$\quad$$\quad$$\quad$Input
	} 
	\rotatebox{90}{$\quad$$\quad$K+CS $\quad$$\quad$$\quad$$\quad$K+CS $\quad$$\quad$$\quad$$\quad$$\quad$K$\quad$$\quad$$\quad$$\quad$$\quad$ K $\quad$$\quad$$\quad$$\quad$K+CS $\quad$$\quad$$\quad$$\quad$$\quad$K+IN$\quad$$\quad$$\quad$$\quad$$\quad$$\quad$K$\quad$$\quad$$\quad$$\quad$$\quad$$\quad$K}
	\rotatebox{90}{
		${\tiny 384\times1280}$
		${\tiny 256\times832}$
		${\tiny 384\times1280}$
		${\tiny 256\times832}$
		${\tiny 384\times1280}$
		${\tiny 320\times1024}$
		${\tiny 256\times832}$
		$\quad$${\tiny 256\times832}$

	}
	\includegraphics[scale=0.13]{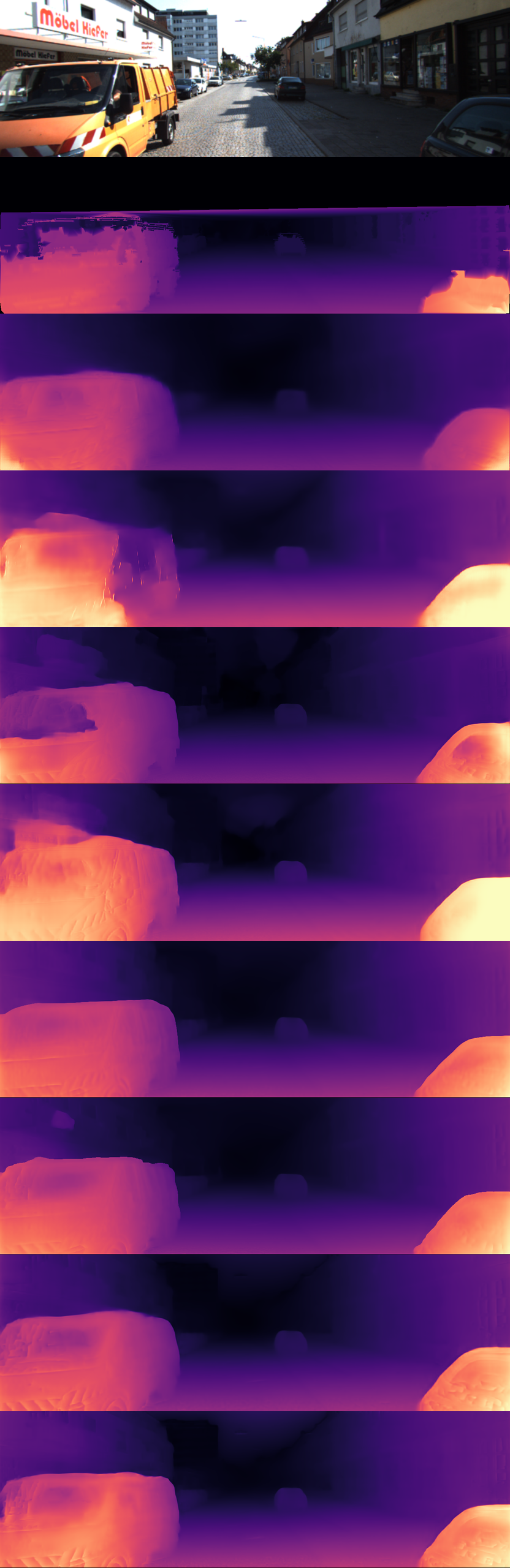}
	\includegraphics[scale=0.13]{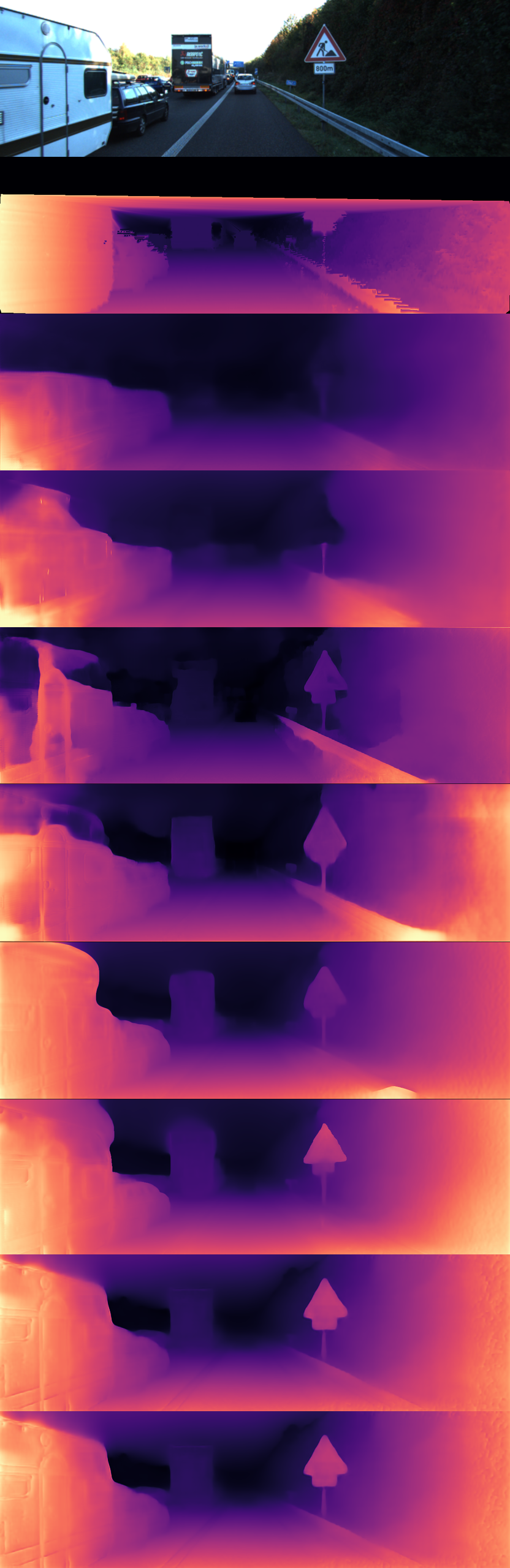}
	\includegraphics[scale=0.13]{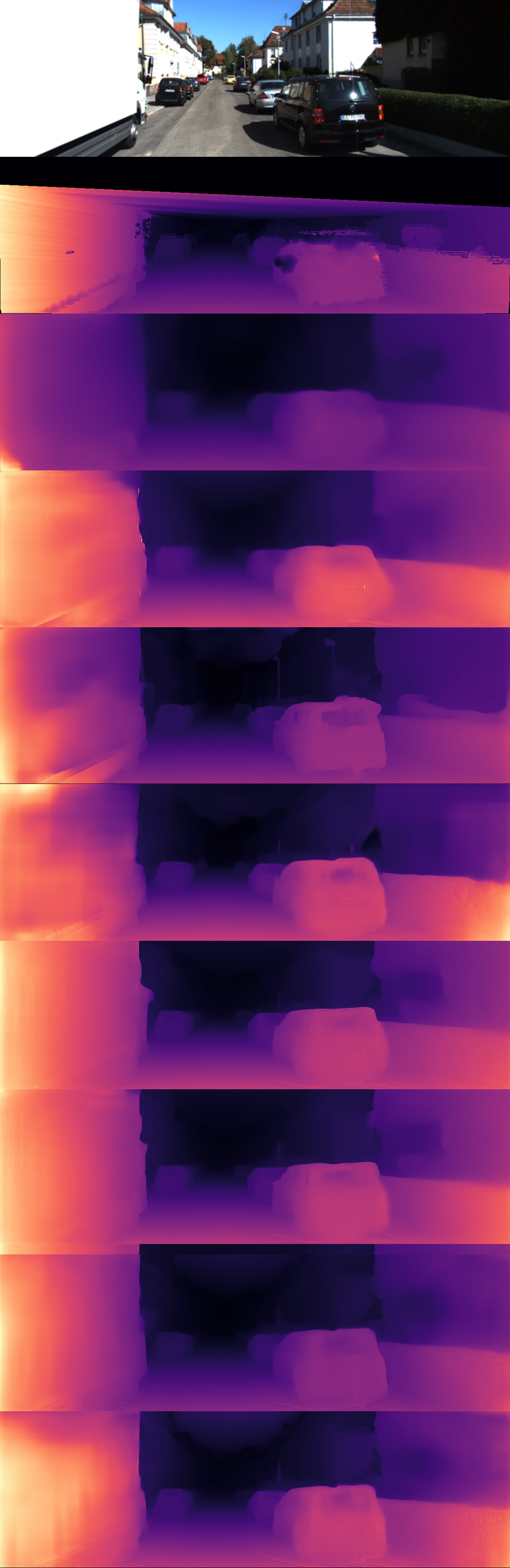}
	\includegraphics[scale=0.13]{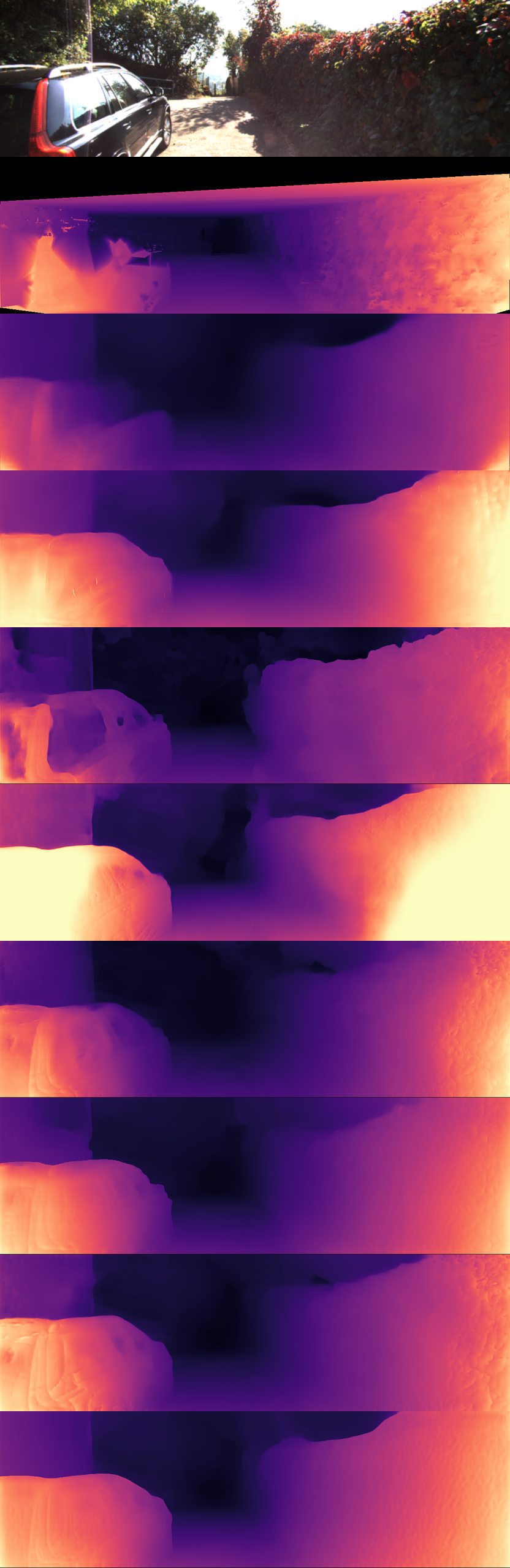}

	\caption{Qualitative comparison of example results of our proposed self-supervised monocular depth estimation method with those of previous state-of-the-art methods as estimated on the KITTI dataset. The ground truth maps were obtained from sparse laser data for visualization only. The brighter an area in a depth map is, the closer it is to the camera. 
	}\label{fig:compare_disp}%
	\vspace{-10pt}	
\end{figure*}

In order to further quantitatively observe the robustness of the algorithm to moving objects and textureless regions, we select images with moving objects and those in which most areas are textureless from 697 test frames in accordance with Eigen’s testing split, resulting in 282 test frames. Tab.  \ref{tab:depth_compared_on_282_moving_object} reports the quantitative results of different methods evaluated on these challenging scenarios. It shows that our proposed method is more robust to these scenarios than the previous methods \cite{zhou2017unsupervised,ranjan2019competitive,bian2019unsupervised,godard2019digging,guizilini20203d}. We suspect that this may be caused by the following reasons:
1) Auto-Mask scheme adopted in \cite{godard2019digging,guizilini20203d} only allows the network to ignore the contribution of objects, which move at the same velocity as the camera, to photometric loss. Nevertheless, this scheme is invalid when the moving object has a different translation speed from the camera. Our scheme based on the flow fields and depth structure could mitigate this impact.
2) The above methods are inefficient for large untextured areas due to not making full use of semantic and contextual information (e.g. the large white area in the seventh column in Fig. \ref{fig:compare_disp}), while our proposed bidirectional feature perception loss could force the network to focus on these information.
3) Our quantitative result is slightly lower than that in \cite{guizilini20203d} in 697 test frames, which should be due to the fact that more fine-grained information could be preserved by their proposed packing-unpacking blocks. 

Tab. \ref{tab:depth_compared_on_ddad} reports the quantitative results evaluated on the DDAD dataset \cite{guizilini20203d} which is a more realistic and challenging benchmark for depth estimation and contains more moving objects. It demonstrates that our proposed method outperforms prior work \cite{guizilini20203d} by a big margin, which also proves the above conjecture from the side.

\begin{table}[htb]\small
	\vspace{-2pt}
	
	\setlength\tabcolsep{2pt}
	\centering

	\begin{tabular}{lcccccc} 
		\toprule 
		\multicolumn{1}{l}{ }&
		\multicolumn{1}{l}{BackBone}& 
		\multicolumn{1}{c}{Param({\scriptsize  M})}& 		
		\multicolumn{1}{c}{InferT({ms})}& 
		\multicolumn{1}{c}{GPU({\scriptsize M})}\\

		\hline
		
		\multirow{3}{*}{DepthNet } &DRN&80.88&10.6&2389 \\ 
		
		&PackNet&128.29&49.3&3981\\

		&RN50(Ours)&32.52&12.1&2143\\
		\hline
	
		\multirow{3}{*}{CameraNet }&PN7*&1.59&1.9&1899\\
	
		&PN7(Ours)&1.59&1.5&1897\\

		&RN18&13.01&4.5&1999\\

		\bottomrule 
	\end{tabular}
	\caption{Comparison of offline infer time on the different backbones with batchsize=1. `*'  indicates that Group Normalization is used after each convolution layer in PN7. All results are evaluated on RTX 3090Ti with the same setting. The resolution is set to 256$\times$832. The time of DepthNet is averaged over 697 test frames in accordance with Eigen’s testing split, and the time of CameraNet is averaged over sequence 09 in the KITTI Odometry dataset. `InferT' indicates infer time.  }\label{tab:param_time_gpu_compare_offline}
	\vspace{-10pt}
\end{table}

\begin{table}[htb]\small
	\vspace{-2pt}
	
	\setlength\tabcolsep{2pt}
	\centering

	\begin{tabular}{lcccccc} 
		\toprule 
		\multicolumn{1}{l}{BackBone}&		
		\multicolumn{1}{c}{TrainT({ms})}& \multicolumn{1}{c}{GPU({\scriptsize M})}\\
		\hline
		
		PackNet+PN7*&1068.3&19817\\

		RN50+PN7(Ours)&307.9&10013\\	
	
		RN50+RN18&321.7&10645\\

		\bottomrule 
	\end{tabular}
	\caption{Comparison of train time on the different backbones with batchsize=2. `*'  indicates that Group Normalization is used after each convolution layer in PN7. All results are evaluated on RTX 3090Ti with the same setting. The resolution is set to 256$\times$832. The time is averaged 1000 iterations on the KITTI RAW dataset. }\label{tab:param_time_gpu_compare_train}
	\vspace{-10pt}
\end{table}

\begin{table}[htb]\small
	\vspace{-2pt}
	
	\setlength\tabcolsep{2pt}
	\centering

	\begin{tabular}{lcc} 
		\toprule 
		\multicolumn{1}{l}{Method}& \multicolumn{1}{c}{Seq. 09}& \multicolumn{1}{c}{Seq. 10}\\
		\hline 
		\multicolumn{1}{l}{ORB-SLAM (short)} & 0.064$\pm$0.141 &0.064$\pm$0.130 \\  
		\multicolumn{1}{l}{ORB-SLAM (full)}& 0.014$\pm$0.008&0.012$\pm$0.011 \\
		\multicolumn{1}{l}{Mean Odometry}& 0.032$\pm$0.026& 0.028$\pm$0.023 \\
		\multicolumn{1}{l}{Zhou et al. \cite{zhou2017unsupervised}}& 0.021$\pm$0.017& 0.020$\pm$0.015 \\
		\multicolumn{1}{l}{Zou et al. \cite{zou2018df}}& 0.017$\pm$0.007& 0.015$\pm$0.009\\

		\multicolumn{1}{l}{Bian et al.$^\ddagger$\cite{bian2019unsupervised}}& 0.016$\pm$0.007& 0.016$\pm$0.015\\
		\multicolumn{1}{l}{{{Godard et al.$^\ddagger$}\cite{godard2019digging}}}& 0.021$\pm$0.009& 0.014$\pm$0.010\\

		\multicolumn{1}{l}{Luo et al. \cite{luo2019every}}& 0.013$\pm$0.007& 0.012$\pm$0.008 \\
		\multicolumn{1}{l}{Mahjourian et al. \cite{mahjourian2018unsupervised}}& 0.013$\pm$0.010& 0.012$\pm$0.011 \\
		\multicolumn{1}{l}{Ranjan et al. \cite{ranjan2019competitive}}& 0.012$\pm$0.007& 0.012$\pm$0.008 \\
		\multicolumn{1}{l}{\textbf{Ours}}& \textit{0.0120$\pm$0.0068}& \textit{0.0118$\pm$0.0081} \\

		\multicolumn{1}{l}{\textbf{{Ours\textdagger}}}& {\textbf{0.0084$\pm$0.0047}}& {\textbf{0.0084$\pm$0.0064}} \\
		 
		\bottomrule
	\end{tabular}
	\caption{Comparison of performance for camera pose estimation. The results were tested on sequences 09 and 10 in the KITTI Odometry dataset. For ORB-SLAM (short), only five frames were taken as input, whereas for ORB-SLAM (full), the entire sequence was taken as input \cite{mur2015orb}. \textdagger \,  indicates that three consecutive frames with a height of 384 and a width of 1280 were used as a training sample for depth and camera pose estimation experiments. $^\ddagger$ indicates the results, which are derived from github's latest weight.}\label{tab:camera_pose_previous_method}
	\vspace{-10pt}
\end{table}

In Tab. \ref{tab:param_time_gpu_compare_offline} and \ref{tab:param_time_gpu_compare_train}, we compare the resource consumption of different models. As you can see from Tab. \ref{tab:param_time_gpu_compare_offline}, the current state-of-the-art self-supervised monocular method \cite{guizilini20203d} requires twice as much memory and takes four times as much reasoning time as ours under the same conditions during inferring.  Furthermore, the previous work \cite{guizilini20203d} takes up more memory and takes longer training times during training as shown in Tab. \ref{tab:param_time_gpu_compare_train}.

The qualitative results shown in Fig. \ref{fig:compare_disp} also prove that the proposed method outperforms the existing state-of-the-art self-supervised methods \cite{ranjan2019competitive,bian2019unsupervised,godard2019digging,guizilini20203d} in the scene consisting of moving objects, occlusions, and textureless regions. More concretely, compared with the existing methods \cite{ranjan2019competitive,bian2019unsupervised,godard2019digging,guizilini20203d}, our method can estimate sharper and smoother scene depths, especially in areas where there are moving objects, occlusions, or textureless regions. For example, in the example in the first column of Fig. \ref{fig:compare_disp}, the brightness of the car in the depth map estimated by our method is closer to the brightness at the corresponding position in the ground truth depth map, while the brightnesses of this car as estimated by the previous methods \cite{ranjan2019competitive,bian2019unsupervised} are darker than that in the ground truth depth map. The windshield of the train in the second column and the large white areas in seventh column, which are a textureless region, are or close to black or not smooth enough in the depth maps predicted by the previous methods \cite{ranjan2019competitive,bian2019unsupervised,godard2019digging,guizilini20203d}. However, the brightness at the same position in the depth map estimated by our method is very similar to that in the ground truth depth map. In other words, our method can work well even in textureless regions and accurately predict the depth of these regions, while the previous methods \cite{ranjan2019competitive,bian2019unsupervised,godard2019digging,guizilini20203d} fail to correctly estimate the depth of such regions. In the third column, although the method \cite{bian2019unsupervised} accurately estimates the depth of the black car in the image than the methods of \cite{ranjan2019competitive} do, it has difficulty accurately predicting the depth in the intersection regions between the black car and the image background because of occlusion effects from the black car. The depth of the moving objects in the fifth, sixth and eighth columns is incorrectly estimated by the previous methods \cite{ranjan2019competitive,bian2019unsupervised,godard2019digging,guizilini20203d}.

\begin{figure*}[!t]
	\centering
	\tiny
	\rotatebox{90}{
		$\quad$$\quad$Ours$\ $ 
		$\quad$$\quad$$\quad$$\quad$Ours$\ $  
		$\quad$$\quad$$\quad$$\quad$Ours$\ $ 
		$\quad$$\quad$$\quad$$\quad$Ours$\ $		
		$\quad$$\quad$Guizilini\cite{guizilini20203d} $ \ $
		$\quad$Godard\cite{godard2019digging} $ \ $
		$\quad$Bian\cite{bian2019unsupervised} $ \ $ 
		$\quad$Ranjan\cite{ranjan2019competitive}$\ $	
	} 
	\rotatebox{90}{$\quad$$\quad$K+CS $\quad$$\quad$$\quad$$\quad$K+CS $\quad$$\quad$$\quad$$\quad$$\quad$K$\quad$$\quad$$\quad$$\quad$$\quad$ K $\quad$$\quad$$\quad$$\quad$K+CS $\quad$$\quad$$\quad$$\quad$$\quad$K+IN$\quad$$\quad$$\quad$$\quad$$\quad$$\quad$K$\quad$$\quad$$\quad$$\quad$$\quad$$\quad$K}
	\rotatebox{90}{
		${\tiny 384\times1280}$
		${\tiny 256\times832}$
		${\tiny 384\times1280}$
		${\tiny 256\times832}$
		${\tiny 384\times1280}$
		${\tiny 320\times1024}$
		${\tiny 256\times832}$
		$\quad$${\tiny 256\times832}$

	}
	\includegraphics[scale=0.13]{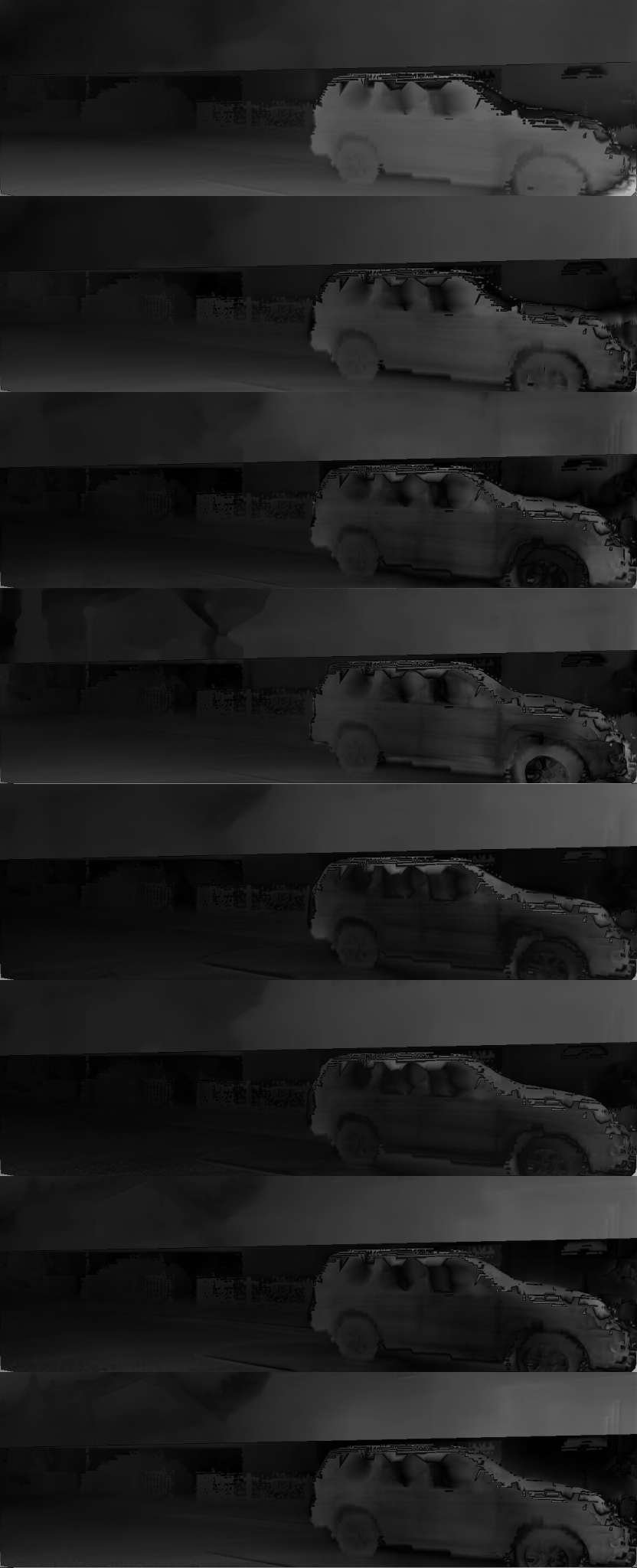}
	\includegraphics[scale=0.13]{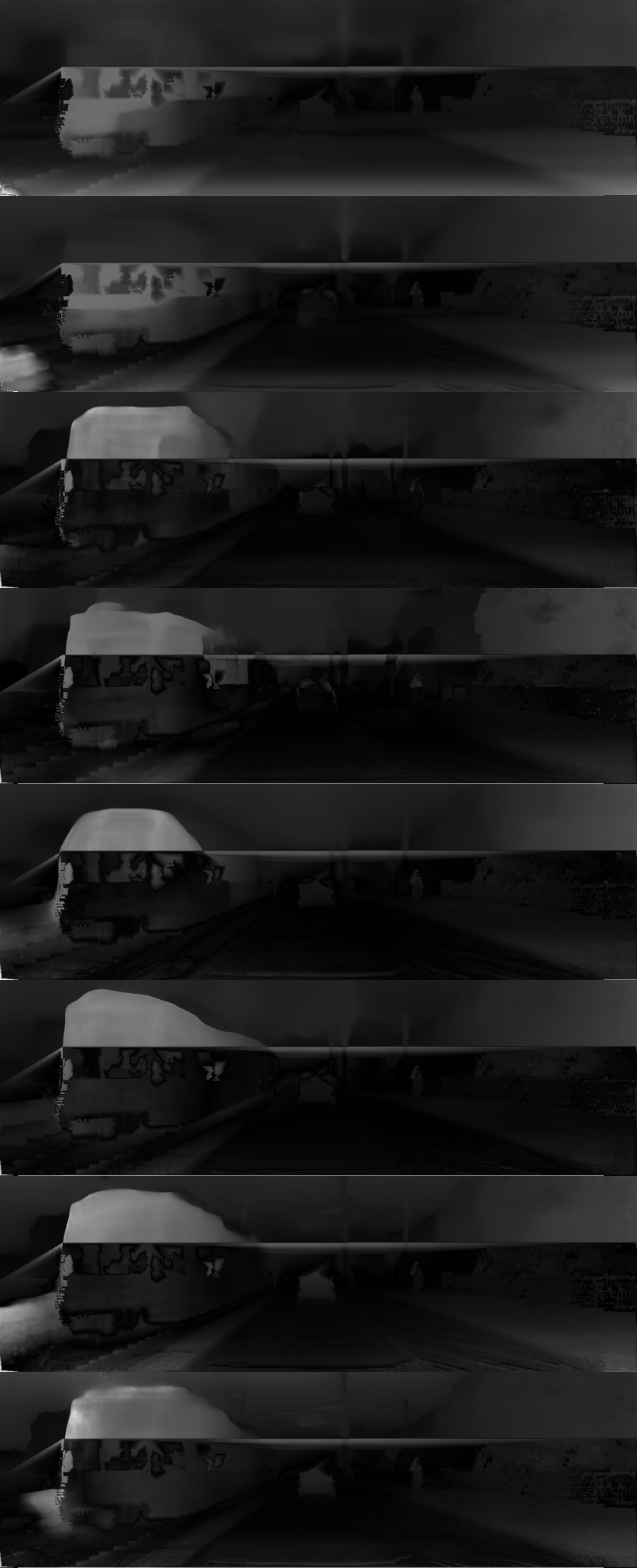}
	\includegraphics[scale=0.13]{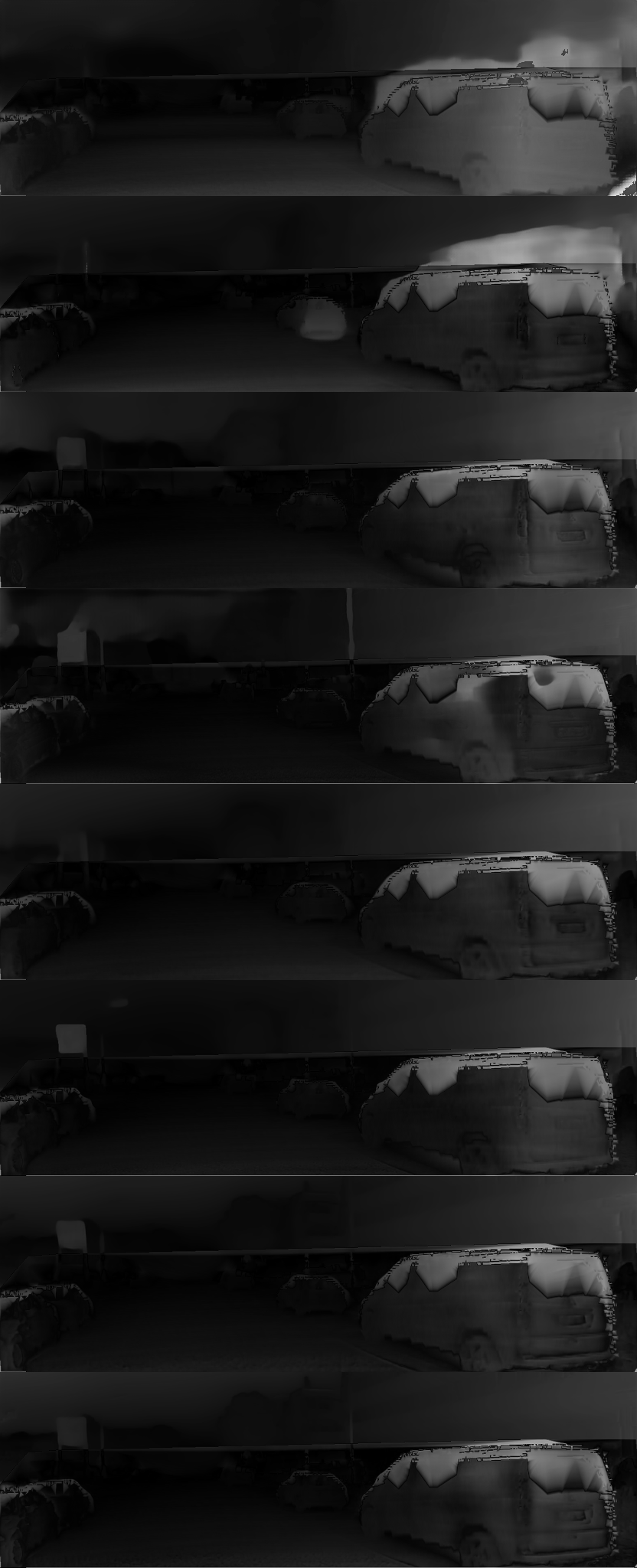}	\includegraphics[scale=0.13]{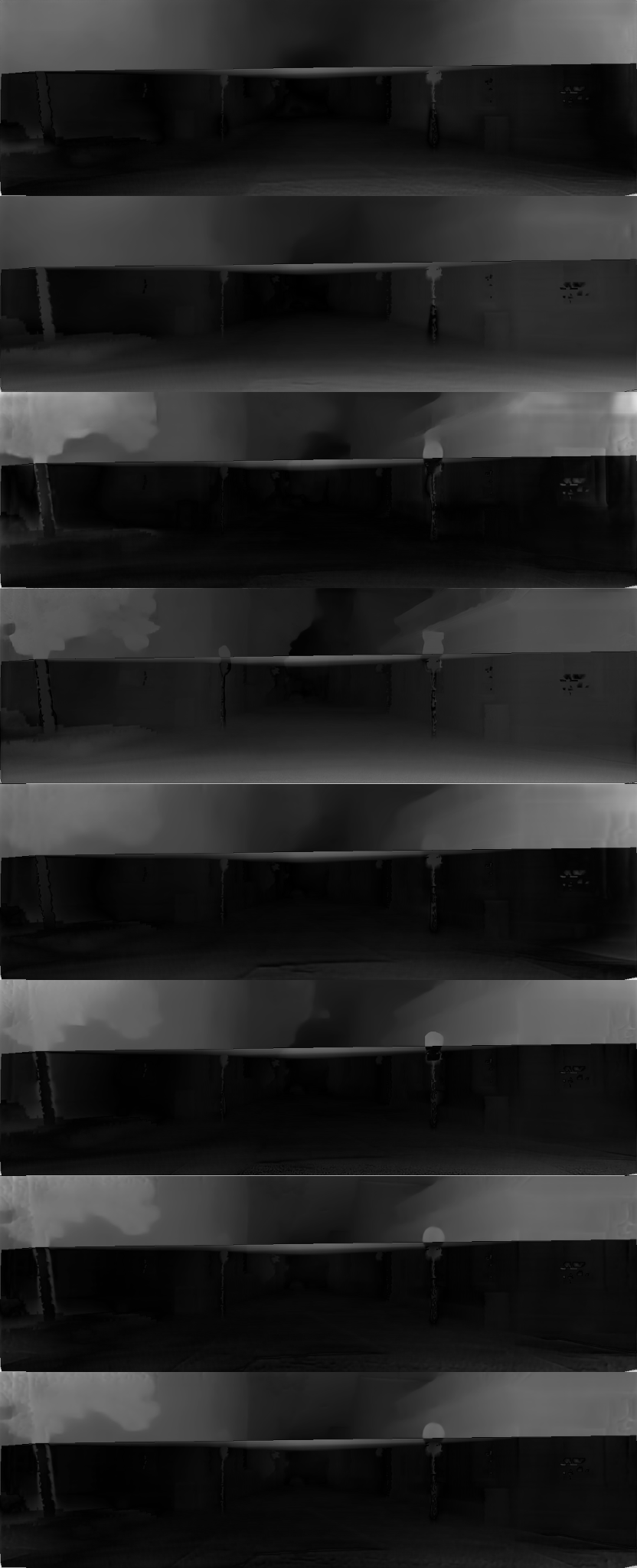}  \\
		
	$\qquad$	\\
      $\qquad$ \\

	\rotatebox{90}{
		$\quad$$\quad$Ours$\ $ 
		$\quad$$\quad$$\quad$$\quad$Ours$\ $  
		$\quad$$\quad$$\quad$$\quad$Ours$\ $ 
		$\quad$$\quad$$\quad$$\quad$Ours$\ $		
		$\quad$$\quad$Guizilini\cite{guizilini20203d} $ \ $
		$\quad$Godard\cite{godard2019digging} $ \ $
		$\quad$Bian\cite{bian2019unsupervised} $ \ $ 
		$\quad$Ranjan\cite{ranjan2019competitive}$\ $		
	} 	 
	\rotatebox{90}{$\quad$$\quad$K+CS $\quad$$\quad$$\quad$$\quad$K+CS $\quad$$\quad$$\quad$$\quad$$\quad$K$\quad$$\quad$$\quad$$\quad$$\quad$ K $\quad$$\quad$$\quad$$\quad$K+CS $\quad$$\quad$$\quad$$\quad$$\quad$K+IN$\quad$$\quad$$\quad$$\quad$$\quad$$\quad$K$\quad$$\quad$$\quad$$\quad$$\quad$$\quad$K}
	\rotatebox{90}{
		${\tiny 384\times1280}$
		${\tiny 256\times832}$
		${\tiny 384\times1280}$
		${\tiny 256\times832}$
		${\tiny 384\times1280}$
		${\tiny 320\times1024}$
		${\tiny 256\times832}$
		$\quad$${\tiny 256\times832}$

	}
	\includegraphics[scale=0.13]{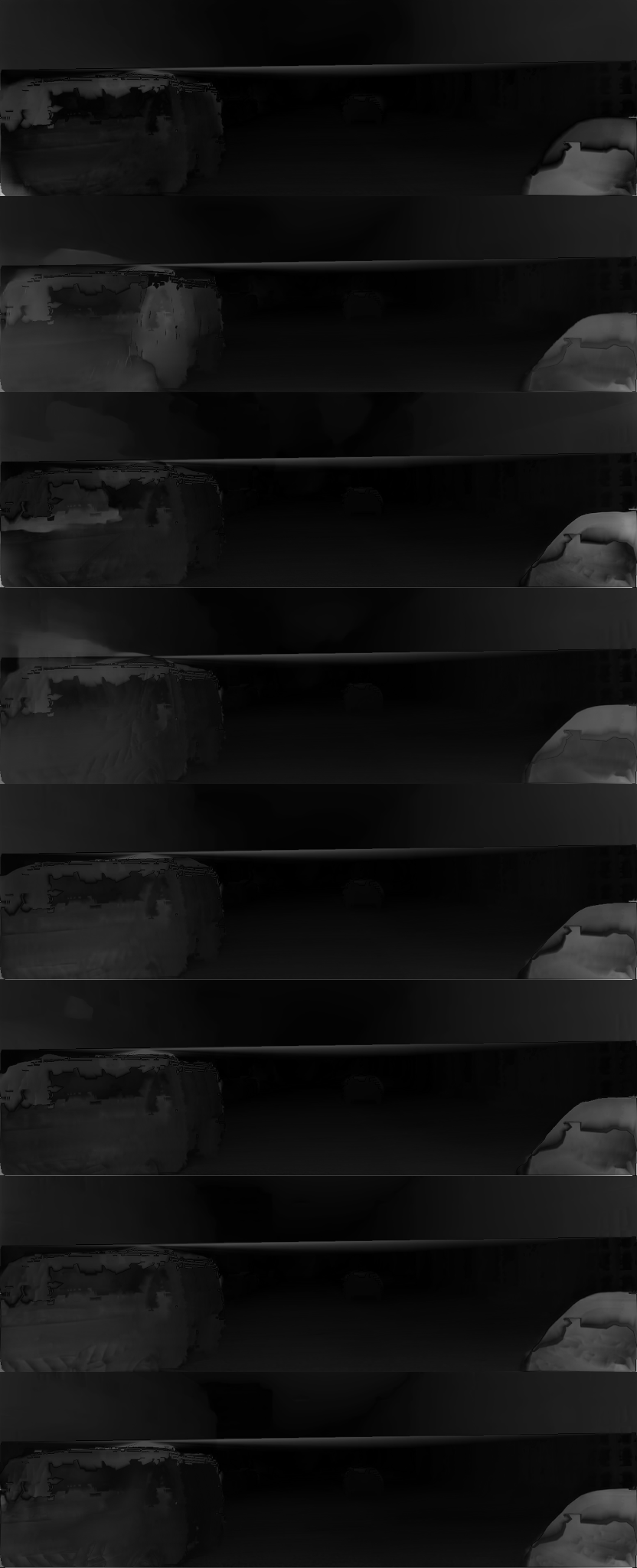}
	\includegraphics[scale=0.13]{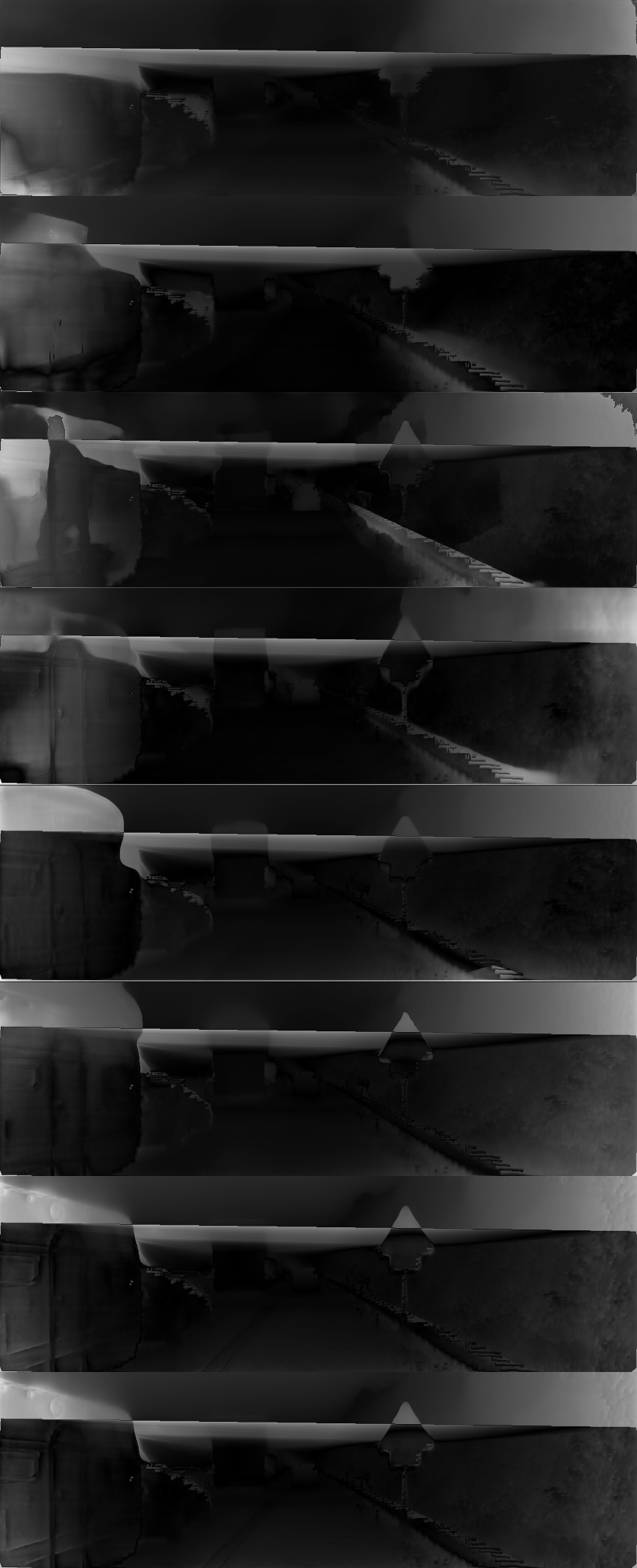}
	\includegraphics[scale=0.13]{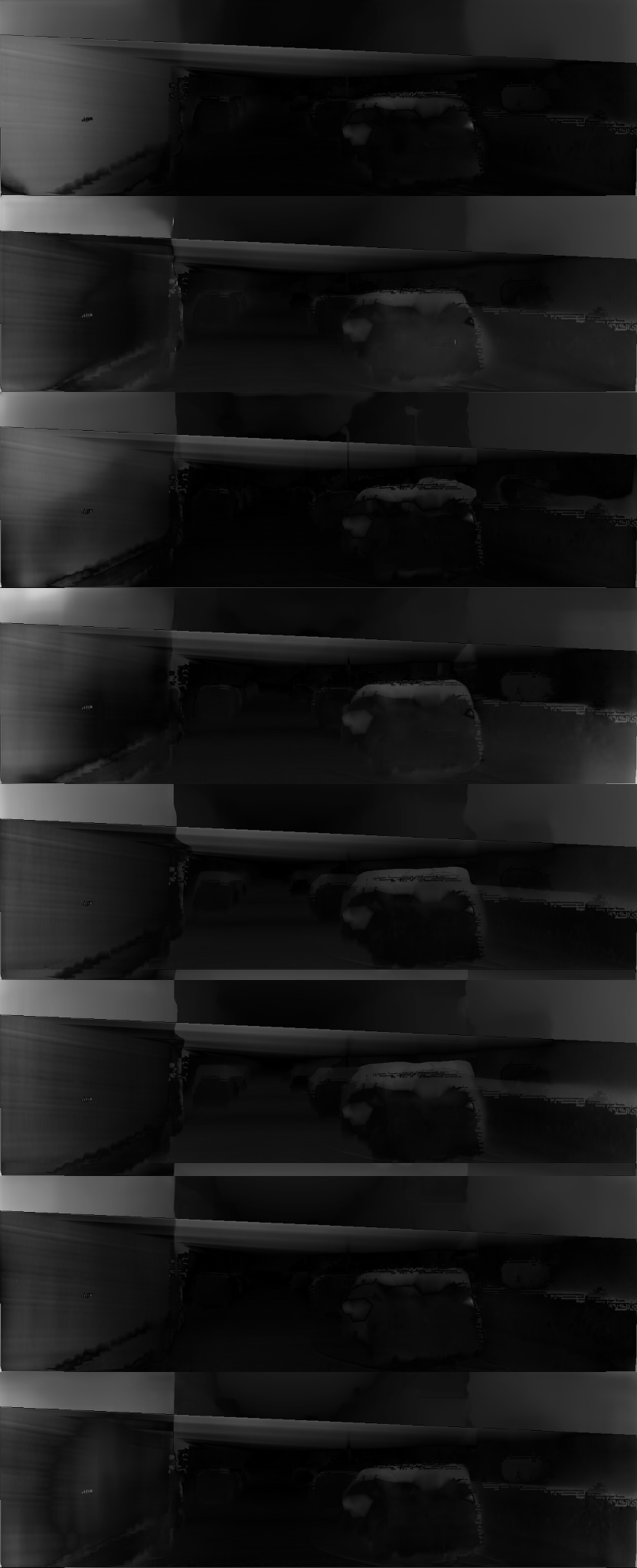}
	\includegraphics[scale=0.13]{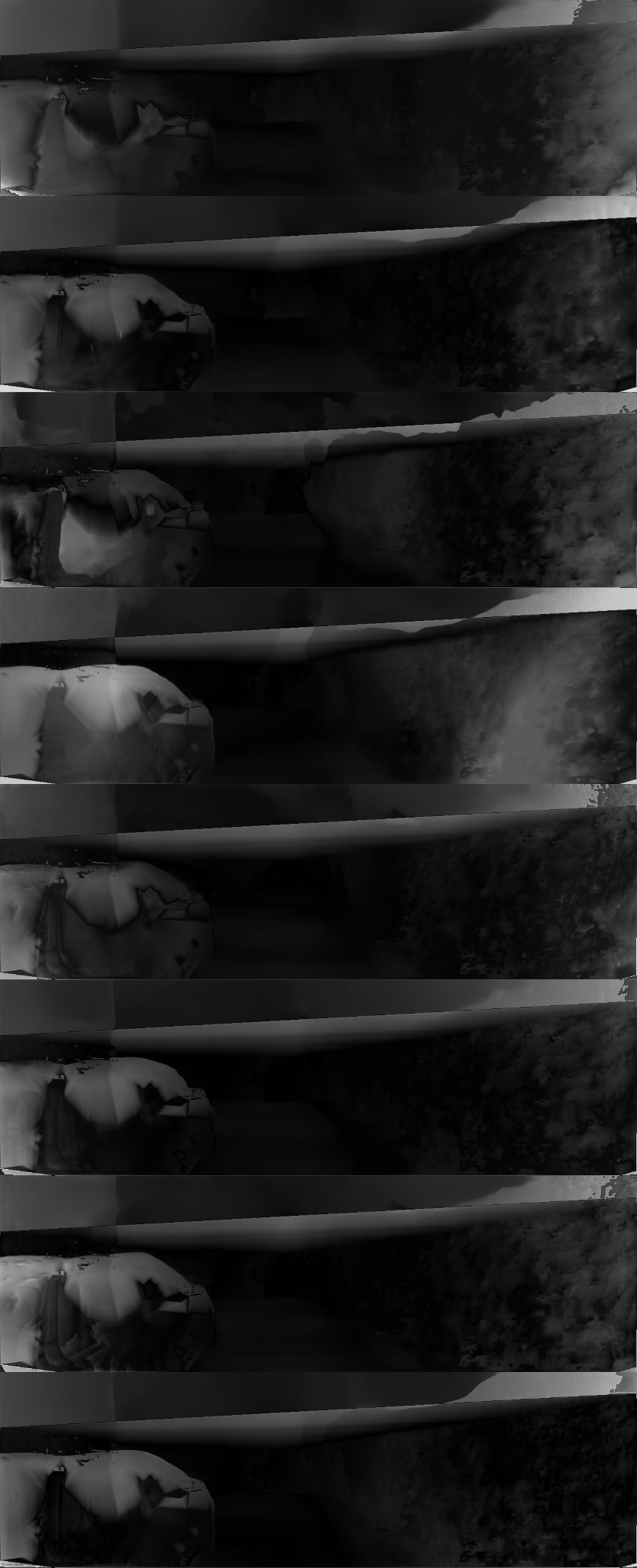}

	\caption{Comparison of error maps between different methods on the KITTI dataset. The order of the figure here corresponds to that of Fig. \ref{fig:compare_disp}.}\label{fig:compare_disp_error}%
	\vspace{-10pt}
\end{figure*}


\begin{figure*}[htbp]
	\centering
	\tiny
	\rotatebox{90}{
		$\quad$$\quad$$\quad$$\quad$Ours$\ $ 			
		$\quad$$\quad$$\quad$$\quad$$\quad$Guizilini\cite{guizilini20203d} $ \ $	
		$\quad$$\quad$$\quad$$\quad$$\quad$$\quad$GT $\quad$$\quad$$\quad$$\quad$$\quad$$\quad$$\quad$Input
	}
\rotatebox{90}{
       $\quad$$\quad$$\quad$${\tiny 384\times640}$		
        $\quad$$\quad$$\quad$${\tiny 384\times640}$}
	\includegraphics[scale=0.13]{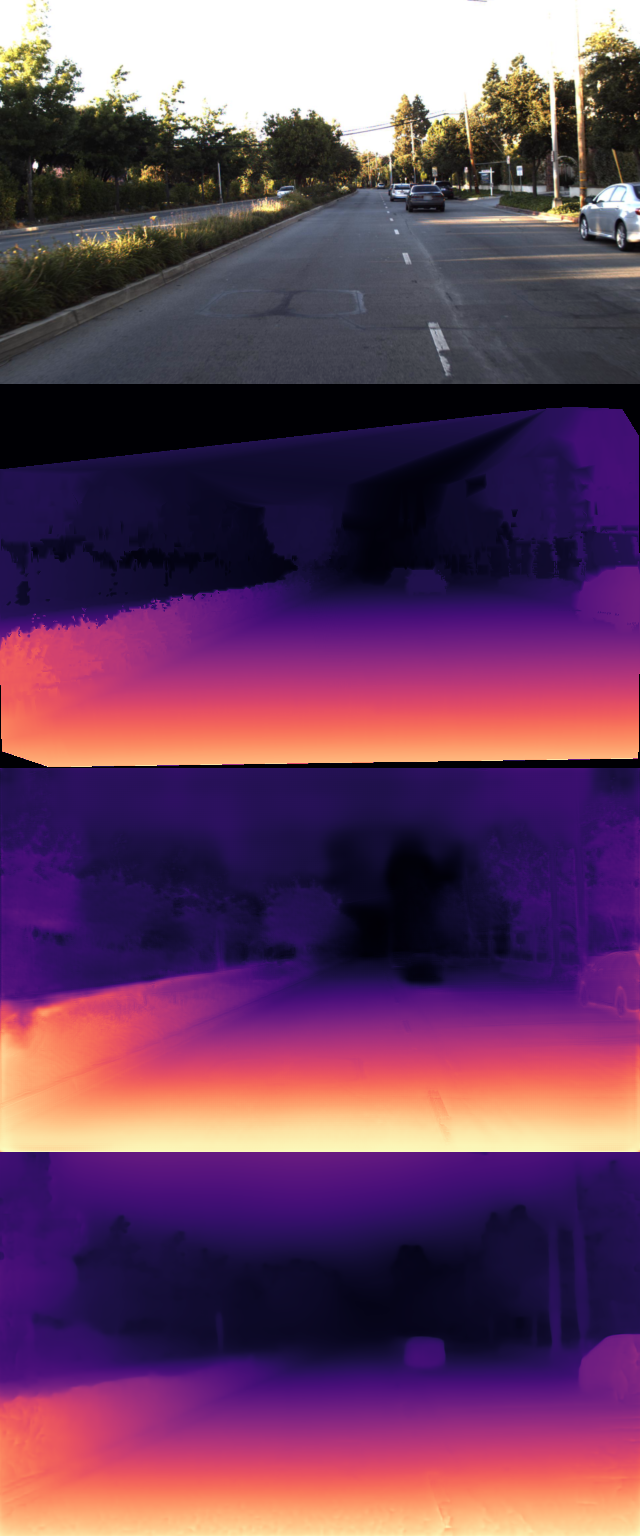}
	\includegraphics[scale=0.13]{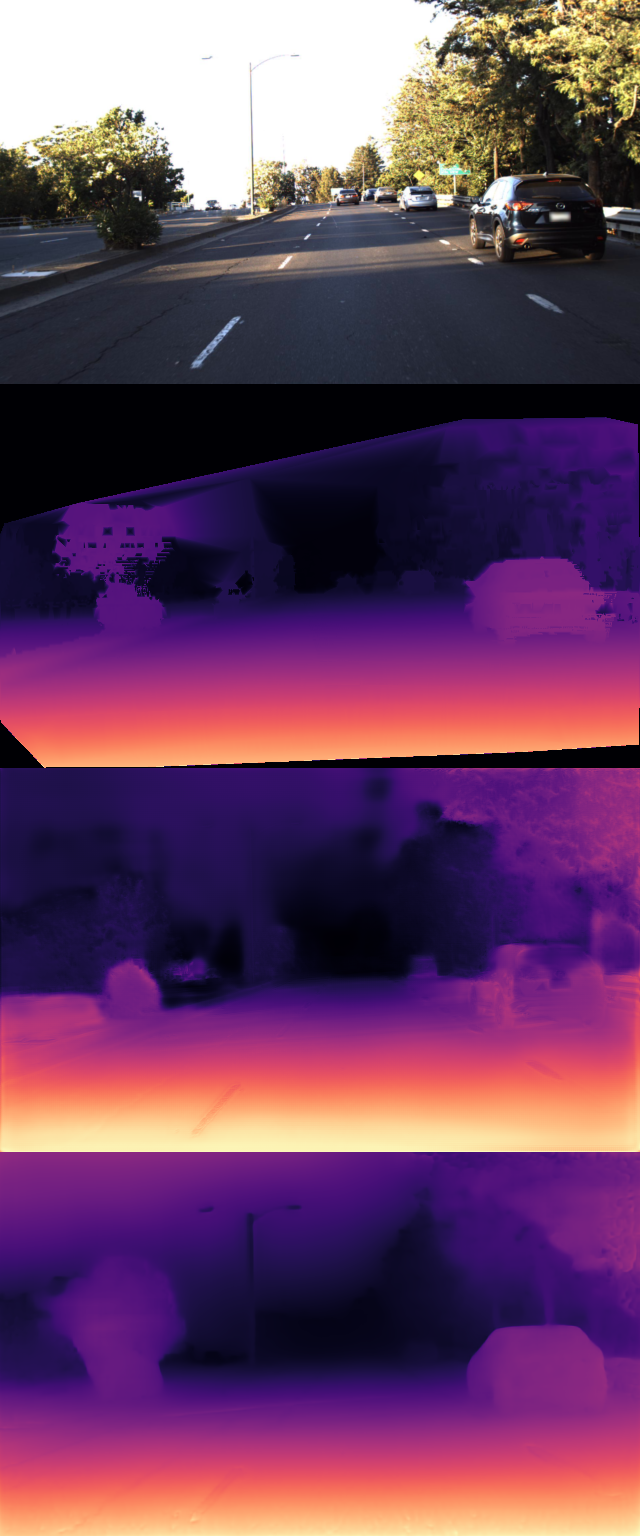}
	\includegraphics[scale=0.13]{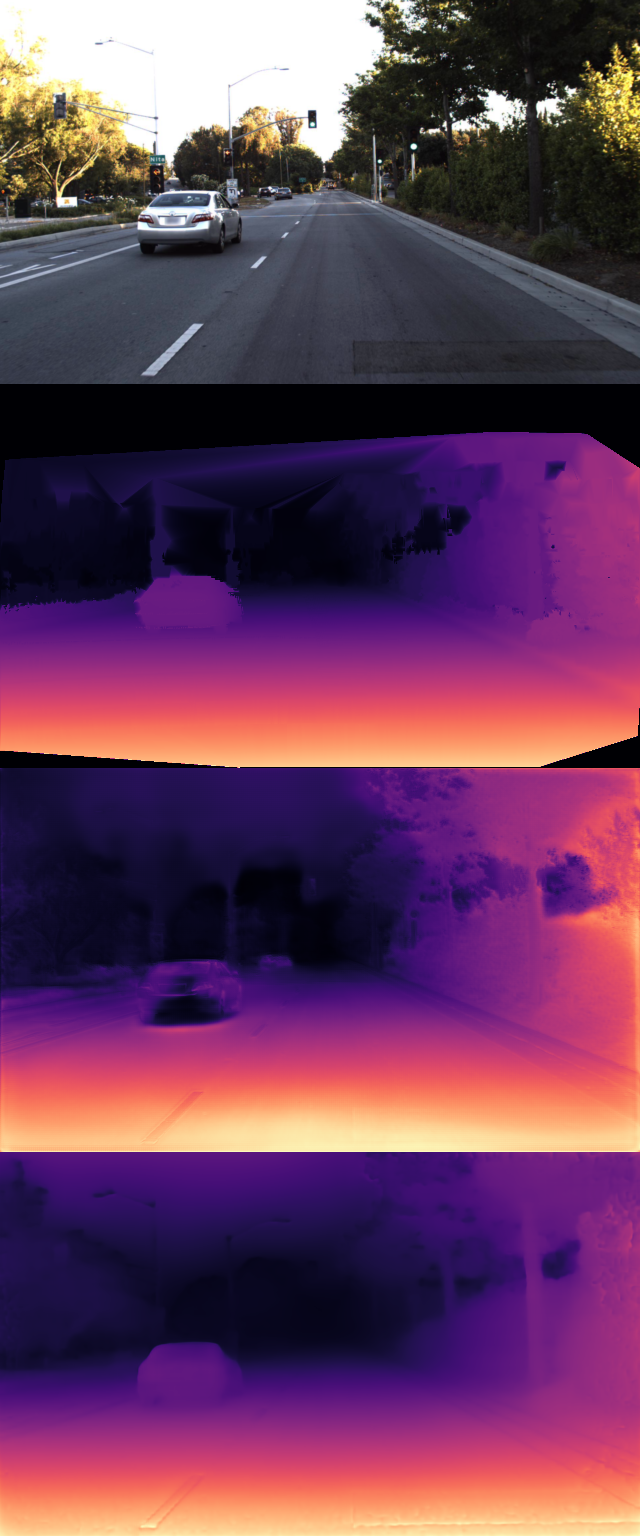}	\includegraphics[scale=0.13]{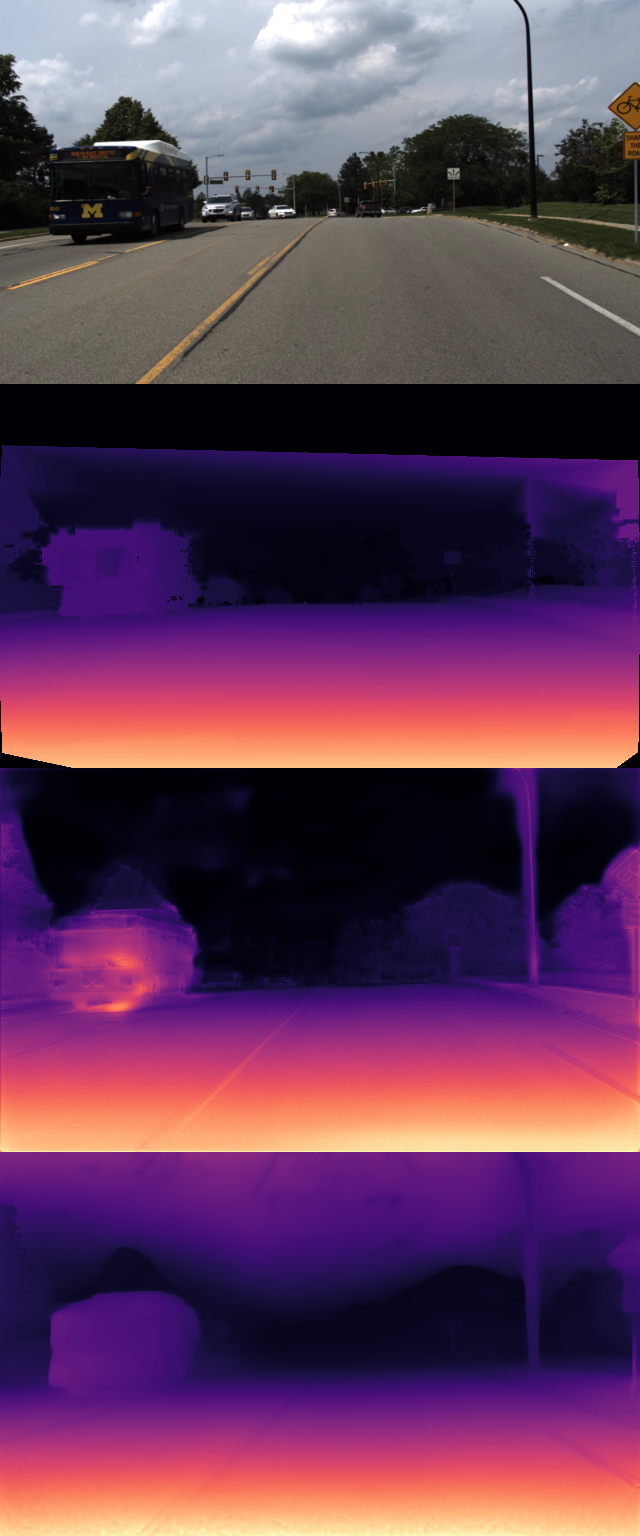} 
	\includegraphics[scale=0.13]{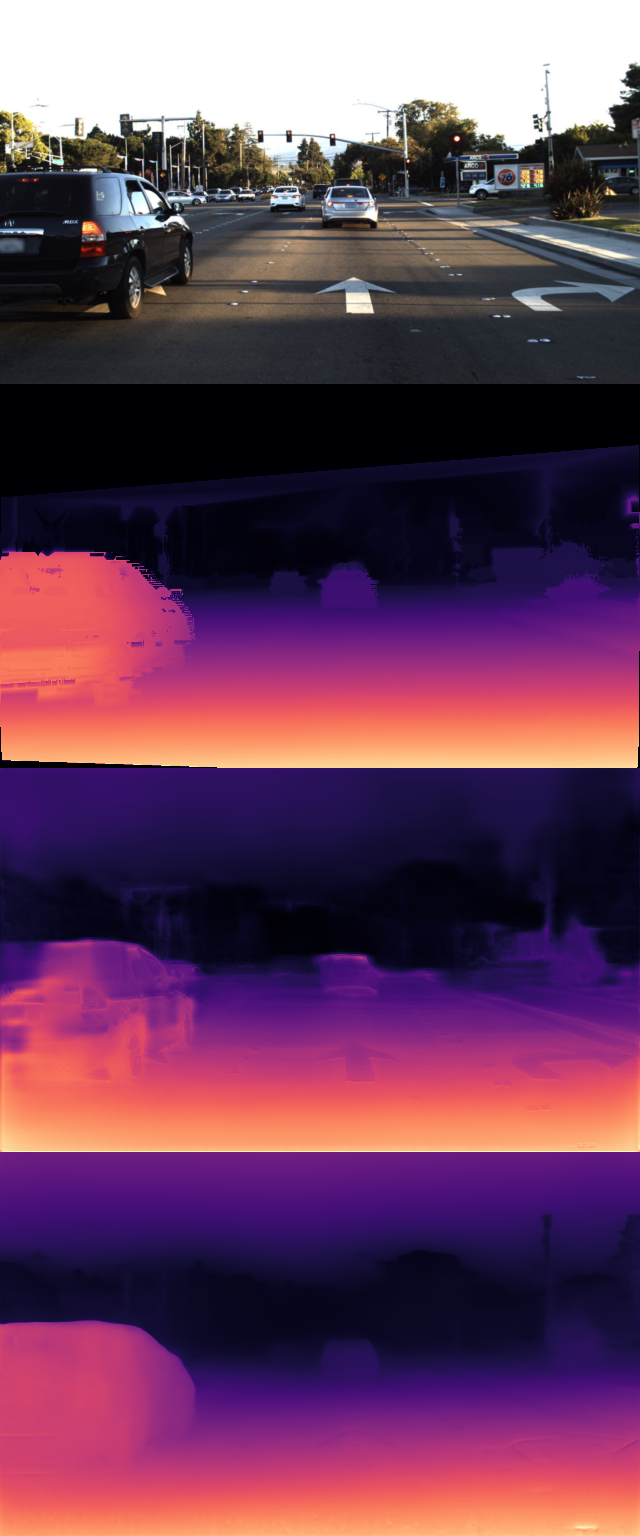}
	 \\
	 
	 $\qquad$	\\
	 $\qquad$ \\

		\tiny
	\rotatebox{90}{
		$\quad$$\quad$$\quad$$\quad$Ours$\ $ 			
		$\quad$$\quad$$\quad$$\quad$$\quad$Guizilini\cite{guizilini20203d} $ \ $	
		$\quad$$\quad$$\quad$$\quad$$\quad$$\quad$GT $\quad$$\quad$$\quad$$\quad$$\quad$$\quad$$\quad$Input
	}
	\rotatebox{90}{
		$\quad$$\quad$$\quad$${\tiny 384\times640}$		
		$\quad$$\quad$$\quad$${\tiny 384\times640}$}
	\includegraphics[scale=0.13]{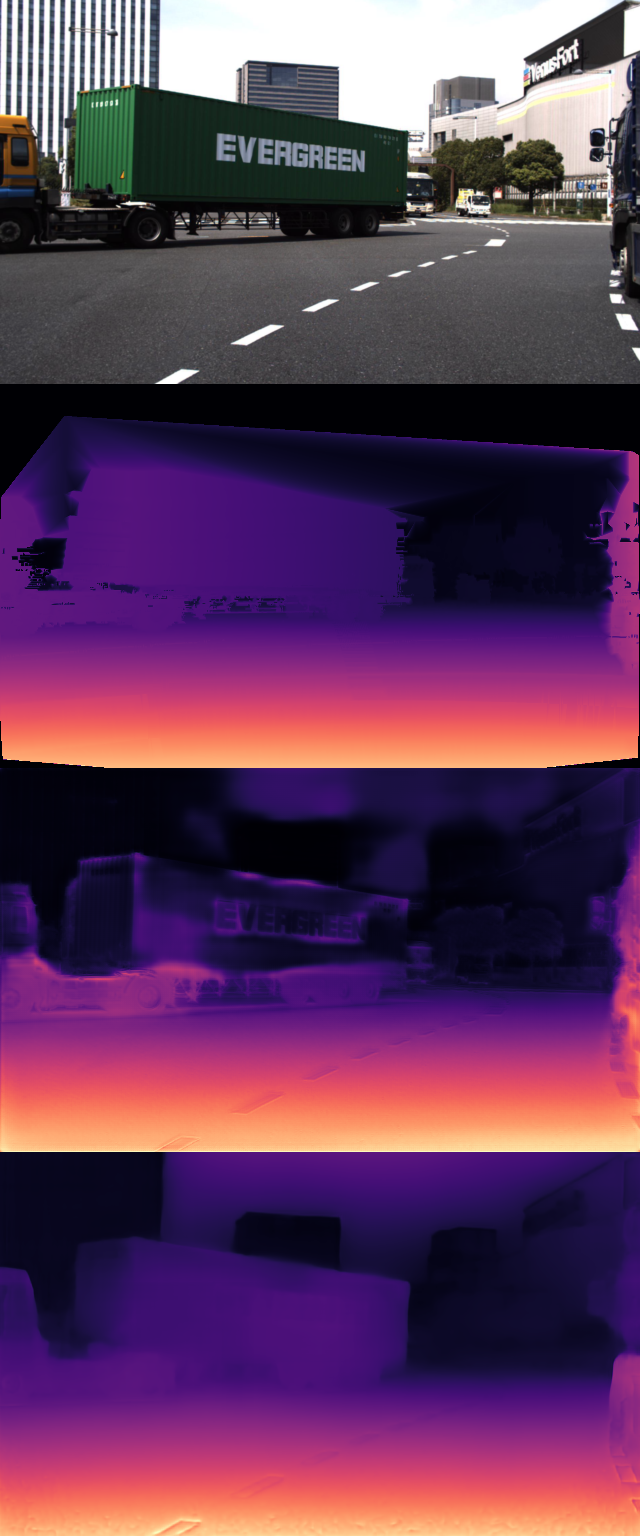}
	\includegraphics[scale=0.13]{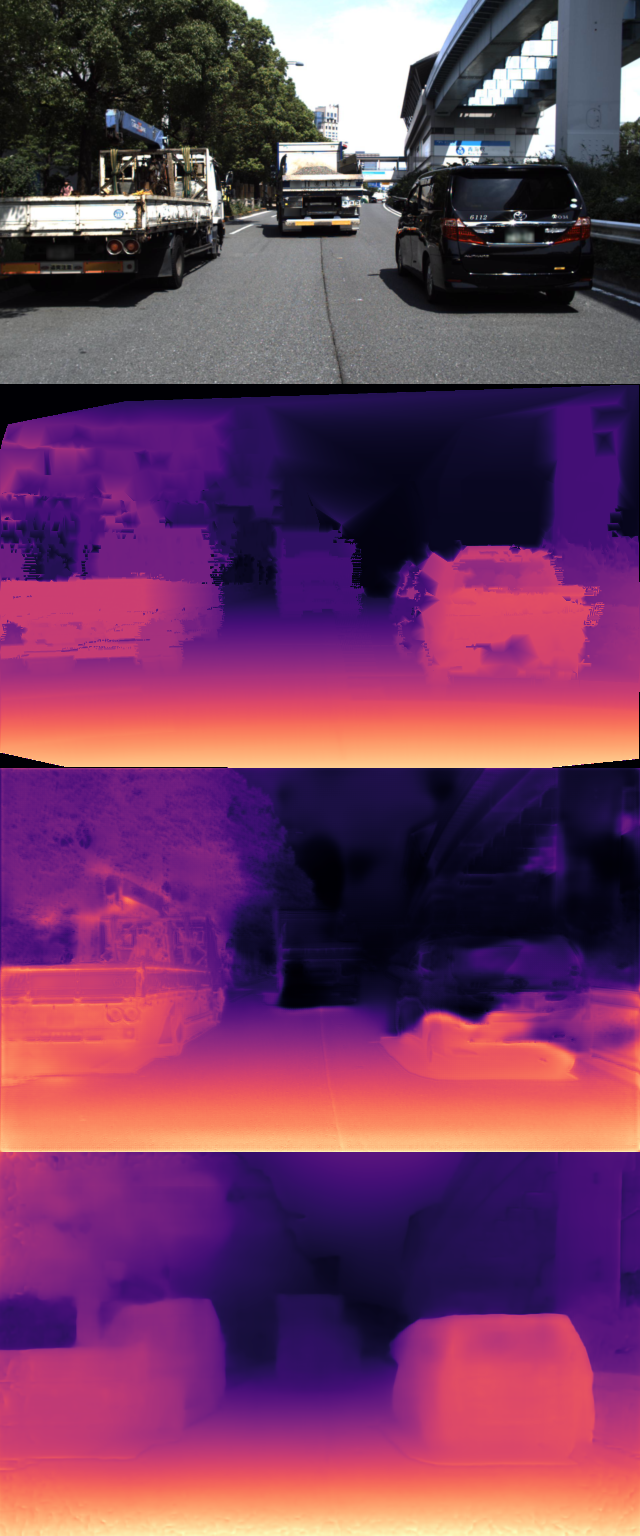}
	\includegraphics[scale=0.13]{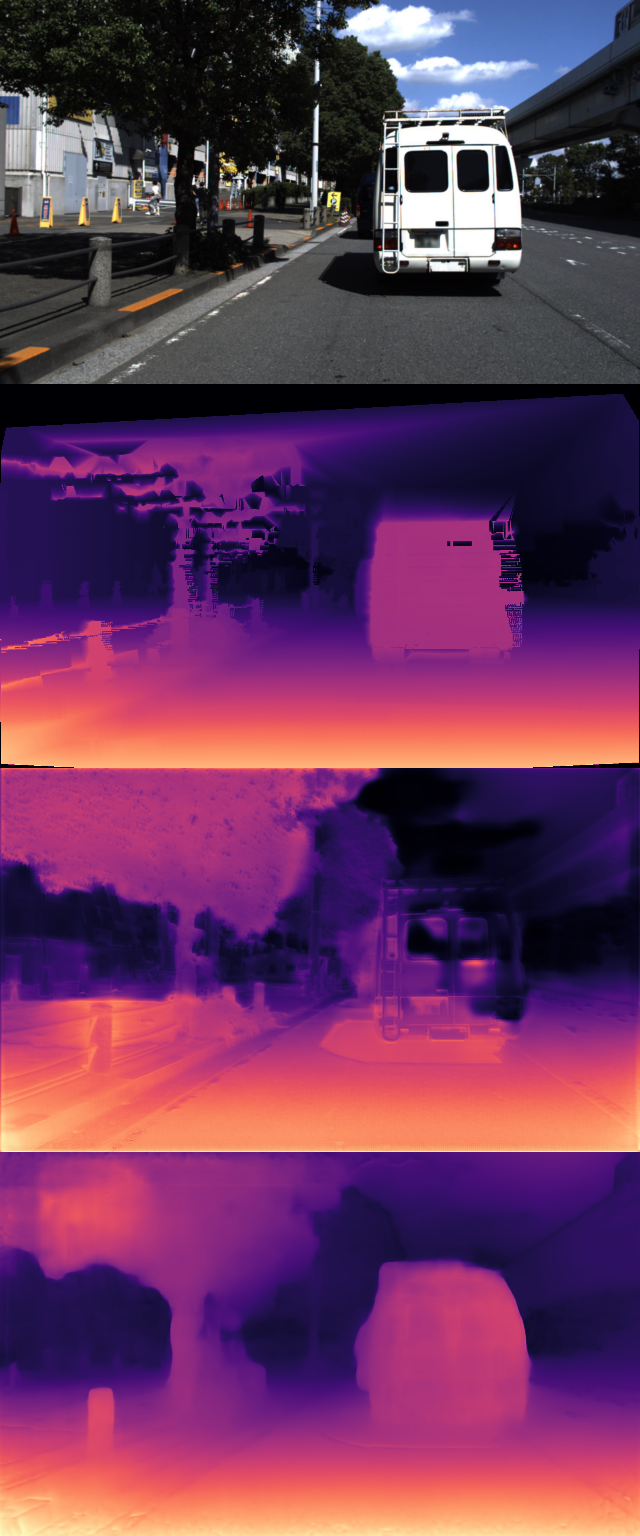}
	\includegraphics[scale=0.13]{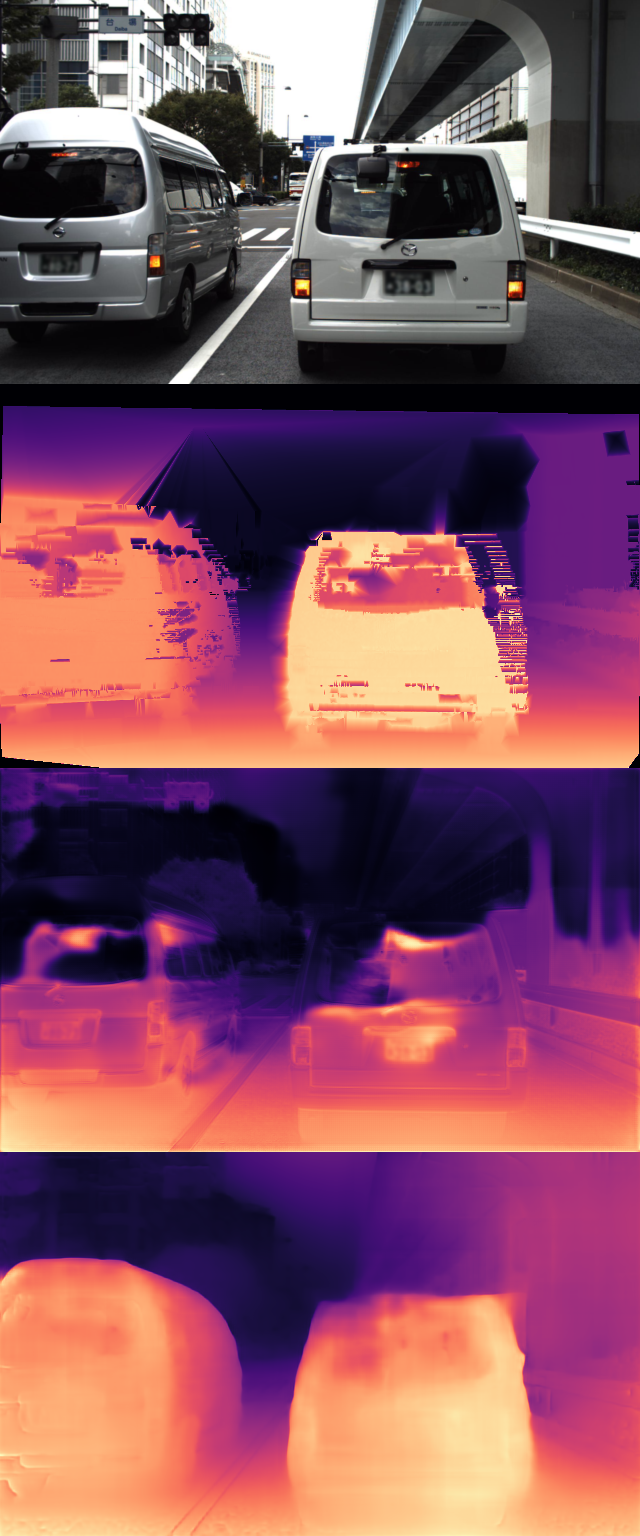}
	\includegraphics[scale=0.13]{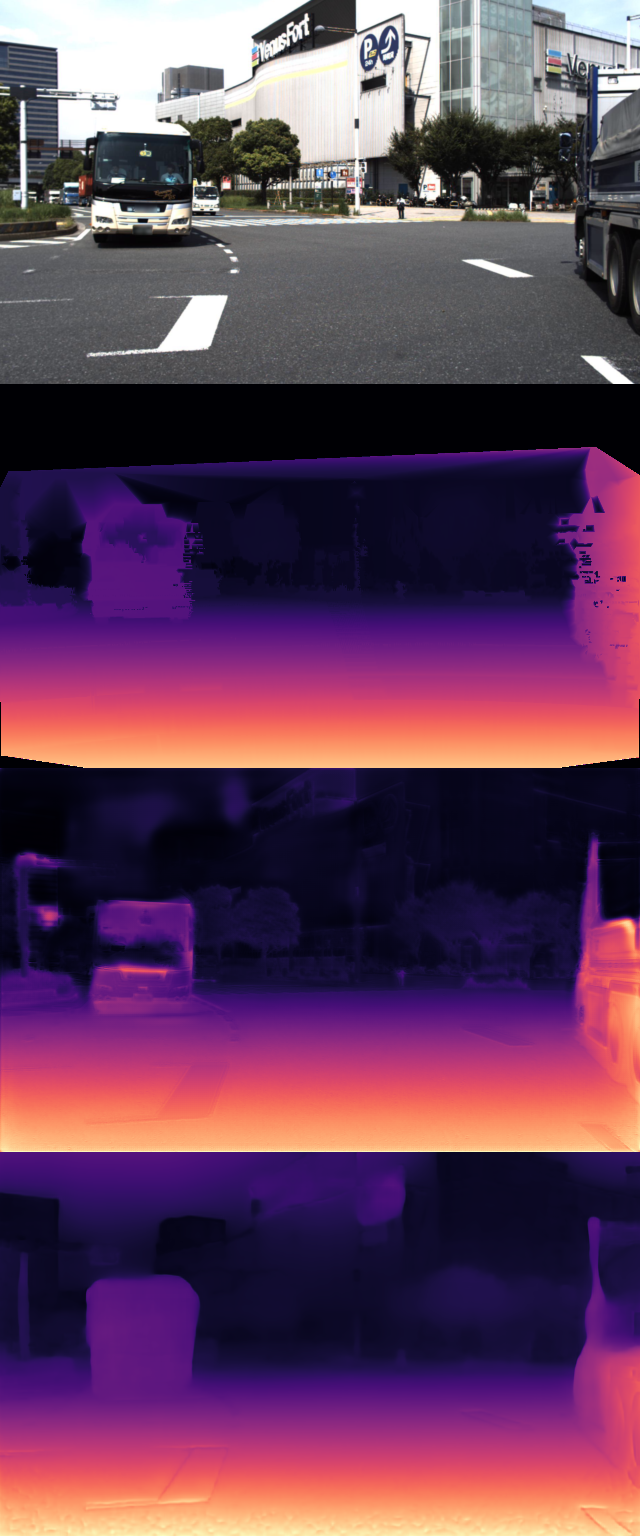}

	\caption{Qualitative comparison of example results of our proposed self-supervised monocular depth estimation method with those of previous state-of-the-art methods as estimated on the DDAD dataset.  }\label{fig:compare_disp_ddad}
	
\end{figure*}


\begin{figure*}[htbp]
	\centering
	\tiny
	\rotatebox{90}{
		$\quad$$\quad$$\quad$$\quad$Ours$\ $ 			
		$\quad$$\quad$$\quad$$\quad$$\quad$Guizilini\cite{guizilini20203d} $ \ $		
	}
	\rotatebox{90}{
		$\quad$$\quad$$\quad$${\tiny 384\times640}$		
		$\quad$$\quad$$\quad$${\tiny 384\times640}$}
	\includegraphics[scale=0.13]{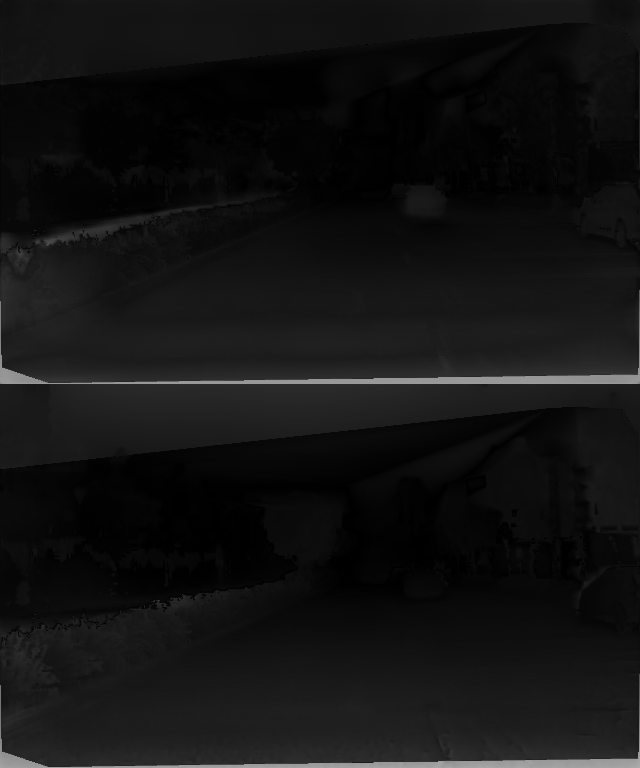}
	\includegraphics[scale=0.13]{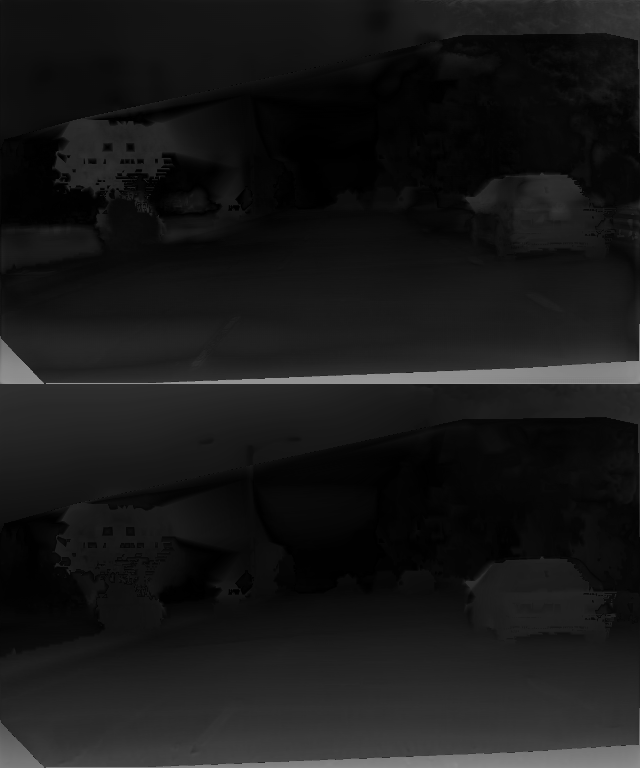}
	\includegraphics[scale=0.13]{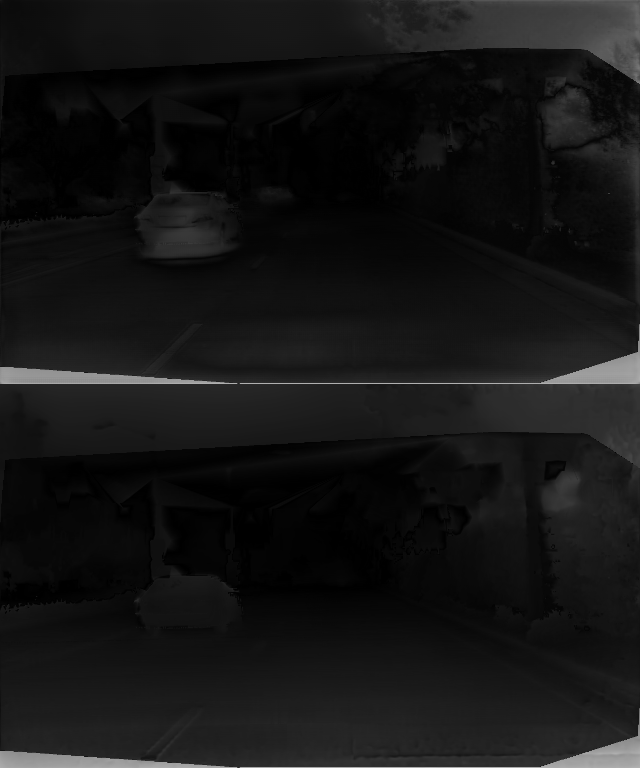}	\includegraphics[scale=0.13]{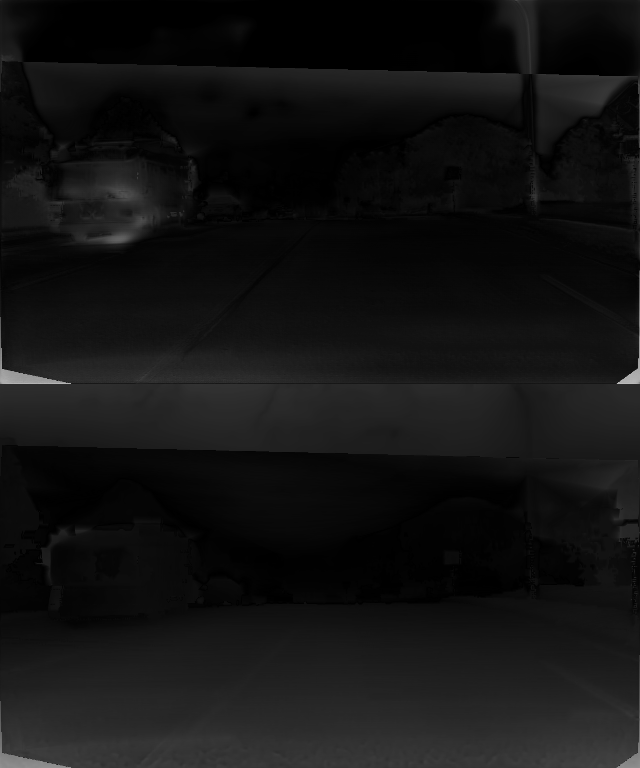} 
	\includegraphics[scale=0.13]{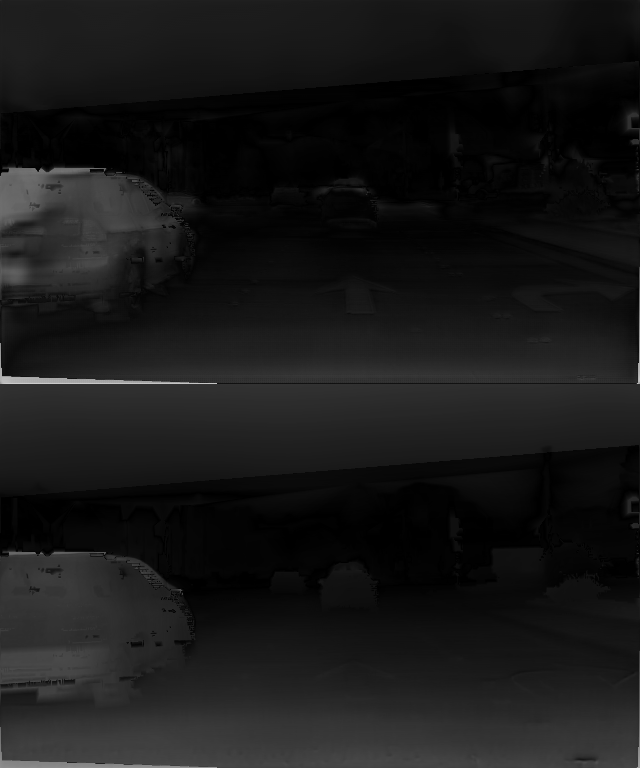}
	\\
	$\qquad$	\\
	
	\tiny
	\rotatebox{90}{
		$\quad$$\quad$$\quad$$\quad$Ours$\ $ 			
		$\quad$$\quad$$\quad$$\quad$$\quad$Guizilini\cite{guizilini20203d} $ \ $		
	}
	\rotatebox{90}{
		$\quad$$\quad$$\quad$${\tiny 384\times640}$		
		$\quad$$\quad$$\quad$${\tiny 384\times640}$}
	\includegraphics[scale=0.13]{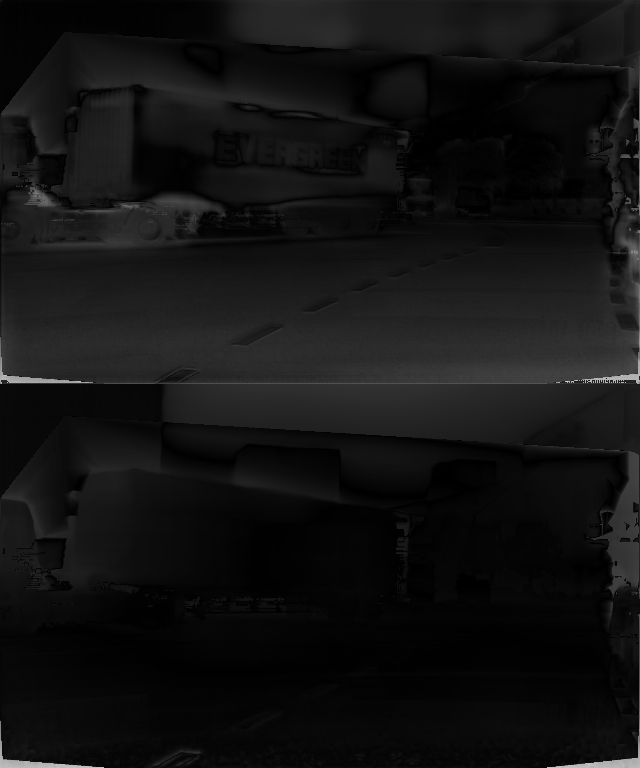}
	\includegraphics[scale=0.13]{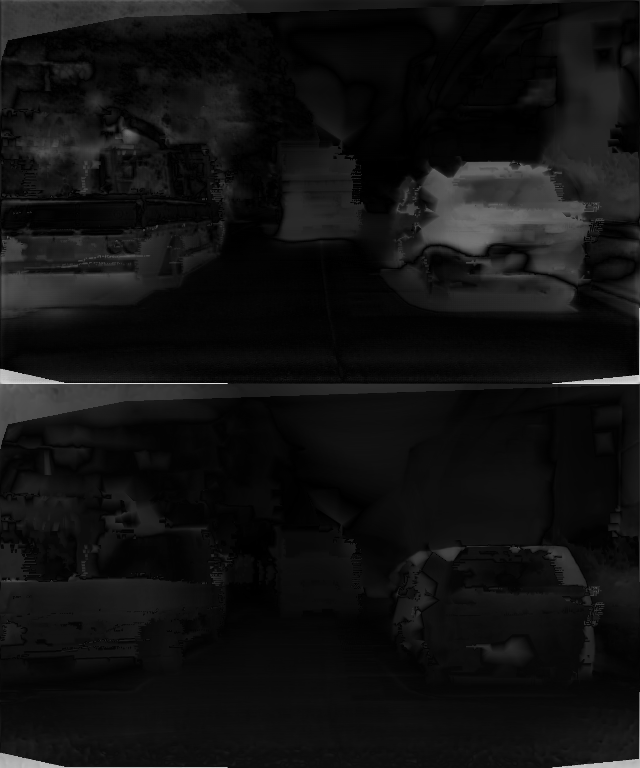}
	\includegraphics[scale=0.13]{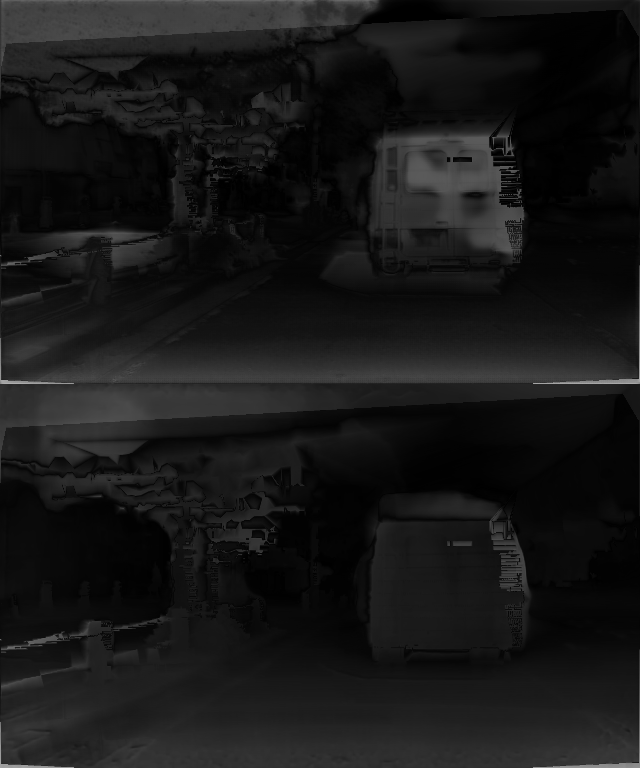}
	\includegraphics[scale=0.13]{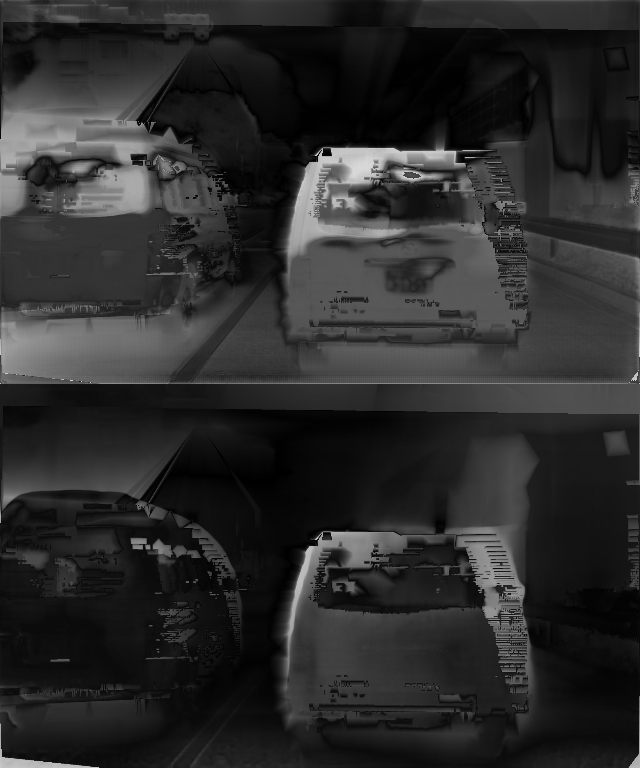}
	\includegraphics[scale=0.13]{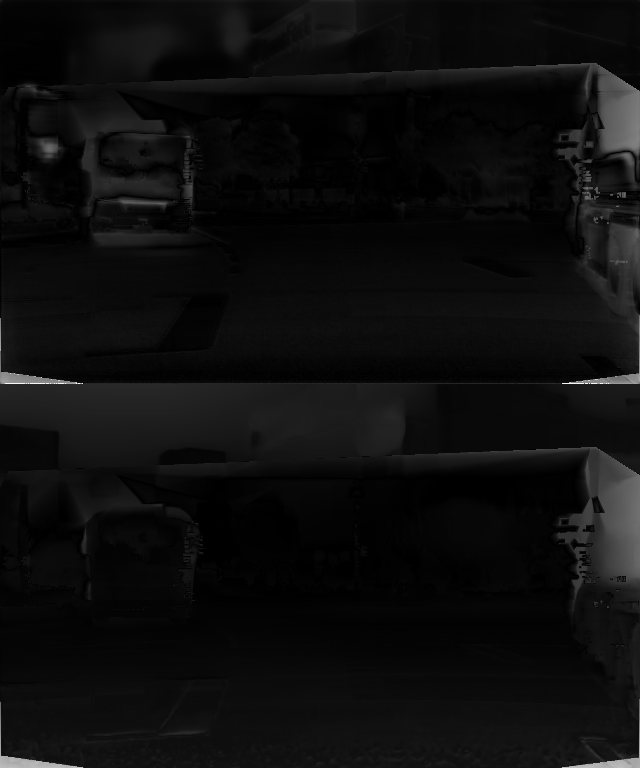}

	\caption{Comparison of error maps between different methods on the DDAD dataset. The order of the figure here corresponds to that of Fig. \ref{fig:compare_disp_ddad}. }\label{fig:compare_disp_ddad_error_map}
	
\end{figure*}

In contrast, our method accurately estimates not only the depth of the black car in the image but also the depth of the intersection regions. In addition, the depth of more distant objects can also be accurately estimated, whereas the depths in the same positions as estimated by the previous methods are 'black holes' \cite{bian2019unsupervised} or greater than the corresponding depth in the ground depth map \cite{ranjan2019competitive}. More importantly, our method is robust not only to a small range of weakly textured regions (e.g., the front windshield of the train in the second column) but also to a large range of textureless regions (e.g., the white regions in the fourth and seventh column, where the previous methods of \cite{ranjan2019competitive,bian2019unsupervised,godard2019digging,guizilini20203d},  tend to predict a either more ambiguous or rougher depth map). In order to be easier to observe the difference between different methods, we also visualized the corresponding error map (The error map here is the absolute value of the difference between the estimated depth map and the ground-truth) shown in Fig. \ref{fig:compare_disp_error}.

In Fig. \ref{fig:compare_disp_ddad}, we compare the qualitative results with the previous method \cite{guizilini20203d} on the DDAD dataset \cite{guizilini20203d}. It shows that our proposed method could achieve more accurate depth than the previous method \cite{guizilini20203d} in moving object regions. Similarly, we also provide corresponding error maps shown in Fig. \ref{fig:compare_disp_ddad_error_map} for observing their difference.

\paragraph{Camera Pose Estimation}

\begin{table*}[htbp]\small
	\setlength\tabcolsep{4pt}
	\centering	
	\begin{tabular}{lccccccccccccc} 
		\toprule 
		\multicolumn{1}{l}{\multirow{2}*{\footnotesize Method}}&		
		\multicolumn{1}{c}{\multirow{2}*{\footnotesize Cap (m)}}&		
		\multicolumn{4}{c}{\footnotesize Error$\downarrow$}& &
		\multicolumn{3}{c}{\footnotesize Accuracy$\uparrow$}\\
		\multicolumn{2}{c}{}&\footnotesize AbsRel&\footnotesize SqRel&\footnotesize RMSE&\footnotesize RMSE log& &\footnotesize $\delta<1.25$ &\footnotesize $ \delta<1.25^2$&\footnotesize $\delta<1.25^3$&	\\ 		
		\midrule
		\multicolumn{1}{l}{{\small Baseline}}&80 &0.1418&0.9628&5.2890&0.2222& &0.8081&0.9406&0.9768&\\   
		\multicolumn{1}{l}{{\small $L_{p}^{bi}$}}&80 &0.1390&1.0420&5.2572&0.2198& &0.8272&0.9417&0.9749&\\

		\multicolumn{1}{l}{{{\small $L_{p}$ + $M_{occ}$}}}& {80}&{0.1385}&{0.9717}&{5.0650}&{0.2085}& &{0.8349}&{0.9505}&{0.9810}&\\%
		
		\multicolumn{1}{l}{{\small $L_{p}^{bi}$ + $M_{occ}^{bi}$}}& 80&0.1262&0.9592&4.8118&0.2026& &0.8566&0.9535&0.9795&\\ 		
		
		\multicolumn{1}{l}{{{\small $L_{p}$ + $M_{occ}$ + $L_{dsc}$}}}&{80} &{0.1271}&{1.0097}&{4.9408}&{0.2037}& &{0.8545}&{0.9527}&{0.9794}&\\
		\multicolumn{1}{l}{{\small $L_{p}^{bi}$ + $M_{occ}^{bi}$ + $L_{dsc}^{bi}$}}&80 &0.1234&0.9984&4.9396&0.1988& &0.8585&0.9548&0.9806&\\

		\multicolumn{1}{l}{{{\small $L_{p}$ + $M_{occ}$ + $L_{dsc}$ + $W_{aw}$}}}&{80} &{0.1222}&{1.0042}&{4.9935}&{0.1990}& &{0.8651}&{0.9558}&{0.9799}&\\

		\multicolumn{1}{l}{{\small $L_{p}^{bi}$ + $M_{occ}^{bi}$ + $L_{dsc}^{bi}$ + $W_{aw}^{bi}$}}&80 &0.1219&0.9833&4.9281&0.1980& &0.8645&0.9558&0.9802&\\

		\multicolumn{1}{l}{{{\small $L_{p}$ + $M_{occ}$ + $L_{dsc}$ + $W_{aw}$ + $L_{feat}$}}}&{80} &{0.1217}&{1.0233}&{5.0100}&{0.1991}& &{0.8690}&{0.9557}&{0.9794}&\\ 
		
		\multicolumn{1}{l}{{\small $L_{p}^{bi}$ + $M_{occ}^{bi}$ + $L_{dsc}^{bi}$ + $W_{aw}^{bi}$ + $L_{feat}^{bi}$}}&80 &0.1199&0.9474&4.9405&0.1965& &0.8630&0.9570&0.9814&\\   

		\multicolumn{1}{l}{{\small $L_{p}^{bi}$ + $M_{occ}^{bi}$ + $L_{dsc}^{bi}$ + $W_{aw}^{bi}$ + $L_{feat}^{bi}$\textdagger}}&{80} &{0.1099}&{0.8286}&{4.6139}&{0.1851}& &{0.8801}&{0.9624}&{0.9828}&\\

		\midrule
		
		\multicolumn{1}{l}{{\small Baseline}}&50&0.1370&0.7844&4.0926&0.2109& &0.8235&0.9486&0.9796&\\
		\multicolumn{1}{l}{{\small $L_{p}^{bi}$}}&50&0.1345&0.8886&4.2324&0.2098& &0.8399&0.9474&0.9772&\\

		\multicolumn{1}{l}{{{\small $L_{p}$ + $M_{occ}$}}}&{50} &{0.1333}&{0.8174}&{4.0294}&{0.1989}& &{0.8492}&{0.9568}&{0.9828}& \\
		
		\multicolumn{1}{l}{{\small $L_{p}^{bi}$ + $M_{occ}^{bi}$}}&50 &0.1228&0.8463&3.9583&0.1951& &0.8675&0.9573&0.9808& \\

		\multicolumn{1}{l}{{{\small $L_{p}$ + $M_{occ}$ + $L_{dsc}$}}}&{50} &{0.1230}&{0.8831}&{4.0035}&{0.1954}& &{0.8692}&{0.9581}&{0.9805}&\\
		
		\multicolumn{1}{l}{{\small $L_{p}^{bi}$ + $M_{occ}^{bi}$ + $L_{dsc}^{bi}$}}&50 &0.1194&0.8794&4.0676&0.1906& &0.8714&0.9596&0.9819&\\

		\multicolumn{1}{l}{{{\small $L_{p}$ + $M_{occ}$ + $L_{dsc}$ + $W_{aw}$}}}&{50}&{0.1185}&{0.8585}&{4.0925}&{0.1906}& &{0.8772}&{0.9603}&{0.9815}&\\ 
		
		\multicolumn{1}{l}{{\small $L_{p}^{bi}$ + $M_{occ}^{bi}$ + $L_{dsc}^{bi}$ + $W_{aw}^{bi}$}}&50&0.1178&0.8575&4.0094&0.1894& &0.8774&0.9608&0.9816&\\

		\multicolumn{1}{l}{{{\small $L_{p}$ + $M_{occ}$ + $L_{dsc}$ + $W_{aw}$ + $L_{feat}$}}}&{50} &{0.1180}&{0.9032}&{4.1571}&{0.1912}& &{0.8804}&{0.9601}&{0.9808}&\\
		
		\multicolumn{1}{l}{{\small $L_{p}^{bi}$ + $M_{occ}^{bi}$ + $L_{dsc}^{bi}$ + $W_{aw}^{bi}$ + $L_{feat}^{bi}$}}&50 &0.1155&0.8169&4.0249&0.1876& &0.8758&0.9619&0.9830&\\
		
		\multicolumn{1}{l}{{\small $L_{p}^{bi}$ + $M_{occ}^{bi}$ + $L_{dsc}^{bi}$ + $W_{aw}^{bi}$ + $L_{feat}^{bi}$}\textdagger}&{50} &{0.1060}&{0.7156}&{3.7449}&{0.1770}& &{0.8917}&{0.9663}&{0.9840}&\\

		\bottomrule 
	\end{tabular}
	\caption{Ablation studies on monocular depth estimation. The results were evaluated on the KITTI Eigen split with the depth capped at 80 m and 50 m. $\delta$ represents the ratio between the estimated depth and ground truth depths. \textdagger \,   indicates that three consecutive frames with a height of 384 and a width of 1280 were used as a training sample for depth and camera pose estimation experiments.}\label{tab:diff_func_ablation_80m_50m}	
	\vspace{-10pt}
\end{table*}

\begin{table}[htbp]\small	
	\setlength\tabcolsep{1.5pt}
	\centering	
	\begin{tabular}{lcccc} 
		\toprule 
		\multicolumn{1}{l}{{\footnotesize Method}}& \multicolumn{1}{c}{{\footnotesize Seq. 09}}& \multicolumn{1}{c}{{\footnotesize Seq. 10}}\\
			
		\hline 
		\multicolumn{1}{l}{{\footnotesize Baseline}} &{0.0369$\pm$0.0369}  &{0.0257$\pm$0.0271} \\  
		\multicolumn{1}{l}{{\footnotesize $L_{p}^{bi}$}}& {0.0133$\pm$0.0066}&{0.0120$\pm$0.0086} \\

		\multicolumn{1}{l}{{{\footnotesize $L_{p}$ + $M_{occ}$}}} &{{0.0170$\pm$0.0062} } &{{0.0144$\pm$0.0085}} \\
		
		\multicolumn{1}{l}{{\footnotesize $L_{p}^{bi}$ + $M_{occ}^{bi}$}} &{0.0130$\pm$0.0064}  &{0.0119$\pm$0.0086} \\

		\multicolumn{1}{l}{{{\footnotesize $L_{p}$ + $M_{occ}$ + $L_{dsc}$ }}}&{0.0167$\pm$0.0066} &{0.0142$\pm$0.0086}  \\
		
		\multicolumn{1}{l}{{\footnotesize $L_{p}^{bi}$ + $M_{occ}^{bi}$ + $L_{dsc}^{bi}$ }}&{0.0129$\pm$0.0067} &{0.0120$\pm$0.0085}  \\ 

		\multicolumn{1}{l}{{\footnotesize $L_{p}$ + $M_{occ}$ + $L_{dsc}$ + $W_{aw}$}} &
		{0.0154$\pm$0.0057}  &   {0.0133$\pm$0.0083} \\ 
		
		\multicolumn{1}{l}{{\footnotesize $L_{p}^{bi}$ + $M_{occ}^{bi}$ + $L_{dsc}^{bi}$ + $W_{aw}^{bi}$}} &
		{0.0123$\pm$0.0065}  &   {0.0119$\pm$0.0083} \\

		\multicolumn{1}{l}{{\footnotesize $L_{p}$ + $M_{occ}$ + $L_{dsc}$ + $W_{aw}$ + $L_{feat}$}}& {0.0148$\pm$0.0061}& {0.0128$\pm$0.0083}\\ 
		
		\multicolumn{1}{l}{{\footnotesize $L_{p}^{bi}$ + $M_{occ}^{bi}$ + $L_{dsc}^{bi}$ + $W_{aw}^{bi}$ + $L_{feat}^{bi}$}}& {0.0120$\pm$0.0068}& {0.0118$\pm$0.0081}\\  

		\multicolumn{1}{l}{{\footnotesize $L_{p}^{bi}$ + $M_{occ}^{bi}$ + $L_{dsc}^{bi}$ + $W_{aw}^{bi}$ + $L_{feat}^{bi}$\textdagger}}& {0.0084$\pm$0.0047}& {0.0084$\pm$0.0064}\\

		\bottomrule
	\end{tabular}
	\caption{Ablation studies on camera pose estimation. The results were tested on sequences 09 and 10 in the KITTI Odometry dataset. \textdagger \,   indicates that three consecutive frames with a height of 384 and a width of 1280 were used as a training sample for depth and camera pose estimation experiments.}\label{tab:diff_objective_function_camera_pose}
	\vspace{-10pt}
\end{table}
 
In Tab. \ref{tab:camera_pose_previous_method}, we compare the results of recent methods based on deep learning with the results of simultaneous localization and mapping based on Oriented FAST and Rotated BRIEF features (ORB-SLAM) \cite{mur2015orb} as a reference. Our model can still achieve better results than ORB-SLAM (full) despite utilizing a rather short sequence. This performance improvement is attributed to the fact that high-level semantic features can be extracted in addition to low-level features. More importantly, although our CameraNet and the methods of \cite{zhou2017unsupervised,zou2018df,luo2019every,ranjan2019competitive}, and \cite{mahjourian2018unsupervised} all have the same network architecture, our method achieves more significant improvements in the ATE. This may be because our method benefits from the proposed objective function, which provides better constraints for network optimization.

\subsection{Ablation Studies}
\begin{figure}[htbp]
	\centering 	
	\includegraphics[scale=0.13]{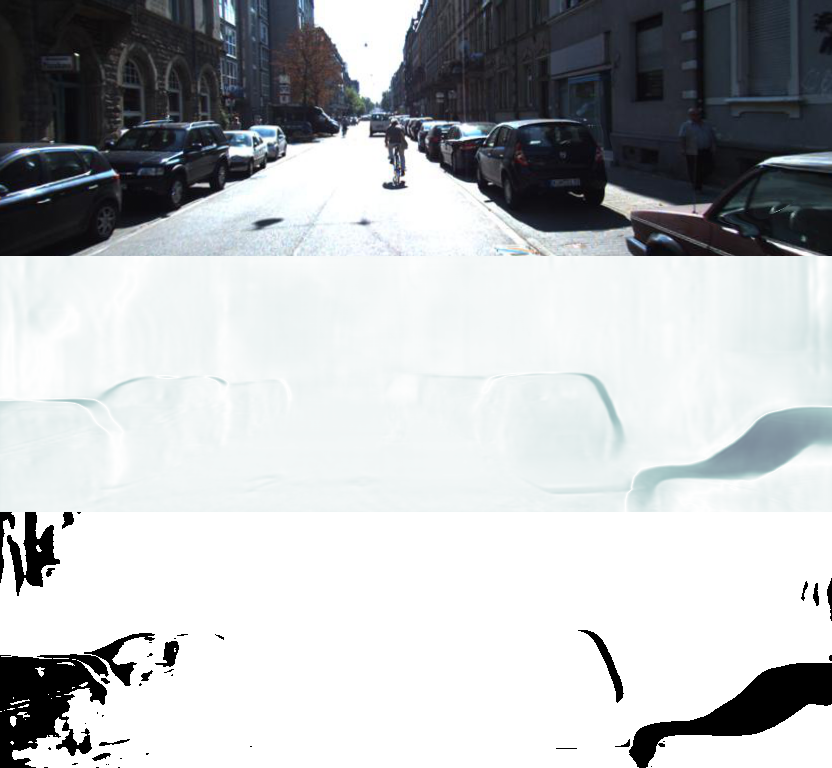}
	\includegraphics[scale=0.13]{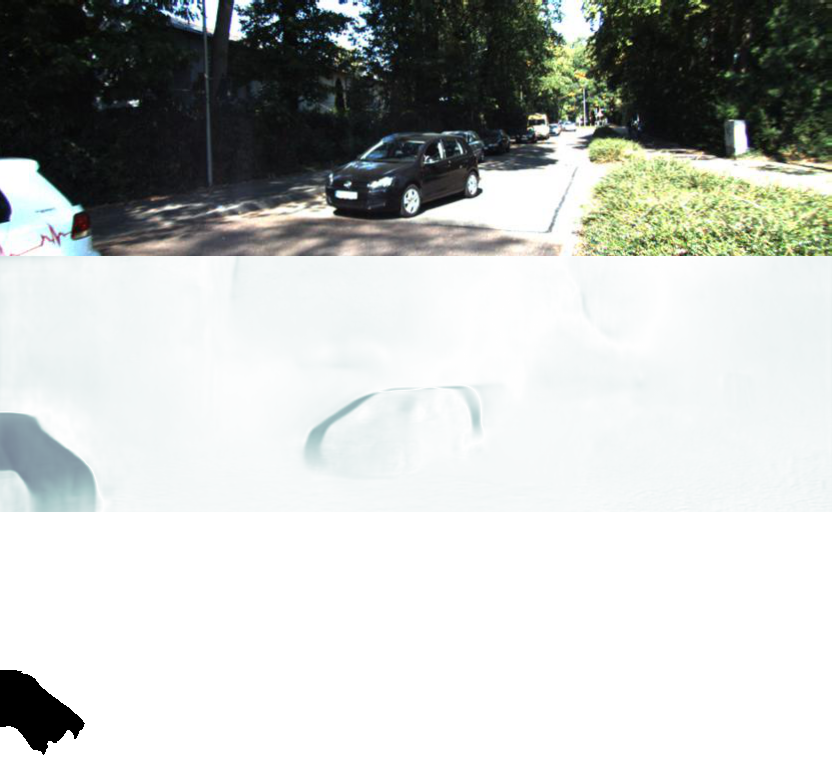}	
	\caption{Qualitative examples of complementary occlusion masks. From top to bottom are the original images, the adaptive weights, and the bidirectional camera flow occlusion masks, respectively. In the first column, the adaptive weights fail to locate the black car in the lower left corner of the image, whereas the bidirectional camera flow occlusion masks succeed in locating the car. The second column shows the opposite situation.}\label{fig:occ_complementry}	
	\vspace{-10pt}
\end{figure}
To better understand the contribution of each element of the objective function proposed in section \ref{sec:method} --- the bidirectional weighted photometric function, which is composed of the bidirectional photometric function ($L_{p}^{bi}$) with bidirectional camera flow occlusion masks  ($M_{occ}^{bi}$) and adaptive weights ($W_{aw}^{bi}$), the bidirectional feature perception loss ($L_{feat}^{bi}$), and the bidirectional depth structure consistency loss ($L_{dsc}^{bi}$) --- to the whole performance, we performed ablation studies, as shown in Tab. \ref{tab:diff_func_ablation_80m_50m} and Tab. \ref{tab:diff_objective_function_camera_pose}.

For our ablation studies, we jointly trained DepthNet and CameraNet with the same network architecture utilizing the proposed objective function combined in different ways in accordance with the idea of the control variable method. Note that the smoothness loss and the SSIM were used by default in all experiments. Tab. \ref{tab:diff_func_ablation_80m_50m} shows the depth estimation results within the range of 80 m and 50 m obtained with different objective function combinations. The corresponding camera pose estimation results are reported in Tab. \ref{tab:diff_objective_function_camera_pose}.

 \begin{figure*}[htbp] 
 	\centering	
 	\includegraphics[scale=0.13]{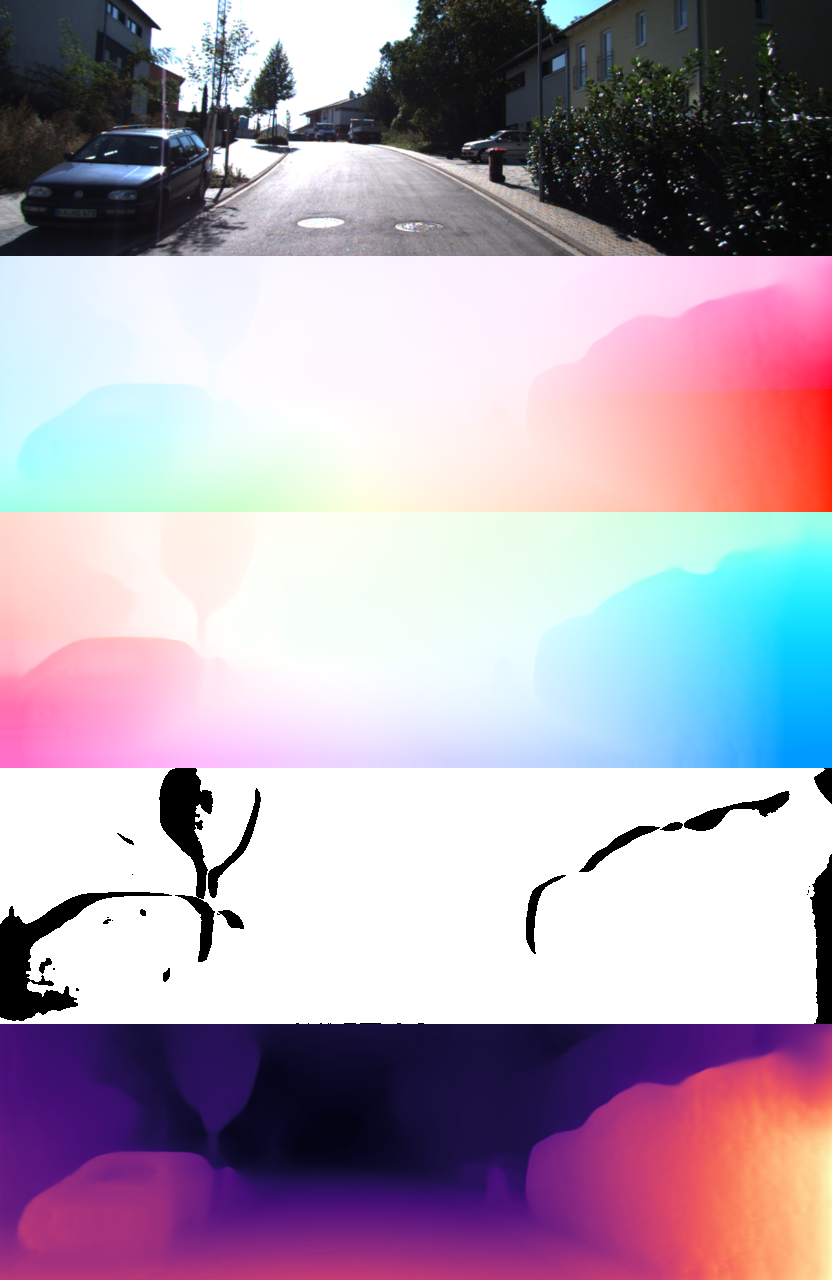}
 	\includegraphics[scale=0.13]{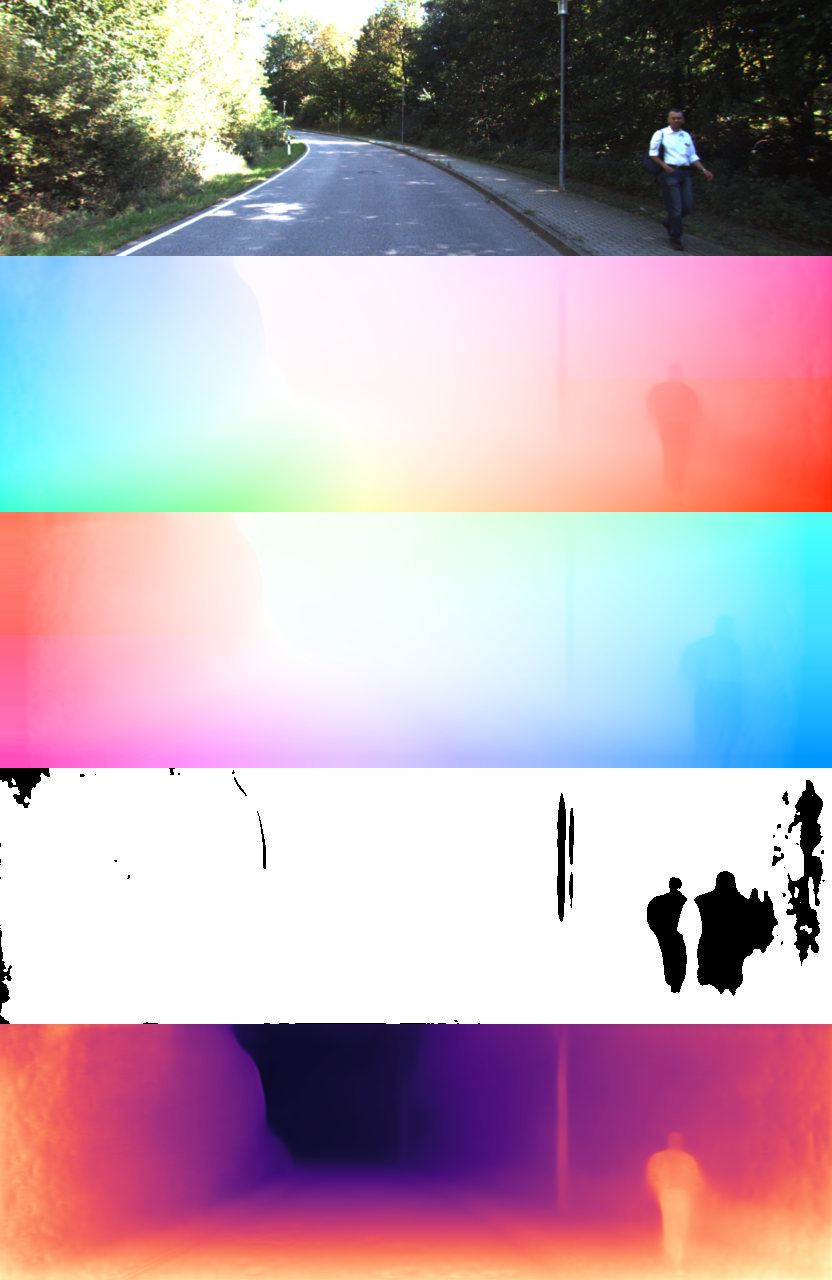}
 	\includegraphics[scale=0.13]{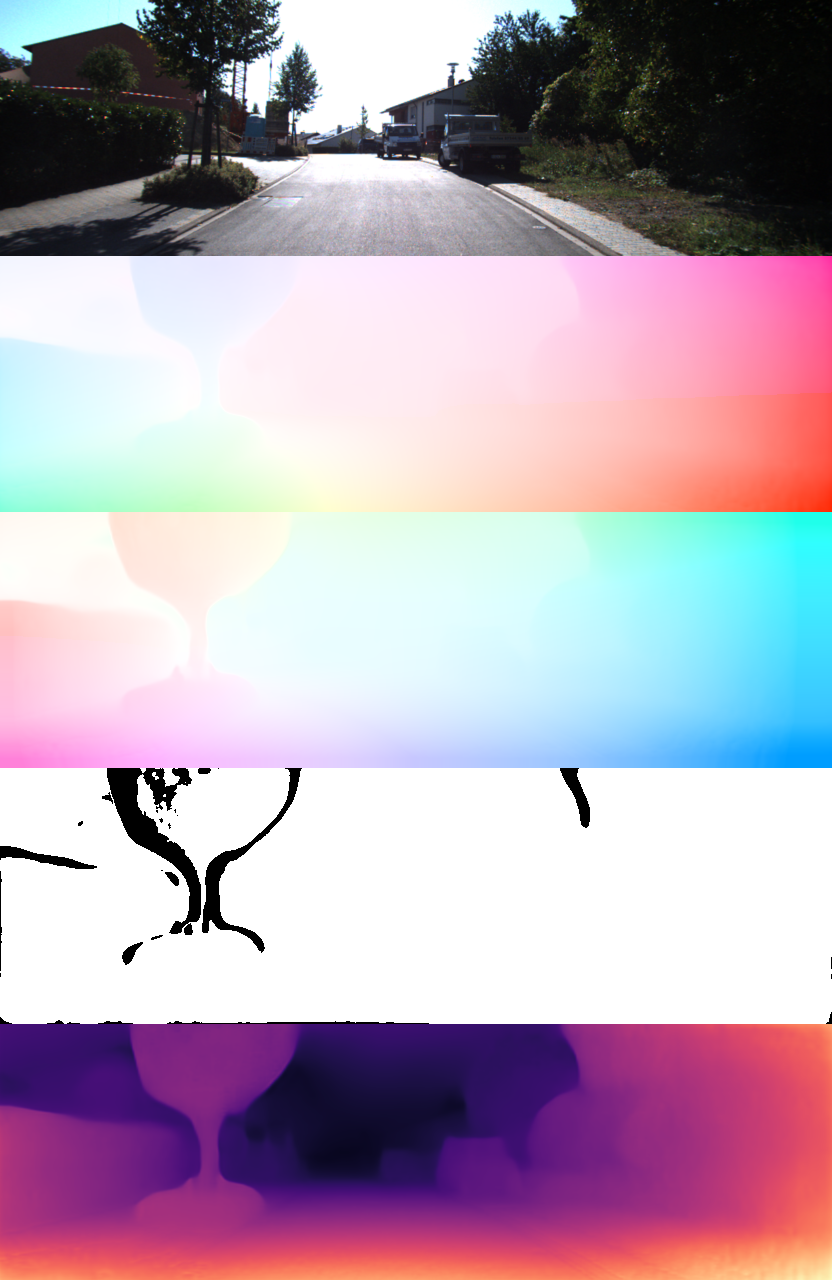}
 	\includegraphics[scale=0.13]{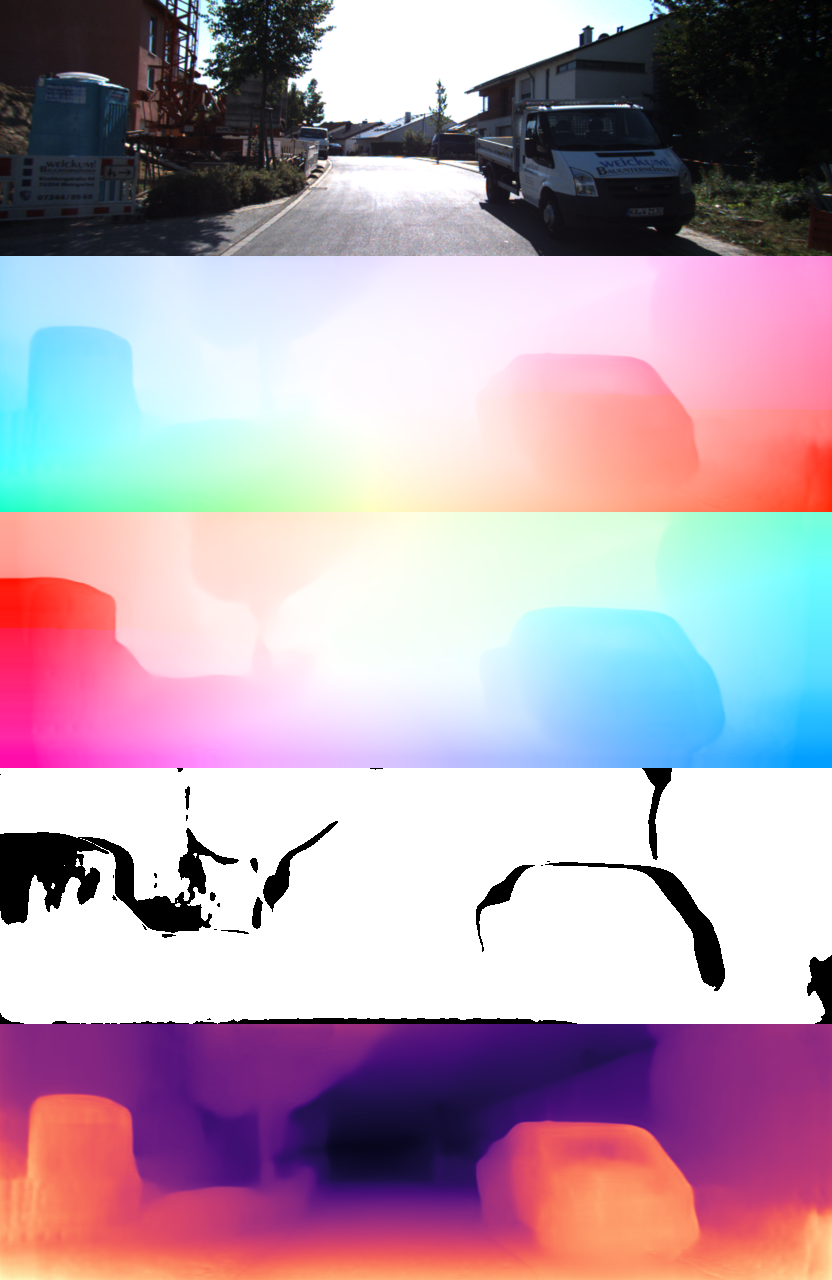}
 	\caption{Visualization analysis of the camera flow occlusion masks. From top to bottom are the original images, the forward camera flows generated from the projection transformation of the reference images to the target images, the backward camera flows synthesized using the differentiable bilinear sampling mechanism, the bidirectional camera flow occlusion masks obtained by checking the consistency of the above two flows, and the estimated depth maps, respectively.} \label{fig:cameraflow_occ}	
 	\vspace{-10pt}	
 \end{figure*}
 \begin{figure*}[htbp]
 	\centering	
 	\includegraphics[scale=0.13]{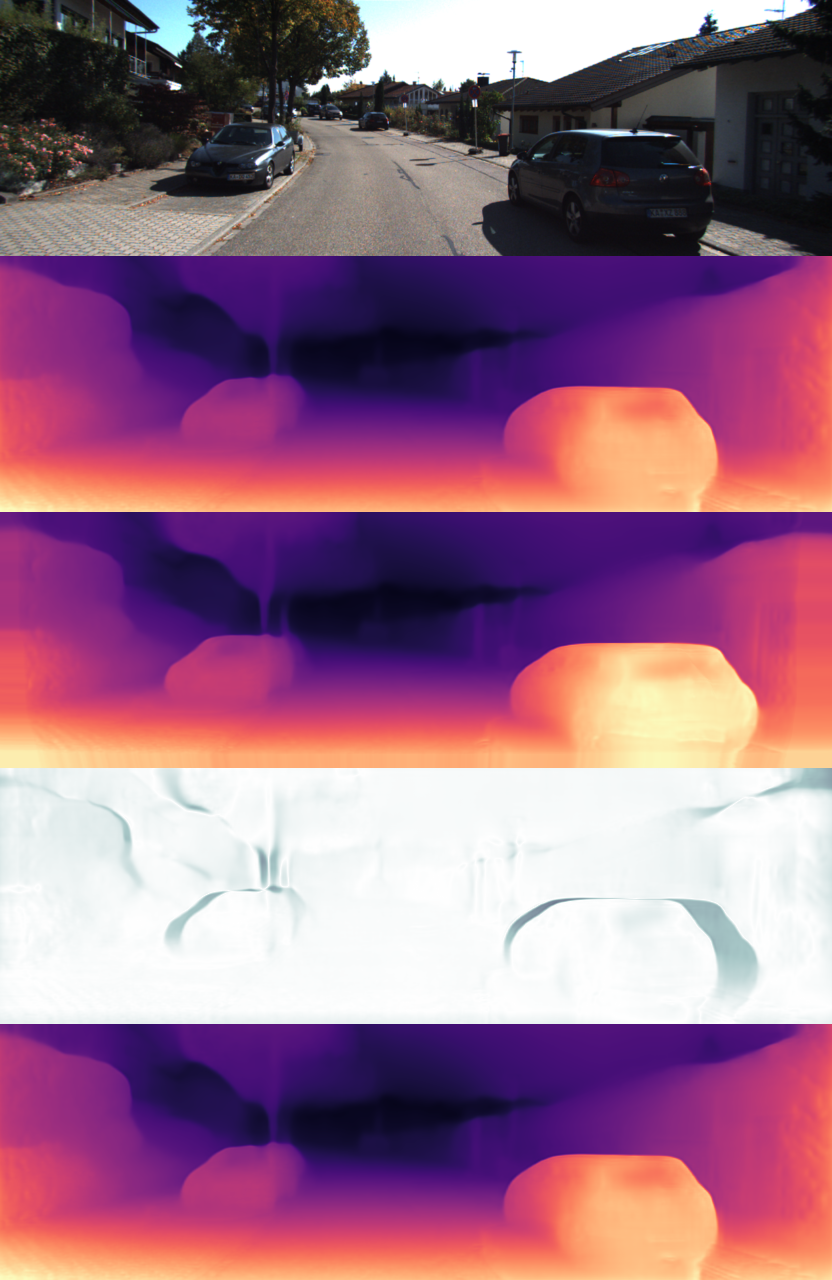}
 	\includegraphics[scale=0.13]{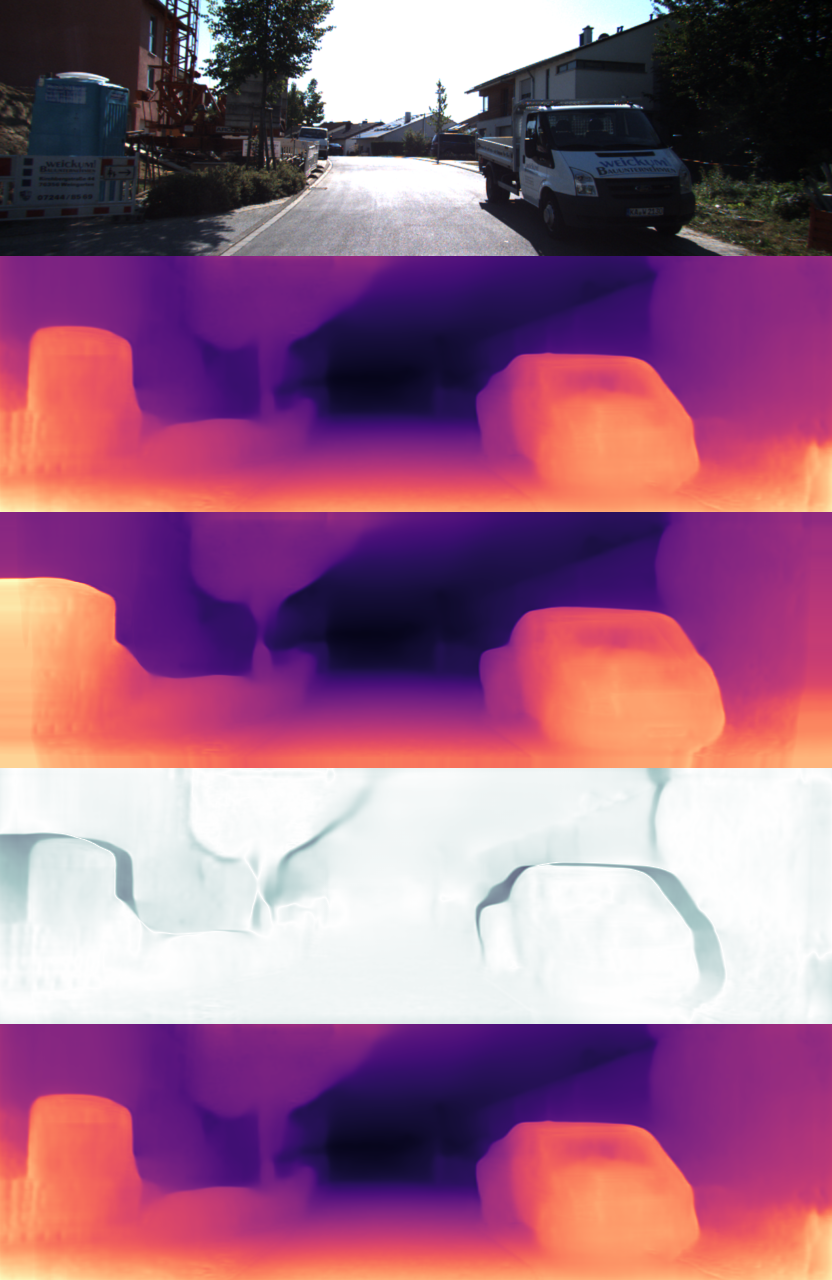}
 	\includegraphics[scale=0.13]{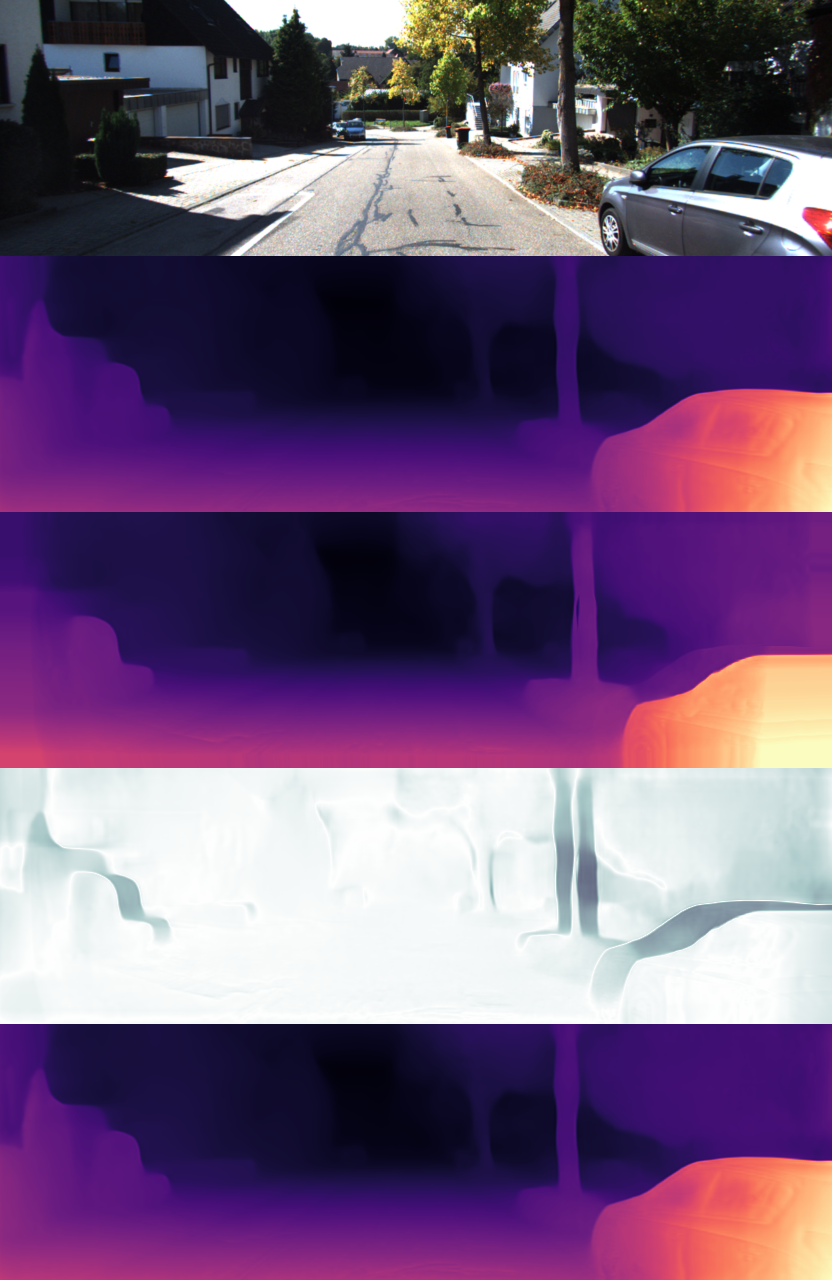}
 	\includegraphics[scale=0.13]{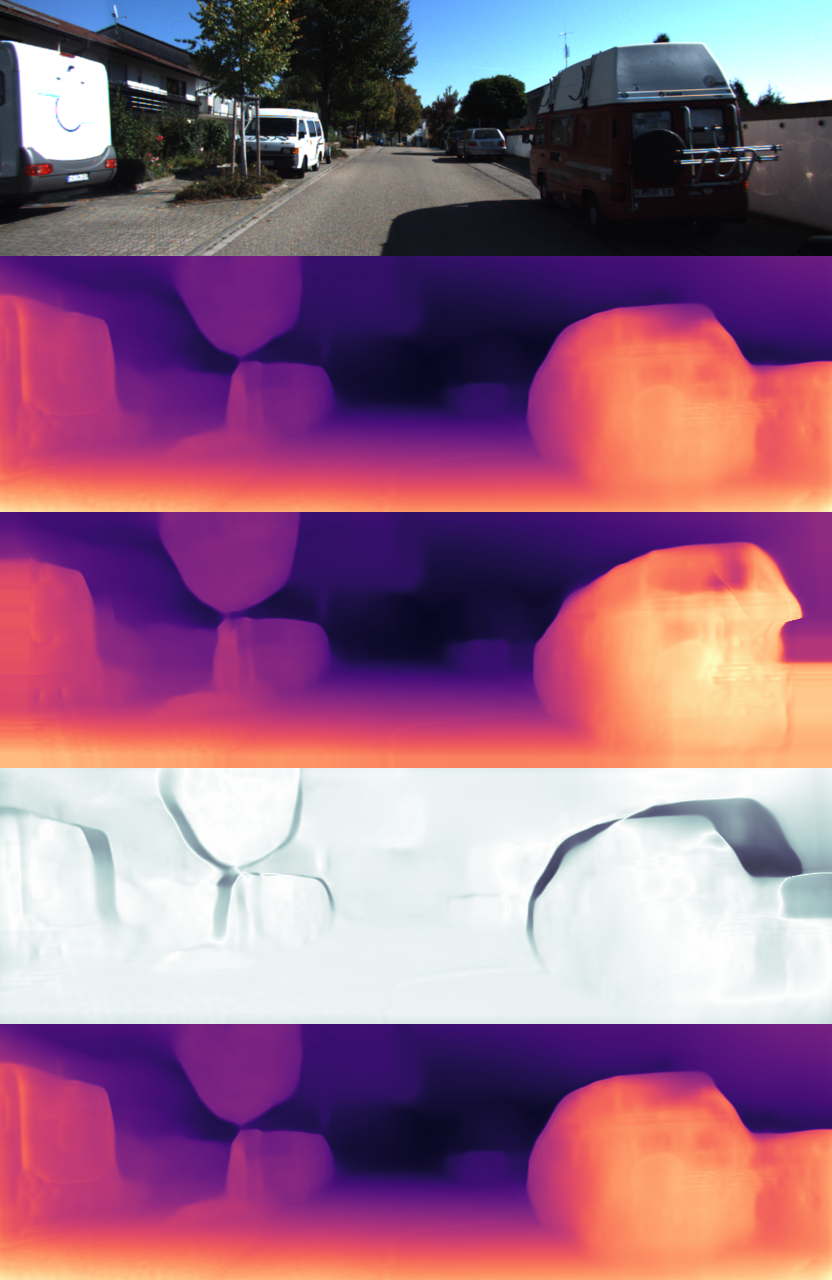}
 	\caption{Visualization analysis of the adaptive weights. From top to bottom are the original images, the depth maps obtained through projection transformation, the depth maps obtained through view synthesis, the adaptive weights obtained by comparing the above two depth maps, and the estimated depth maps, respectively.} \label{fig:depthocc}
 	\vspace{-10pt}
 \end{figure*}
 
 \begin{figure*}[htbp]	
 	\centering	
 	\includegraphics[scale=0.13]{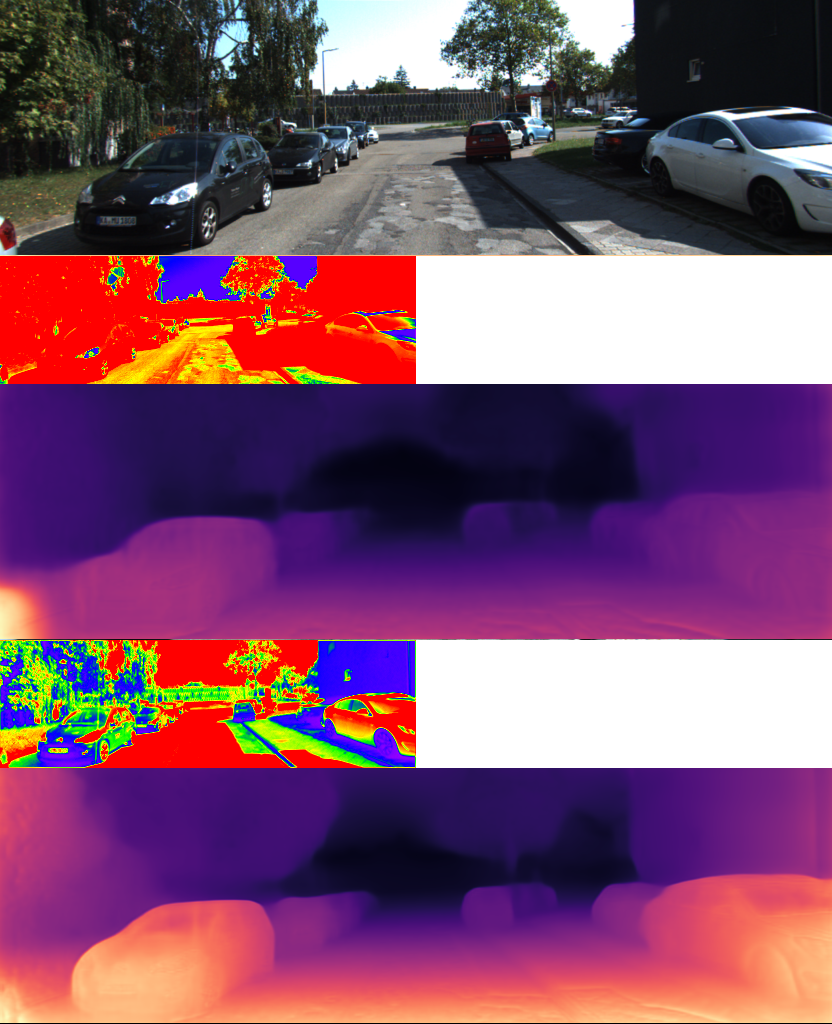}
 	\includegraphics[scale=0.13]{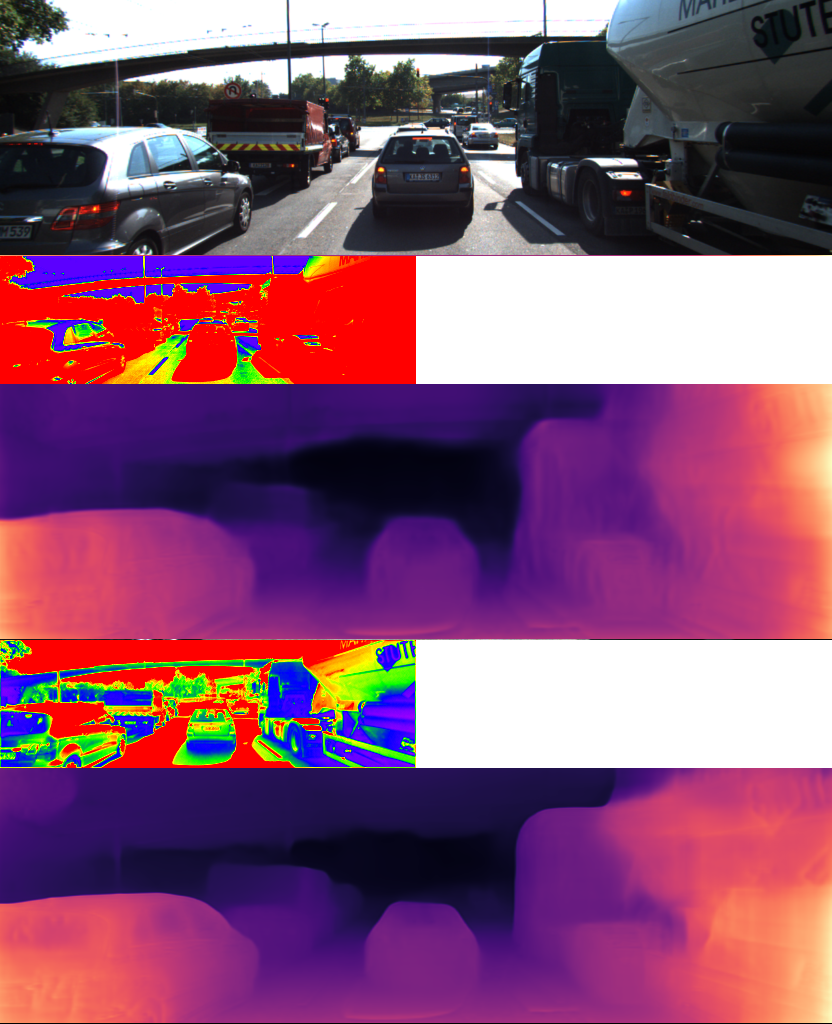}
 	\includegraphics[scale=0.13]{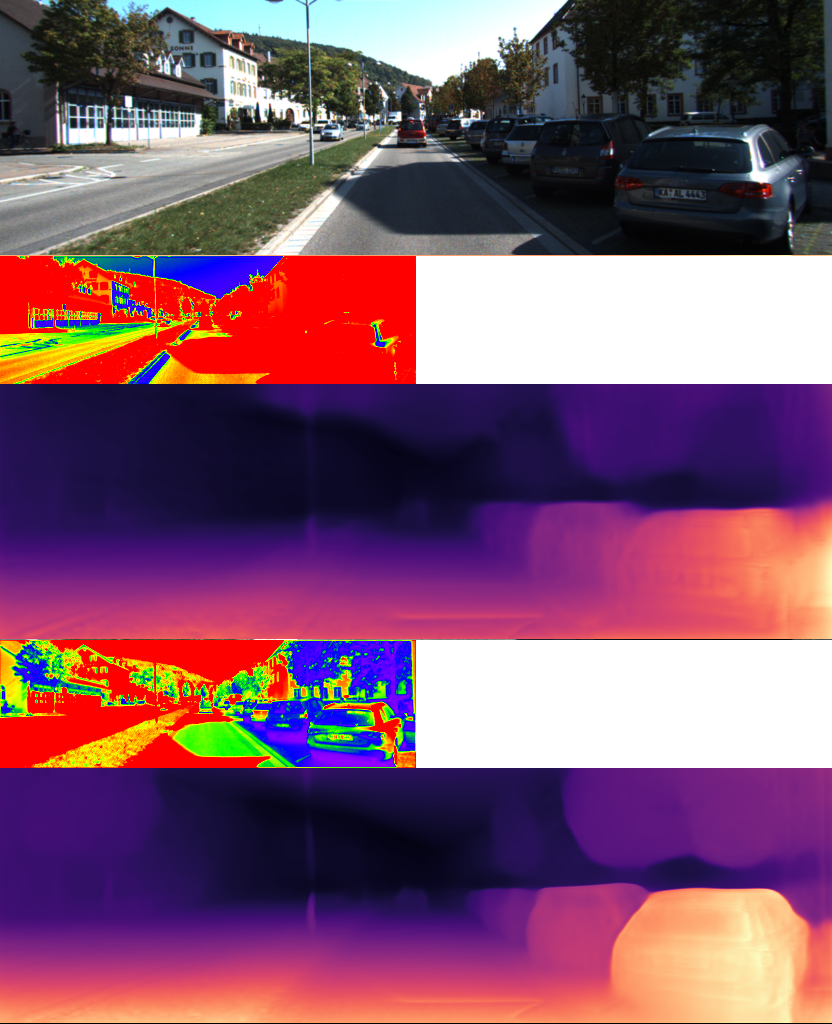}
 	\includegraphics[scale=0.13]{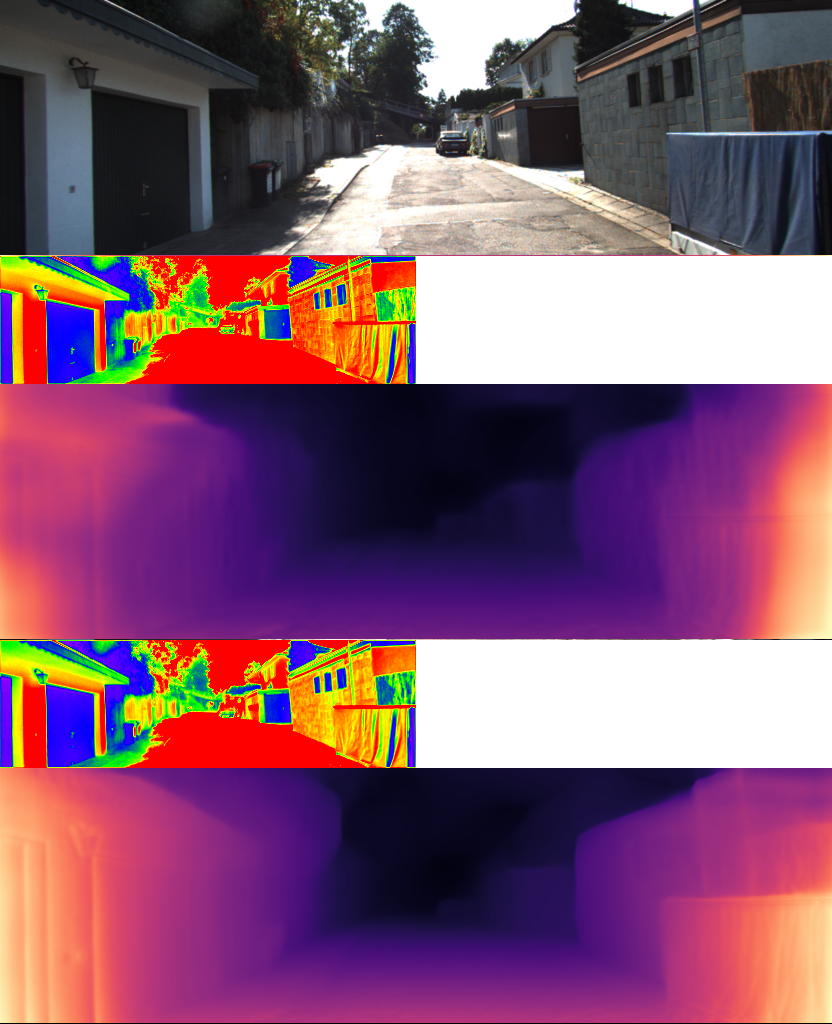}
 	\caption{Visualization analysis of the learned feature maps. Here, we select only the principal feature map for visualization utilizing principal component analysis. From top to bottom are the original images, the feature maps without the bidirectional feature perception loss, the depth maps without the bidirectional feature perception loss, the feature maps with the bidirectional feature perception loss, and the depth maps with the bidirectional feature perception loss, respectively.} \label{fig:featvis}
 	\vspace{-10pt}
 \end{figure*}
To analyze the performance changes caused by the proposed bidirectional photometric function, as the baseline method, we trained DepthNet and CameraNet with the same network architecture utilizing only the reconstruction error between the target view and the warped reference view. The results in Tab. \ref{tab:diff_func_ablation_80m_50m} indicate that the performance metric $\delta <1.25$ could be significantly improved using the proposed bidirectional photometric function; simultaneously, the corresponding ATE of CameraNet was greatly reduced, as seen from the data in Tab. \ref{tab:diff_objective_function_camera_pose}. Moreover, to further verify the effect of the bidirectionality component on the objective function obtained by improving the normal photometric function utilizing the proposed other component, we also conducted additional experiments on the objective function composed of different unidirectionality components, such as the unidirectional photometric function ($L_{p}$), the unidirectional camera flow occlusion masks  ($M_{occ}$), the unidirectional adaptive weights ($W_{aw}$), the unidirectional feature perception loss ($L_{feat}$), and the unidirectional depth structure consistency loss ($L_{dsc}$). The results show that the error indexes all have a different degree of decline compared with the corresponding unidirectional method, except $SqRel$, which is generated from $Baseline$ and $L_{p}^{bi}$, respectively. We hypothesized that this phenomenon might be caused by the presence of occluding objects in the scene. As analyzed above, the absolute trajectory error has a similar trend.

To better understand whether the bidirectional camera flow occlusion masks and adaptive weights play important roles in handling moving objects and occlusions in a scene during inference, we visualized these model components as shown in Fig. \ref{fig:cameraflow_occ} and Fig. \ref{fig:depthocc}. Both the bidirectional camera flow occlusion masks and the adaptive weights can effectively locate moving objects and occlusions in a scene as seen from Fig. \ref{fig:cameraflow_occ} and Fig. \ref{fig:depthocc}. In addition, it can be seen from the quantitative experimental results in Tab. \ref{tab:diff_func_ablation_80m_50m} and Tab. \ref{tab:diff_objective_function_camera_pose} that the performances of both DepthNet and CameraNet are individually improved. Therefore, the moving objects and occlusions in a scene can be well handled. As a result, the implicit assumptions necessary for view synthesis --- that there are no moving objects or occlusions in the scene of interest --- can be satisfied. Note that although moving objects and occlusions can be located using these two components of our method, they cannot always be located successfully if only one component is used (see, e.g.,  Fig. \ref{fig:occ_complementry}). This phenomenon is also supported by the quantitative results shown in Tab. \ref{tab:diff_func_ablation_80m_50m}. More importantly, the bidirectional depth structure consistency constraint employed to obtain the adaptive weights can also improve the quality of the estimated depth maps, as seen in Tab. \ref{tab:diff_func_ablation_80m_50m}.

To analyze the effects of the bidirectional feature perception loss on the model, we selected the principle feature map formed from the features extracted by the encoder and analyzed it using principal component analysis. The visualization results are shown in Fig. \ref{fig:featvis}, where the original images, the feature maps without bidirectional the feature perception loss, the depth maps without the bidirectional feature perception loss, the feature maps with the bidirectional feature perception loss, and the depth maps with the bidirectional feature perception loss are sequentially shown in the first to fifth rows. As seen from Fig. \ref{fig:featvis}, compared to those learned without the bidirectional feature perception loss, the visual representations learned with the bidirectional feature perception loss show larger variations in textureless regions, such as the pure white/black cars in the first column, the white oil tank in the second column, the the shadows of the cars in the second and third columns, and the wall in the fourth column. The corresponding estimated depth maps are also smoother and sharper, consistent with the quantitative results in Tab.  \ref{tab:diff_func_ablation_80m_50m}.

\begin{table}[htbp]\small
	\vspace{-2pt}
	
	\setlength\tabcolsep{2pt}
	\centering	

	\begin{tabular}{lcccccc} 
		\toprule 
		\multicolumn{1}{l}{CameraNet}&
		\multicolumn{1}{l}{Scheme}& \multicolumn{1}{c}{TrainT({ms})}& \multicolumn{1}{c}{GPU({\scriptsize M})}\\
		\hline 
	
		\multirow{2}{*}{PN7 }&	
		Pinv &116.9&2903\\&
		Cinv(Ours) &74.5&2815\\
		\hline
		
		\multirow{2}{*}{RN18 }&	Pinv &151.3&3935\\&
		Cinv(Ours) &80.1&3429\\
		\bottomrule 
	\end{tabular}
	\caption{Comparison of acquisition methods of the inverse pose with batchsize=2. All results are evaluated on RTX 3090Ti with the same setting. The resolution is set to 256$\times$832. The time is averaged 1000 iterations on the KITTI RAW dataset. `Cinv/Pinv' indicates that compute inverse pose/predict inverse pose. }\label{tab:param_time_gpu_compare_train_inverse}
	\vspace{-10pt}
\end{table}

In Tab. \ref{tab:param_time_gpu_compare_train_inverse}, we investigate the differences between acquisition methods of the inverse pose. The results in Tab. \ref{tab:param_time_gpu_compare_train_inverse} shows that calculating the inverse pose is better than predicting the inverse pose in terms of both training time and required GPU. Furthermore, this advantage becomes more significant as the depth of the model increases. More importantly, compared with the scheme of predicting the inverse using the network, both the forward and backward transformations are explicitly constrained to be invertible and the scale of both the forward and backward poses is also explicitly constrained to be consistent.

\vspace{-10pt}
\section{Conclusions}\label{sec:conclusion}
In this paper, we have presented an end-to-end self-supervised learning pipeline that utilizes the task of view synthesis to obtain the supervision signal for depth and camera pose estimation from unlabeled monocular video. Our experimental results indicate that the proposed method outperforms previous related work. The proposed bidirectional weighted photometric loss function can fully reveal the information captured by the limited available data and handle dynamic scenes effectively. Second, textureless regions in a scene can be given more attention by using our feature perception loss function. Moreover, we can enforce consistency between depth maps, further improving the quality of the depth estimates. Despite competitive performance in a benchmark evaluation, the scale-drift issue still remains, causing us to need to align our estimation results with the ground truth during evaluation. Additionally, our method assumes that the camera intrinsics are given, thus preventing its application to arbitrary Internet videos acquired with unknown camera types. We plan to address these problems in future work. 

{
\bibliographystyle{IEEEtran}
\bibliography{reference}

}
\vspace{-10pt}
\begin{IEEEbiography}
	[{\includegraphics[width=0.8in,height=1.in,clip]{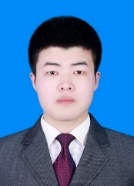}}]
	{Fei Wang} is currently pursuing the Ph.D. degree in University of Chinese Academy of Sciences, Shenzhen Institute of Advanced Technology. His current research interests include computer vision, structure from motion, robotics and deep learning.
\end{IEEEbiography}
\vspace{-10pt}
\begin{IEEEbiography}
	[{\includegraphics[width=0.8in,height=1.in,clip]{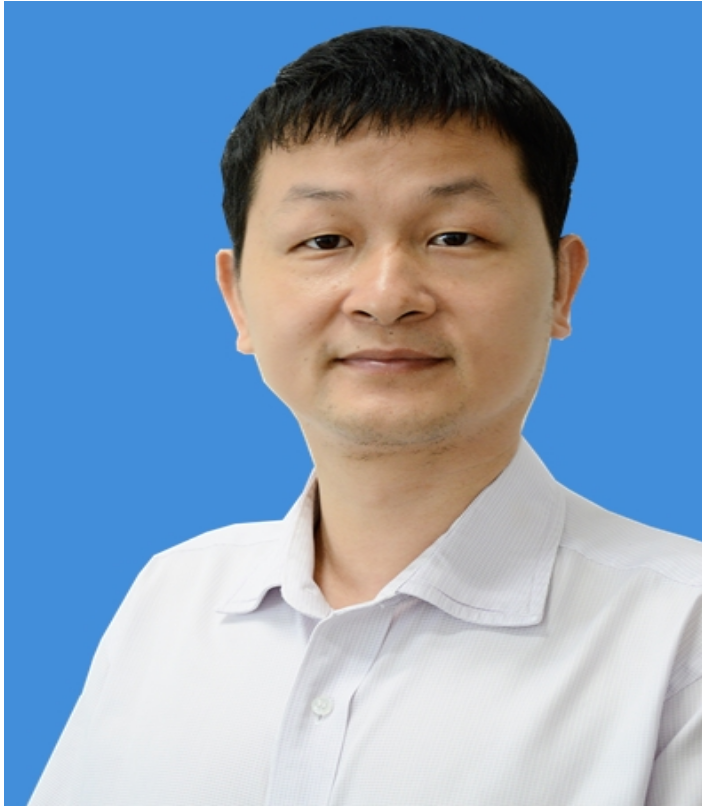}}]
	{Jun Cheng} received the B.Eng. and M.Eng. degrees from the University of Science and Technology of China, Hefei, China, in 1999 and 2002, respectively, and the Ph.D. degree from The Chinese University of Hong Kong, Hong Kong, in 2006. He is currently with the Shenzhen Institute of Advanced Technology, Chinese Academy of Sciences, Shenzhen, China, as a Professor and the Director of the Laboratory for Human Machine Control. His current research interests include computer vision, robotics, machine intelligence, and control.
\end{IEEEbiography}
\vspace{-10pt}
\begin{IEEEbiography}
	[{\includegraphics[width=0.8in,height=1.in,clip]{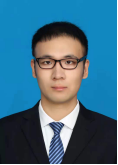}}]
	{Penglei Liu} is currently pursuing the Ph.D. degree in University of Chinese Academy of Sciences, Shenzhen Institute of Advanced Technology. His current research interests include robot control, neural network applications and machine learning.
\end{IEEEbiography}
\end{document}